\documentclass[10pt,twocolumn,letterpaper]{article}

\usepackage[pagenumbers]{cvpr} %

\newcommand{\method}{\mbox{StereoSpace}}

\usepackage{enumitem}
\usepackage{amssymb}
\usepackage{wrapfig}
\usepackage{subcaption}
\usepackage{arydshln}     %
\usepackage[table,dvipsnames]{xcolor} %
\usepackage{siunitx}
\usepackage{svg}
\usepackage{pifont}
\usepackage{mathtools}
\usepackage{multirow} 
\usepackage{soul}
\usepackage{comment}
\usepackage{overpic}

\usepackage{microtype}
\usepackage{inconsolata}

\usepackage{fontawesome5}

\newcommand{\envtree}{\textcolor{ForestGreen}{\faTree}}
\newcommand{\envhouse}{\textcolor{Cyan}{\faHome}} 
\newcommand{\baseshuffle}{\textcolor{orange}{\faRandom}}

\definecolor{rulegray}{gray}{0.75}
\definecolor{baselinered}{rgb}{1,0,0}
\newcommand{\tabdividerlegend}{%
  \textcolor{rulegray}{\makebox[8mm][s]{\dotfill}}%
}

\newcommand{\mdown}[1]{\textbf{#1}\,(\ensuremath{\downarrow})}

\definecolor{firstcolor}{HTML}{BDE6CD}%
\definecolor{secondcolor}{HTML}{E2EEBC}%
\definecolor{thirdcolor}{HTML}{FFF8C5}%

\newcommand{\fst}[1]{\cellcolor{firstcolor}\bfseries #1}
\newcommand{\snd}[1]{\cellcolor{secondcolor}#1}
\newcommand{\trd}[1]{\cellcolor{thirdcolor}#1}

\definecolor{cvprblue}{rgb}{0.21,0.49,0.74}
\usepackage[pagebackref,breaklinks,colorlinks,allcolors=cvprblue]{hyperref}

\def\paperID{XXX} %
\def\confName{XXX}
\def\confYear{XXX}

\def\paperID{57} 
\def\confName{CVPR}
\def\confYear{2026}

\title{StereoSpace: Depth-Free Synthesis of Stereo Geometry \\ via End-to-End Diffusion in a Canonical Space}

\author{\!\!\!\!Tjark Behrens$^1$ \quad\!\! Anton Obukhov$^3$ \quad\!\! Bingxin Ke$^1$ \quad\!\! Fabio Tosi$^2$ \quad\!\! Matteo Poggi$^2$ \quad\!\! Konrad Schindler$^1$%
\\
\small{
$^1$ETH Zurich \quad\quad\quad $^2$University of Bologna \quad\quad\quad $^3$HUAWEI Bayer Lab
} \\
\small{
Project page: \href{https://hf.co/spaces/prs-eth/stereospace}{https://hf.co/spaces/prs-eth/stereospace}
}
}

\begin{document}

\twocolumn[{
\renewcommand\twocolumn[1][]{#1}
\maketitle
\begin{center}
  \centering
  \vspace{-0.5cm}
  \includegraphics[width=\linewidth]{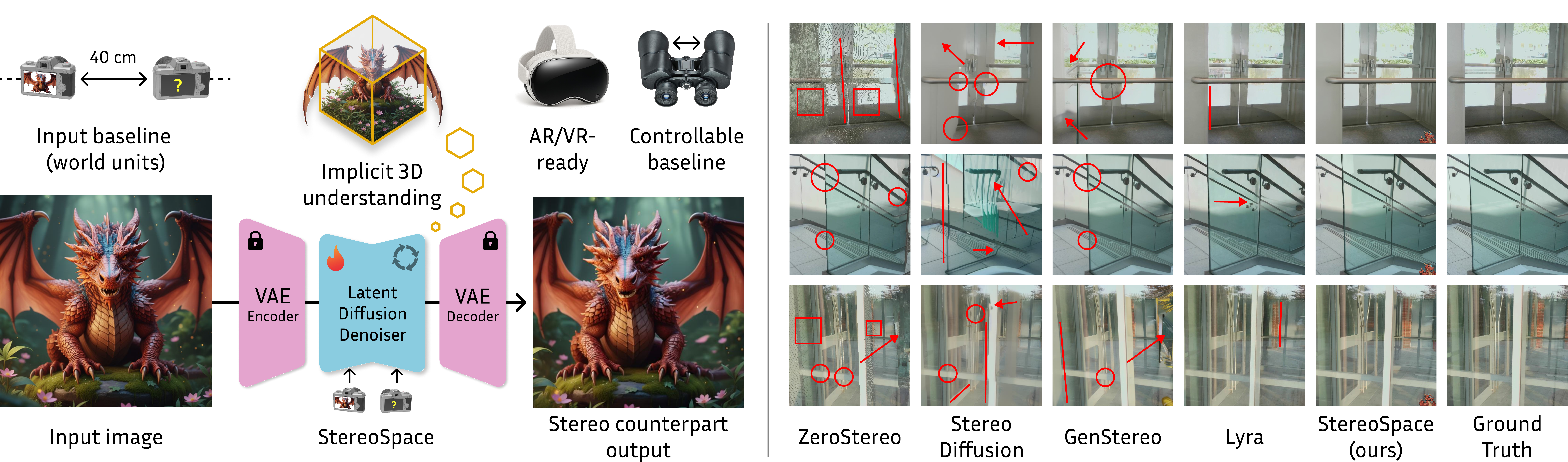}\vspace{-0.2cm}
  \captionof{figure}{
    \textbf{\method{} for generating stereo 
    from monocular images.
    } 
    \textit{Left:}
    Built on a foundational LDM, our framework efficiently leverages learned priors for end-to-end view synthesis.
    The target baseline in world units acts as conditioning for precise view control.
    \textit{Right:}
    Implicit scene understanding allows us to tackle the most complex cases where geometry cues alone are insufficient for novel view synthesis.
    Best viewed zoomed-in. 
    Legend: 
    \mbox{\raisebox{-0.2ex}{$\square$} (warping)}, 
    \mbox{\raisebox{0.15ex}{\scriptsize$\bigcirc$} (breaks)}, 
    \mbox{--- (bends)}, 
    \mbox{$\rightarrow$ (ghosting)}.
    \method{} consistently outperforms recent monocular competition, including generative 3DGS models like Lyra~\cite{bahmani2025lyra}.
  }
  \label{fig:teaser}
\end{center}
}]

\begin{abstract}
We introduce StereoSpace, a diffusion-based framework for monocular-to-stereo synthesis that models geometry purely through viewpoint conditioning, without explicit depth or warping. A canonical rectified space and the conditioning guide the generator to infer correspondences and fill disocclusions end-to-end. To ensure fair and leakage-free evaluation, we introduce an end-to-end protocol that excludes any ground truth or proxy geometry estimates at test time. The protocol emphasizes metrics reflecting downstream relevance: iSQoE for perceptual comfort and MEt3R for geometric consistency. StereoSpace surpasses other methods from the warp \& inpaint, latent-warping, and warped-conditioning categories, achieving sharp parallax and strong robustness on layered and non-Lambertian scenes. This establishes viewpoint-conditioned diffusion as a scalable, depth-free solution for stereo generation.
\end{abstract}
\section{Introduction}
\label{cpt1}
\enlargethispage{\baselineskip}
Stereo imaging provides a principled way to infer three-dimensional structure from two-dimensional observations captured from slightly displaced viewpoints.

As this mechanism mirrors human depth perception, stereo imaging also represents a key technology for spatially immersive entertainment. In 3D cinema, virtual reality, and augmented reality, presenting slightly offset stereo images to human eyes stimulates binocular disparity and produces a convincing illusion of depth for the viewer. 
This exploitation of a biologically grounded perception mechanism enhances visual realism and audience engagement, making stereo imaging not only a tool for 3D reconstruction but also a cornerstone in spatially immersive visual media.
\enlargethispage{\baselineskip}

However, acquiring high-quality stereo imagery for entertainment purposes can be both costly and technically demanding, as it requires precise camera alignment, synchronization, and calibration, with even minor mismatches possibly leading to visual discomfort or distortion in the perceived 3D effect. 
Consequently, generating stereo pairs out of single monocular images has emerged as an attractive alternative for reducing production costs and potentially enable conversion of any 2D content into 3D media.

The simplest and most intuitive way to achieve this is to recover the depth of the single image and use it to project its pixels over another camera frame placed on its right, e.g., by a simple forward warping operation \cite{wang2024stereodiffusion,wang2025zerostereo,qiao2025genstereo}. 
This approach assumes the availability of a reliable and generalizable depth estimator, which is a reasonable requirement given the latest advances in single-image depth estimation~\cite{yang2024depth,wang2025moge}. As a result, the stereo generation process becomes akin to an inpainting task, aimed at filling empty pixels after warping rather than a 3D generative task.
However, we argue that this shortcut exposes the entire generation process to failure cases inherited from the depth estimator itself. Among these, the presence of multiple depth layers within a scene -- occurring, for instance, in the presence of glass or transparent surfaces -- represents a particularly severe challenge, given the inability of depth estimation models to deal with such complex structures.

Driven by these arguments, in this paper we reformulate the stereo generation problem as an image-conditioned 3D generative task by getting rid of depth estimation as a preliminary requirement. Inspired by the recent success achieved by the simple repurposing of diffusion-based image generators for dense predictive tasks \cite{ke2024repurposing,ke2025marigold}, we follow the same path and design a novel depth-free framework for stereo image generation.
The key to compensating for the lack of structural guidance provided by depth is a proper encoding of relative displacements between viewpoints in a metric, pose-canonicalized frame.
Relative extrinsics and dense intrinsics encodings define a canonical \texttt{\textbf{\method{}}}, which suggests the name for our framework.
This lets users set the stereo baseline $B$ directly in physical units at inference, yielding predictable control and generalization across baselines, as shown in Fig. \ref{fig:teaser} on the left.

We build \method{} around the rich text-to-image generative prior of Stable Diffusion~\cite{salimans2022progressive}.
A mixture of off-the-shelf single-baseline stereo datasets and custom-made multi-baseline datasets, rendered by means of novel view synthesis techniques \cite{Tosi2023Nerf, li2025scenesplat}, ensures efficient transfer of the generative prior to the downstream task. 
To assess \method{}'s generation quality, we establish a novel evaluation protocol based on iSQoE \cite{tamir2025isqoe}, aimed at quantifying perceptual comfort, and MEt3R \cite{asim2025met3r}, to evaluate geometric consistency of the generated image with the input.
This evaluation is carried out on four real-world stereo datasets, covering both indoor scenes and outdoor driving scenarios, as well as multi-layered structures where conventional, depth-based approaches struggle, as shown in the right part of the teaser figure.
Our results confirm the effectiveness of \method{} in generating stereo images despite the lack of explicit depth estimation performed in advance.

Our main contributions can be summarized as follows:

\begin{itemize}
\item \method{} for single-image conditional generation of counterpart views, free from explicit geometric shortcuts 
\item End-to-end training procedure for efficient transfer of the rich task-agnostic foundation prior to the task at hand
\item A novel perceptual and geometry-aware evaluation
\end{itemize}

\section{Related Work}
\label{cpt2}

\noindent\textbf{Novel View Synthesis (NVS)}. Recent breakthroughs in novel view synthesis are rooted in neural rendering and implicit and explicit 3D scene representations. Neural Radiance Fields (NeRFs) introduced implicit neural representations by mapping spatial coordinates to density and radiance, rendered via differentiable volume integration. Despite capturing fine geometry and view-dependent effects, NeRFs require slow per-scene optimization and dense ray marching. Subsequent work improved efficiency~\cite{muller2022instant, chen2022tensoRF}, anti-aliasing~\cite{barron2021mipnerf,barron2022mipnerf360}, sparse-view reconstruction~\cite{yu2021pixelnerf, chen2021mvsnerf}, and dynamic scene handling~\cite{park2021nerfies, pumarola2020d, gao2021dynamic}. NeRFs trained from monocular sequences have also been exploited to generate stereo training data by rendering rectified stereo pairs with controllable baselines~\cite{Tosi2023Nerf}.
3D Gaussian Splatting (3DGS)~\cite{kerbl20233d}, instead, introduced an explicit representation based on anisotropic Gaussian primitives rendered by efficient rasterization, enabling real-time performance. Building on this, recent works enhanced geometry~\cite{huang20242d}, appearance~\cite{jiang2024gaussianshader}, dynamics~\cite{luiten2024dynamic}, and generalization~\cite{charatan2024pixelsplat, chen2024mvsplat}.
While these methods aim at general novel view synthesis from arbitrary views in a multi-view setting, our work focuses on synthesizing a single translationally-shifted viewpoint (the stereo pair) from a monocular image using a diffusion model.

\noindent\textbf{Diffusion Models for Vision Tasks.} Diffusion Models (DM)~\cite{ho2020denoising,yang2023diffusion} have transformed generative modeling, achieving state-of-the-art results in image~\cite{dhariwal2021diffusion}  and video synthesis~\cite{ho2022video, blattmann2023align}. Latent Diffusion Models (LDMs)~\cite{rombach2022latent} further improve efficiency by operating in compressed latent space, while conditional variants such as ControlNet~\cite{zhang2023adding} and T2I-Adapter \cite{mou2024t2i} enable fine-grained control through structural inputs (\eg, depth or edges). Beyond generation, DMs have been adapted for structured prediction, including monocular depth or surface normals~\cite{ke2025marigold,ke2024repurposing, qiu2024richdreamer, fu2024geowizard}, segmentation~\cite{baranchuk2021label,xu2023open,rahman2023ambiguous,tian2024diffuse}, object detection~\cite{chen2023diffusiondet,he2024diffusion}, and inpainting~\cite{lugmayr2022repaint}. For multi-view generation, existing methods generate multiple consistent views that span object-centric~\cite{shi2023mvdream, liu2023zero} and scene-level~\cite{sargent2024zeronvs, tang2023mvdiffusion} settings. Some approaches~\cite{gao2024cat3d} reconstruct multi-view images into explicit 3D representations via expensive per-scene optimization, while camera-controlled video diffusion models~\cite{wang2024motionctrl, zhou2025stable, he2024cameractrl, xu2024camco, bahmani2025ac3d} enable viewpoint conditioning, with recent works adopting Plücker coordinates for pixel-wise control; feed-forward 3D models~\cite{bahmani2025lyra, liang2024wonderland} then distill scene knowledge from such video priors. Following the paradigm of camera-conditioned diffusion, we adapt LDMs for stereo generation, synthesizing geometrically consistent stereo pairs through a dual-stream architecture conditioned on Plücker ray embeddings.

\noindent\textbf{Monocular-to-Stereo Generation.} Research on monocular-to-stereo synthesis has evolved from explicit 
geometry-based pipelines to diffusion-driven generation. Early work, such as Deep3D~\cite{xie2016deep3d}, introduced end-to-end neural architectures 
trained on stereo pairs to predict disparity-like maps for direct view 
synthesis. Recent diffusion-based methods, instead, can be categorized into distinct paradigms. \textit{Warp-and-inpaint} approaches~\cite{wang2025zerostereo, zhang2024spatialme, dai2024svg, zhao2024stereocrafter, shvetsova2025m2svid, huang2025restereo,lv2024spatialdreamer} estimates disparity from monocular depth, forward-warp the input image, and inpaint disoccluded regions using diffusion priors, with~\cite{yu2025mono2stereo} focusing specifically on tiny-baseline setups.
\textit{Latent warping}~\cite{wang2024stereodiffusion, shi2024stereocrafterzero} performs disparity shifts directly in diffusion latent space without explicit training, while \textit{Warped conditioning}~\cite{qiao2025genstereo} applies warping to canonical coordinate embeddings that condition the denoising process.  \textit{Training-free} variants~\cite{dai2024svg}, instead, operate without fine-tuning. Other extensions include stereo matching through synthesis~\cite{wang2025zerostereo} and multi-baseline depth estimation~\cite{liu2025dms}.  
Our work differs fundamentally by learning stereo geometry directly via viewpoint conditioning in canonicalized space, enabling metric baseline control through viewpoint conditioning, and achieving \textit{depth-free} handling of complex multi-layer scenes with strong cross-baseline generalization.

\begin{figure*}[t]
  \centering
  \includegraphics[width=\linewidth]{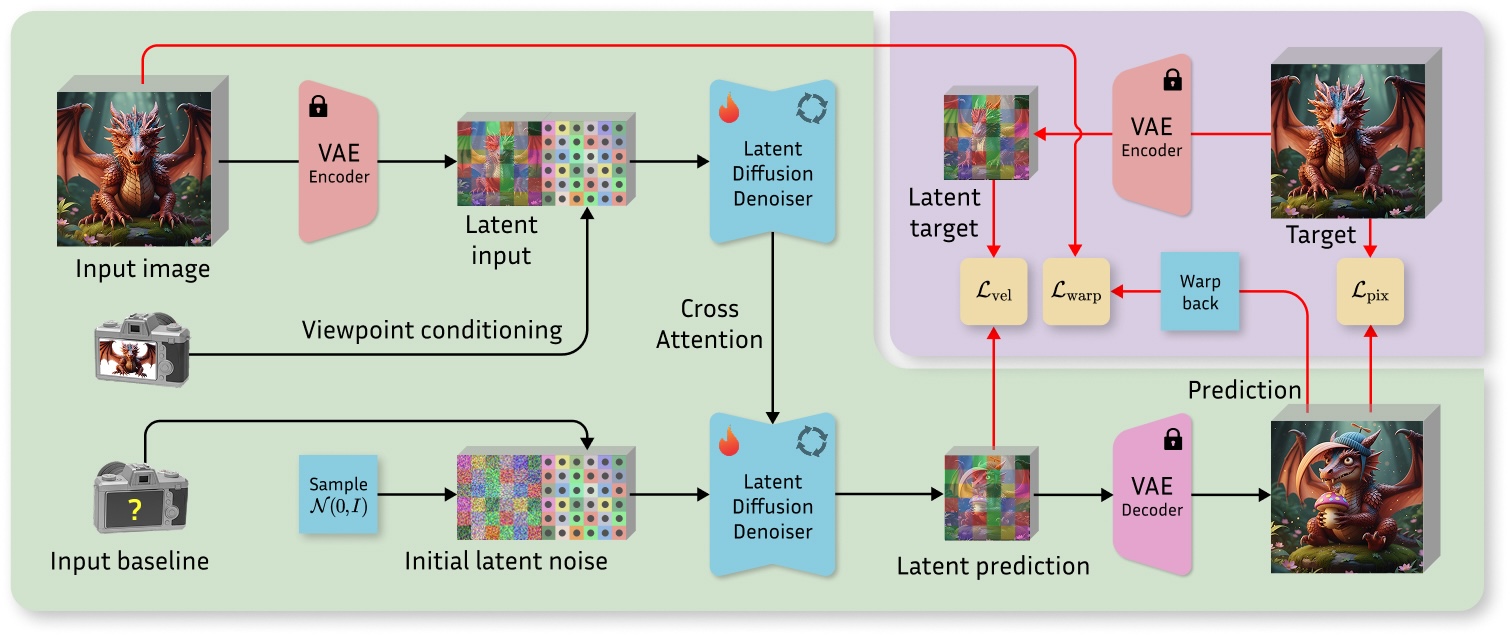}
  \vspace{-0.7cm}
  \caption{\textbf{Architecture overview.} 
  The model uses a dual U-Net initialized from Stable Diffusion~v2.0. 
  The top branch operates on the source view latent as well as the viewpoint condition.
  The target baseline is encoded similarly and is concatenated with the  latent code of the counterpart view.
  Latent and pixel-space losses supervise fine-tuning, wherein target view synthesis leverages source view features through end-to-end cross attention.
  Red arrows denote operations at training time only.
  Refer to Sec.~\ref{cpt3} for details on conditioning and warping.
  }\vspace{-0.3cm}
  \label{fig:stereospace_architecture}
\end{figure*}

\section{Method}
\label{cpt3}

We consider a standard rectified pinhole stereo setting with known intrinsics $f$ and a metric baseline $B$. In rectified stereo, epipolar lines are horizontal and disparities vary only along the image x-axis. Hence, the mapping from source to target view is fully determined by the relative calibration $(f,B)$, rather than by the absolute camera poses in the world. Put differently, in stereo view synthesis, the critical variable is the inter-camera geometry, not absolute position in space.

To exploit this, we introduce \texttt{StereoSpace} that canonicalizes any rectified stereo pair by re-expressing its extrinsics in a shared frame. The teaser figure (Fig.~\ref{fig:teaser}, left) visualizes our formulation of StereoSpace: the center of the stereo rig is fixed at the origin and the two cameras are constrained to lie on the x-axis, separated by the baseline $B$. This creates a common metric baseline along which all cameras move horizontally. Canonicalization in this space concentrates the training distribution: the model no longer needs to explain variation due to arbitrary world poses and can instead focus on the stereo-induced appearance changes and epipolar-consistent correspondences.

Unlike warping-based pipelines, we avoid building an intermediate depth volume. We argue that the representational capacity of a diffusion model is sufficient to learn stereo view synthesis directly in our StereoSpace. In practice, the generative model is conditioned on the source image and the relative calibration $(f,B)$ and predicts the target view. This yields a \emph{depth-free} formulation: geometry is injected through the conditioning variables and the canonical frame, not through an explicit 3D reconstruction.

\subsection{Viewpoint Conditioning}

We formulate stereo view generation as conditional diffusion: given a rectified source view $I_s$, the model synthesizes the paired target view $I_t$ at the horizontally displaced viewpoint, conditioned on the known stereo configuration. We follow the standard latent diffusion setup and train with the velocity parameterization used in recent work~\cite{salimans2022progressive,karras2022elucidating}. Concretely, for a noisy latent $z_t$ of the target, we minimize:
\begin{equation}
    \mathcal{L}_{\mathrm{vel}} = \mathbb{E}_{(I_s, I_t), \epsilon, t} \big[ \, \left\| v - v_\theta\left(z_t, t ; I_s, \Phi\right) \right\|^2 \, \big],
\end{equation}
where $v$ is the ground truth velocity for the chosen noise schedule, $v_\theta$ is the predicted velocity, and $\Phi$ denotes the viewpoint-conditioning signal.

The purpose of $\Phi$ is to impose our canonical \emph{StereoSpace} by providing exact camera control. We add this viewpoint conditioning $\Phi$ as Plücker embeddings in the diffusion process. Each camera ray is represented as a normalized 6D Plücker vector (moment and direction) that encodes the intrinsics and extrinsics along that sightline while being invariant to translations along the ray~\cite{jia2024plucker}. Distances in the StereoSpace are kept metric. Hence, Plücker rays retain a metric structure, and the diffusion process receives a geometry-aware, scale-preserving description of the target camera. Although we use Plücker embeddings as our default parameterization, the method is not tied to it: other camera-aware parameterizations, as well as mixtures thereof, can be substituted without substantial differences (Sec.~\ref{sec:ablation}).

Through the imposed canonicalization into StereoSpace via viewpoint conditioning, a single model can operate across multiple baselines and focal lengths observed during training and can handle datasets containing several stereo rigs, without being tied to a specific calibrated setup, which in turn allows generalization beyond those configurations at inference time.

\subsection{Architecture}

Our model adopts a dual U-Net diffusion backbone for stereo view generation~\cite{qiao2025genstereo}: a reference U-Net encodes the source view into semantically rich features, while a denoising U-Net synthesizes the target view conditioned on these features, providing a natural trade-off between semantic preservation and geometric adaptation~\cite{seo2024genwarp,hu2024animateanyone}. Both U-Nets are initialized from Stable Diffusion checkpoints~\cite{rombach2021highresolution}, transferring strong semantic and structural priors from large-scale image pretraining to our stereo setting. Viewpoint information $\Phi$ is injected via pixel-wise Plücker rays~\cite{plucker1865} computed for both source and target images in the canonical StereoSpace. This dense per-pixel pose representation is injected via Adaptive Layer Normalization into the ResNet blocks of both U-Nets~\cite{zheng2024free3d}, and additionally concatenated to the input latents~\cite{gao2024cat3d}, so that the diffusion process can attend directly to the underlying 3D ray configuration. The complete architecture is depicted in Fig.~\ref{fig:stereospace_architecture}.

\subsection{Warping Loss}\label{sec:warp}

DDIM~\cite{songdenoising} sampling admits a closed-form expression for the clean sample at $t=0$, which we interpret as the predicted target-view image $\widehat{I}_t$. We first supervise $\widehat{I}_t$ with a photometric loss that combines structural similarity and per-pixel~$\ell_1$:
\begin{equation}
\mathcal{L}_{\mathrm{pix}} = \alpha\bigl(1 - \mathrm{SSIM}(\widehat{I}_t, I_t)\bigr) + (1-\alpha)\,\|\widehat{I}_t - I_t\|_{1}.
\end{equation}
Unlike prior warping-based approaches, our model is not conditioned on disparity. Hence, the ground truth source-view disparity $d_s$ is used purely as a supervision, injecting explicit scene geometry into learning. We define a differentiable backward-warping operator $\mathcal{W}_{d_s}$ that maps $\widehat{I}_t$ into the source frame using $d_s$ and known camera geometry. A binary validity mask $M$ removes pixels that are out-of-bounds or fail a left–right consistency check, thus avoiding penalties on occlusions and invalid regions. The warp-consistency loss is then given by the masked $\ell_1$ residual:
\begin{equation}
    \mathcal{L}_{\mathrm{warp}} = \frac{1}{\|M\|_1 }\, \big\|  M \odot \big(\mathcal{W}_{d_s}(\widehat{I}_t) - I_s\big) \big\|_{1},
\end{equation}
where $\odot$ denotes element-wise multiplication and the norms are taken over spatial dimensions. Our final training objective combines the velocity, photometric, and warp-consistency terms:
\begin{equation}\label{eq:total_loss}
    \mathcal{L}_{\mathrm{total}} = \mathcal{L}_{\mathrm{vel}} + \lambda_{\mathrm{pix}} \mathcal{L}_{\mathrm{pix}} + \lambda_{\mathrm{warp}}\,\mathcal{L}_{\mathrm{warp}}.
\end{equation}

\begin{figure*}[t]
  \centering
  \renewcommand{\tabcolsep}{2pt}
  \begin{tabular}{cc}
       \includegraphics[width=0.5\linewidth]{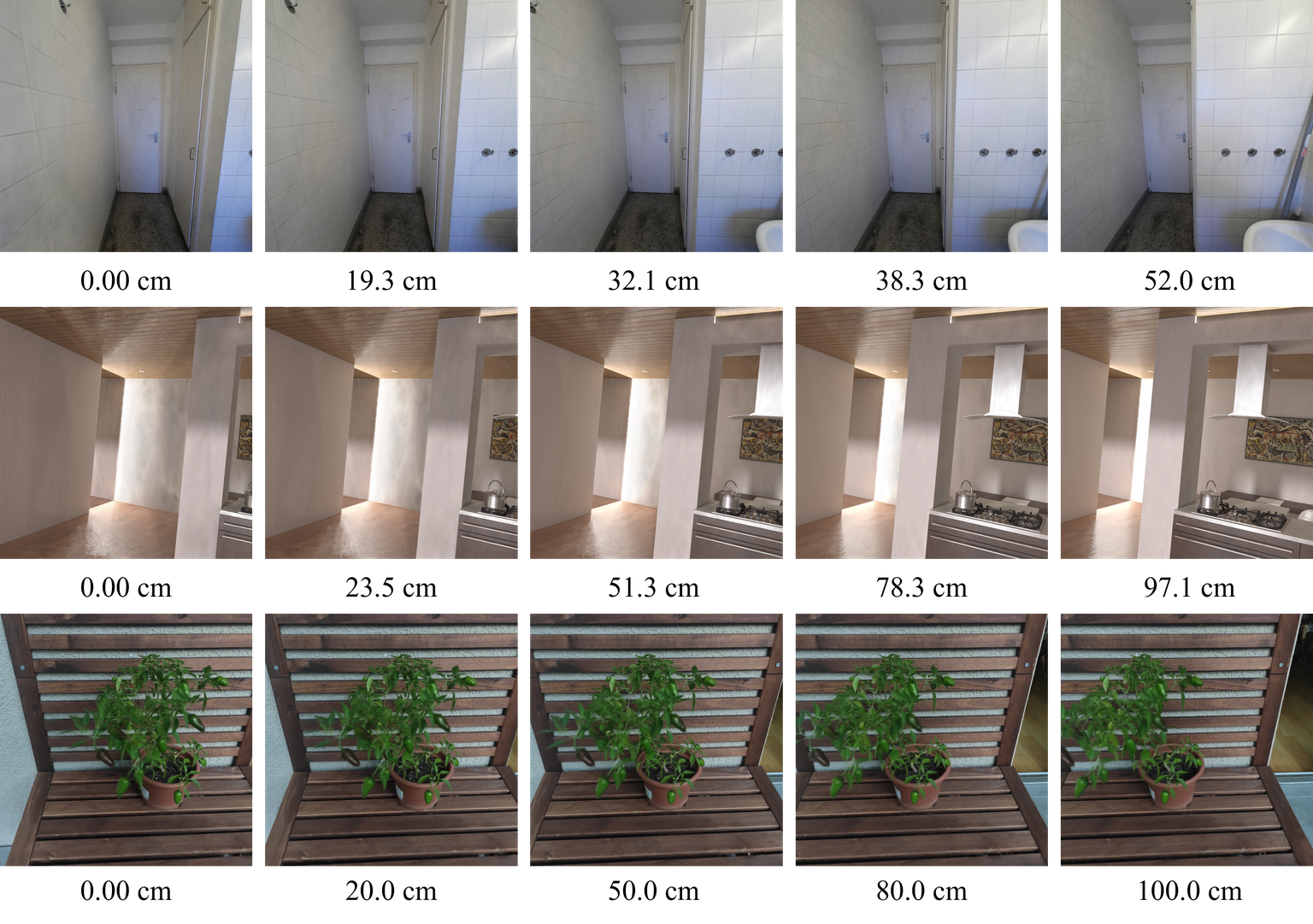} &
       \includegraphics[width=0.5\linewidth]{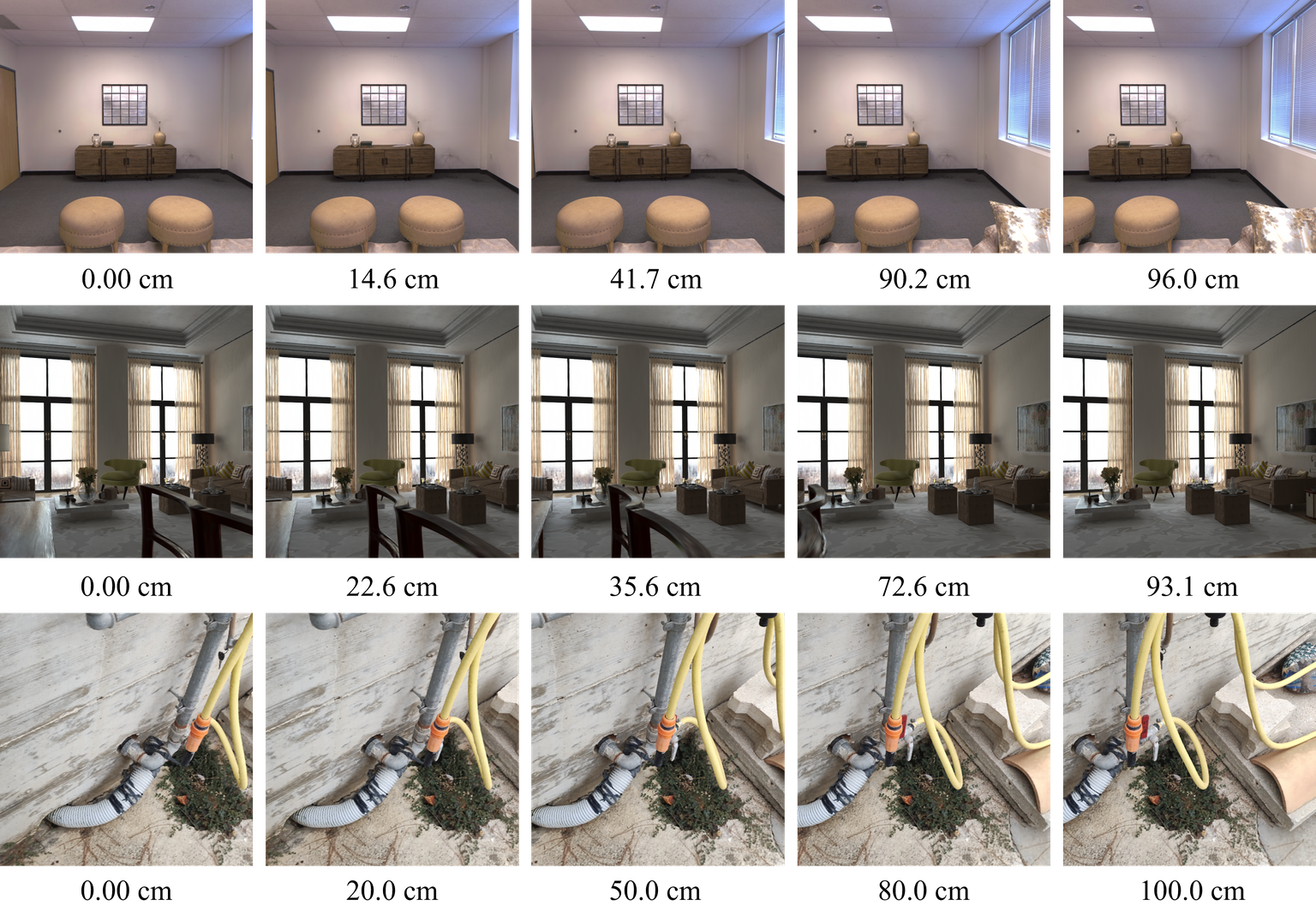} \\
  \end{tabular}\vspace{-0.4cm}
  \caption{\textbf{Multi-baseline training data.} Samples from SceneSplat-7K (top two rows) and NeRF-Stereo (bottom row). Scenes span indoor and outdoor environments and are observed under multiple, controlled baselines, providing explicit cues on how stereo geometry changes with baseline length; numbers below each view indicate the distance (cm) to the left-most view in the corresponding tuple.} 
  \label{fig:multi-baseline}\vspace{-0.3cm}
\end{figure*}

\section{Experimental Results}
\label{cpt4}

We now introduce our experimental evaluation.

\subsection{Training Datasets} 
\label{sec:trainingdatasets}

We train StereoSpace on a mixed-dataset strategy combining $\sim$750K single-baseline stereo pairs from 12 synthetic and photorealistic datasets. The main data sources include TartanAir ~\cite{wang2020tartanair} (306K pairs), Dynamic Replica ~\cite{karaev2023dynamic} (145K), IRS~\cite{wang2019irs} (103K), Falling Things~\cite{tremblay2018falling} (61K), and LayeredFlow~\cite{wen2024layeredflow} (31K), among others ~\cite{cabon2020virtualkitti,jing2024infinigen,jospin2022simstereo,Tosi2021unreal,mehl2023spring,tokarsky2024plt,buttler2012sintel}, covering indoor and outdoor scenes. Importantly, we incorporate multi-baseline data to help the model understand baseline effects, a key distinction from prior work. Specifically, we use 27K multi-view tuples from NeRF-Stereo~\cite{Tosi2023Nerf} and 5K from SceneSplat-7K~\cite{li2025scenesplat} (Hypersim~\cite{hypersim}, Replica~\cite{replica19arxiv}, ScanNet++~\cite{yeshwanthliu2023scannetpp} subsets). For the Gaussian Splats provided in SceneSplat-7K, we recover the training cameras, constrain virtual viewpoints to their geometric hull, and cluster cameras by orientation to obtain locally consistent epipolar groups from which we render short virtual-baseline stacks along stereo directions. Examples are shown in Fig.~\ref{fig:multi-baseline}. Each tuple contains 5-7 rectified views along a shared baseline direction, generating 10-21 stereo pairs per tuple. 
Smaller datasets are resampled to $10\%$ of the largest dataset size for balance. Multi-baseline tuples are weighted by $10{\times}\#\text{tuples}$ to account for multiple pairs per tuple.

\subsection{Implementation Details}
\textbf{Architecture.} For our proposed StereoSpace architecture, we fine-tune both streams of the dual U-Net on top of a Stable Diffusion 2.0 checkpoint. The first convolution of each branch is modified to accept 10 channels: 4-channel VAE latent concatenated with 6D Plücker ray embeddings. New channels are zero-initialized to preserve pretrained weights. All parameters are trainable except the highest-resolution up-block of the reference U-Net, frozen to stabilize appearance features. Semantic conditioning uses a frozen CLIP ViT-H/14 (LAION-5B~\cite{schuhmann2022laionb}) providing 1024-dim features matching the SD 2.0's cross-attention dimension. 

\noindent\textbf{Training and Inference.} We train on mixed synthetic and photorealistic stereo data (Sec.~\ref{sec:trainingdatasets}). Images are resized to $768 \times 768$ by short-side scaling and center cropping to preserve the rectified geometry. Paired crops are generated on-the-fly for consistent left-right transforms. We use AdamW \cite{loshchilov2017adamw} with learning rate $1{\times}10^{-5}$, $(\beta_1, \beta_2){=}(0.9, 0.999)$, weight decay=$1{\times}10^{-2}$, $\epsilon{=}1{\times}10^{-8}$, constant schedule, and gradient clipping at 1.0.  For the total loss in Eq.~\ref{eq:total_loss}, we set the loss weights to $\lambda_{\mathrm{pix}} = 1.0$ and $\lambda_{\mathrm{warp}} = 0.3$. Per-device batch size is 1 with gradient accumulation 6, yielding an effective batch size of $6\times N_{\text{GPU}}$, with $N_{\text{GPU}}=12$. We employ the \texttt{diffusers}~\cite{diffusers} DDIM scheduler with velocity parameterization, enabling a deterministic mapping and closed-form solution at (t{=}0)
for pixel-space reconstruction. The scheduler uses 1000 steps with a scaled-linear ($\beta$) schedule, zero-SNR ($\gamma_{\text{SNR}}=5.0$) and a $0.05$ noise offset. Classifier-free guidance is applied on CLIP and reference U-Net features with 0.1 unconditional drop ratio. Inference uses 50 steps and guidance scale of 1.5.

\subsection{Setup}
\textbf{Evaluation Protocol.} Assessing geometric consistency in stereo generation remains challenging: most pipelines are still evaluated solely with photometric metrics (PSNR, SSIM, LPIPS), which often assign the highest scores to over-smoothed, averaged predictions that visibly wash out high-frequency detail and depth edges~\cite{yang2024viewfusion, asim2025met3r}.
Recent work has sought to standardize these metrics in the context of tiny-baseline stereo~\cite{yu2025mono2stereo}, but this evaluation protocol remains tailored to a setting that is markedly different from ours. Moreover, these metrics are not expressive for ranking perceptual realism, as they penalize pixel misalignment between generated and ground-truth images more heavily than blur or unrealistic hallucinations.

We therefore propose an evaluation protocol that combines MEt3R~\cite{asim2025met3r} and iSQoE~\cite{tamir2025isqoe}, two complementary metrics that operate on orthogonal axes: MEt3R measures stereo consistency by lifting per-pixel semantic features (DINO+FeatUp~\cite{caron2021dino,fu2024featup}) into a common 3D frame predicted by a pretrained geometry model (MASt3R~\cite{leroy2024grounding}), reprojecting them into both views, and computing symmetric cosine-similarity maps $S$:
\begin{equation}
    \mathrm{MEt3R}(\mathbf{I}_1,\mathbf{I}_2)=1-\tfrac{1}{2}\big(S(\mathbf{I}_1,\mathbf{I}_2)+S(\mathbf{I}_2,\mathbf{I}_1)\big).
\end{equation}
The perceptual estimate of stereo fidelity and viewing comfort is provided with iSQoE~\cite{tamir2025isqoe}, a learned stereoscopic QoE predictor that maps a left–right pair to a single scalar trained on VR preference data.
In Sec.~\ref{sec:res_single} and~\ref{sec:res_multi}, we demonstrate that iSQoE and MEt3R provide more informative assessments than PSNR and SSIM, both qualitatively and quantitatively.

Furthermore, previous evaluation protocols~\cite{wang2024stereodiffusion, qiao2025genstereo} for stereo generation suffer from test-time leakage, as they condition inference on ground truth disparity, and thus remove the need to predict geometry. We instead evaluate all methods strictly end-to-end, without access to ground truth geometry, and account for monocular scale ambiguity via per-scene calibration: for each method and scene, we choose the camera baseline (ours) or depth-to-disparity scale that best aligns the generated stereo pair with the ground truth, using a coarse-to-fine search that minimizes the RMSE between SGBM~\cite{hirschmuller2008sgbm} disparity maps on real versus synthesized views over jointly valid pixels. The selected scale is then fixed for all metrics reported, equalizing difficulty across methods while avoiding any target-side information during generation.

\noindent\textbf{Baselines.} We compare against state-of-the-art monocular-to-stereo view generators: ZeroStereo's Stereo-Gen~\cite{wang2025zerostereo} inpainting module (warp-and-inpaint, DAv2~\cite{yang2024depth} backbone), StereoDiffusion~\cite{wang2024stereodiffusion} (latent warping), and GenStereo~\cite{qiao2025genstereo} (warped conditioning). Within our end-to-end evaluation pipeline, we perform the coarse-to-fine search over the baseline and scale parameter in $[0.025, 1]$, respectively. To probe the behavior of large open-world generative models on this task, we additionally evaluate Lyra~\cite{bahmani2025lyra}, where we approximate the right view by simulating a small camera translation along its $x$-axis and selecting the frame that minimizes the same RMSE criterion used for scale calibration.

\noindent\textbf{Datasets.}
We perform architectural ablations on Middlebury 2014~\cite{scharstein2014middlebury} and evaluate the final model on DrivingStereo~\cite{yang2019driving}, which we select over KITTI~\cite{Menze_2015_CVPR} due to its more favorable aspect ratio, as well as on Booster~\cite{ramirez2022booster} and LayeredFlow~\cite{wen2024layeredflow}.
This suite collectively balances canonical stereo evaluation with diverse, real-world conditions (multi-layered depth, weather changes, non-Lambertian surfaces). For all datasets, we rescale the shorter image side to each model's native input resolution and then apply a square center crop to match the required aspect ratio while preserving stereo geometry. To ensure a fair comparison, all generated stereo pairs are finally downsampled to a common resolution of $512{\times}512$ before computing metrics. \medskip

\subsection{Ablation Study}\label{sec:ablation}

Before comparing with state-of-the-art models, we study the impact of the different components in StereoSpace.

\noindent\textbf{Viewpoint Conditioning.} Our dual U-Net architecture exposes multiple injection points for viewpoint conditioning. We compare three alternatives on Middlebury (Tab.~\ref{tab:ablation}): (i) a CLIP~\cite{radford2021clip} text embedding that encodes the desired stereo baseline as a prompt (``Stereo baseline $x$\,cm to the left/right'') into the denoising U-Net, (ii)~PRoPE-style~\cite{li2025prope} projective attention that injects full camera frustums into the cross-attention between the two U-Nets, and (iii) a dense Plücker ray embedding of the target camera,. Each conditioning alone already surpasses GenStereo on both iSQoE and MEt3R, indicating that our viewpoint-conditioned diffusion framework is effective. Adding PRoPE on top of Plücker does not yield further improvements, indicating that stacking multiple conditioning signals is not beneficial in this regime. Plücker conditioning achieves the best scores and is used as our default, it further exposes camera intrinsics explicitly and avoids an additional text encoder.

\begin{table}
\renewcommand{\tabcolsep}{10pt}
\resizebox{0.48\textwidth}{!}{
    \begin{tabular}{@{}l
                  S[table-format=1.4] S[table-format=1.4] S[table-format=1.4]
                  S[table-format=1.4] S[table-format=1.4] S[table-format=1.4]
                  @{}}
    \toprule
    & \multicolumn{2}{c}{\textbf{Middlebury 2014 \cite{scharstein2014middlebury}}} \\
    \cmidrule(lr){2-3}
    \textbf{Method} & \multicolumn{1}{c}{\mdown{iSQoE}} &
      \multicolumn{1}{c}{\mdown{MEt3R}} \\
    \midrule
    GenStereo \cite{qiao2025genstereo} & 0.6933 & 0.1339 \\
    \hline
    StereoSpace w/ text & 0.6841 & \snd 0.0907 \\
    StereoSpace w/ Plücker & \fst \bf 0.6823 & \fst \bf 0.0901 \\
    StereoSpace w/ PRoPe & 0.6865 & 0.0937 \\
    StereoSpace w/ Plücker+PRoPE & \snd 0.6828 & 0.0945  \\
    \hline
    StereoSpace w/ Plücker & \fst \bf 0.6823 & \snd 0.0901 \\
    StereoSpace wo/ multi-baseline & 0.6907 & 0.1095 \\
    StereoSpace w/ warping loss & \snd 0.6829 & \fst \bf 0.0893 \\
    \bottomrule
  \end{tabular}}\vspace{-0.3cm}
  \caption{\textbf{Ablation study on Middlebury dataset.} Metrics are iSQoE~($\downarrow$) and MEt3R~($\downarrow$). \colorbox{firstcolor}{\textbf{Best}} and \colorbox{secondcolor}{second-best} scores highlighted over subcategories.} \label{tab:ablation}\vspace{-0.3cm}
\end{table}

\noindent\textbf{Training Data and Disparity Loss.} To assess the impact of additional multi-view stereo data (NeRF-Stereo, SceneSplat), we train a variant of our model without these sources, using effectively the same training corpus as GenStereo. Performance drops noticeably on both iSQoE and MEt3R compared to our full dataset, confirming the benefit of the extra multi-view supervision, yet this reduced-data variant still outperforms GenStereo, supporting the robustness of our viewpoint conditioning design. Finally, we enable an auxiliary disparity loss during training. This variant achieves a modest improvement in MEt3R at the cost of a slight degradation in iSQoE, consistent with our hypothesis that disparity supervision encourages stronger geometric alignment while mildly trading off perceived viewing comfort.

\begin{table*}[t]
  \centering
  \resizebox{\textwidth}{!}{
  \setlength{\tabcolsep}{15pt}
  \begin{tabular}{@{}lll
                  S[table-format=1.4] S[table-format=1.4] S[table-format=1.4]
                  S[table-format=1.4] S[table-format=1.4] S[table-format=1.4]
                  @{}}
    \toprule
    & & & \multicolumn{2}{c}{\textbf{Middlebury 2014 \cite{scharstein2014middlebury}}} & \multicolumn{2}{c}{\textbf{Drivingstereo \cite{yang2019driving}}} \\
    \cmidrule(lr){4-5}\cmidrule(lr){6-7}
    \textbf{Method} & 
    \textbf{Category} & \textbf{Depth} & \multicolumn{1}{c}{\mdown{iSQoE}} &
      \multicolumn{1}{c}{\mdown{MEt3R}} &
      \multicolumn{1}{c}{\mdown{iSQoE}} &
      \multicolumn{1}{c}{\mdown{MEt3R}} \\
    \midrule
    StereoDiffusion \cite{wang2024stereodiffusion} & 
    \textit{Warp-and-inpaint} & DAv2 \cite{yang2024depth} & 0.7475 & 0.1933 & \trd 0.7887 & 0.1015 \\
    ZeroStereo \cite{wang2025zerostereo} & 
    \textit{Latent warping} & DAv2 \cite{yang2024depth} & 0.7423 & 0.2057 & 0.7964 & \trd 0.0798 \\
    GenStereo \cite{qiao2025genstereo} & 
    \textit{Warped conditioning} & DAv2 \cite{yang2024depth} & \snd 0.6933 & \trd 0.1339 &  \snd 0.7850 & \snd 0.0728 \\
    Lyra \cite{bahmani2025lyra} & \textit{3DGS generative model}
    & MoGe-2 \cite{wang2025moge} & \trd 0.7184 & \snd 0.1163 &  0.7891 & 0.0949 \\
    \midrule
    \bf StereoSpace (ours) & \textit{Depth-free} & - & \fst \bf0.6829 & \fst \bf 0.0893 &  \fst \bf \fst \bf0.7829 & \fst \bf0.0717 \\
    \bottomrule
  \end{tabular}}\vspace{-0.3cm}
  \caption{\textbf{Results on Middlebury and Drivingstereo.} Metrics are iSQoE~($\downarrow$) and MEt3R~($\downarrow$). For each model, we report its category, followed by the depth estimator it uses at inference time.
  \colorbox{firstcolor}{\textbf{Best}}, \colorbox{secondcolor}{second-} and \colorbox{thirdcolor}{third-best} scores highlighted. 
  }
  \label{tab:eval_middlebury_drivingstereo}
\end{table*}

\begin{table*}[t]
  \centering
  \resizebox{\textwidth}{!}{
  \setlength{\tabcolsep}{15pt}
  \begin{tabular}{@{}lll
                  S[table-format=1.4] S[table-format=1.4] S[table-format=1.4]
                  S[table-format=1.4] S[table-format=1.4] S[table-format=1.4]
                  @{}}
    \toprule
    & & & \multicolumn{2}{c}{\textbf{Booster \cite{ramirez2022booster}}} & \multicolumn{2}{c}{\textbf{LayeredFlow \cite{wen2024layeredflow}}} \\
    \cmidrule(lr){4-5}\cmidrule(lr){6-7}
    \textbf{Method} & 
    \textbf{Category} & \textbf{Depth} & \multicolumn{1}{c}{\mdown{iSQoE}} &
      \multicolumn{1}{c}{\mdown{MEt3R}}  &
      \multicolumn{1}{c}{\mdown{iSQoE}} &
      \multicolumn{1}{c}{\mdown{MEt3R}} \\
    \midrule
    StereoDiffusion \cite{wang2024stereodiffusion} & 
    \textit{Warp-and-inpaint} & DAv2 \cite{yang2024depth} & 0.7248 & 0.2011 & 0.8046 & 0.3074 \\
    ZeroStereo \cite{wang2025zerostereo} & 
    \textit{Latent warping} & DAv2 \cite{yang2024depth} & 0.7503 & 0.3171 & 0.8108 & 0.3630 \\
    GenStereo \cite{qiao2025genstereo} & 
    \textit{Warped conditioning} & DAv2 \cite{yang2024depth} & \snd 0.6901 & \trd 0.1457 & \snd 0.7678 & \trd 0.2275 \\
    Lyra \cite{bahmani2025lyra} & 
    \textit{3DGS generative model} & MoGe-2 \cite{wang2025moge} & \trd 0.6989 & \snd 0.1293 & \trd 0.7802 & \snd 0.1877 \\
    \midrule
    \bf StereoSpace (ours) & \textit{Depth-free} & - & \fst \bf 0.6764 & \fst \bf 0.1013 & \fst \bf 0.7489 & \fst \bf 0.1619 \\
    \bottomrule
  \end{tabular}}\vspace{-0.3cm}
  \caption{\textbf{Results on Booster and LayeredFlow.} Metrics are iSQoE~($\downarrow$) and Met3r~($\downarrow$). \colorbox{firstcolor}{\textbf{Best}}, \colorbox{secondcolor}{second-} and \colorbox{thirdcolor}{third-best} scores highlighted.
  For each model, we report the category it belongs to, followed by the depth model it uses at inference time. 
  }
  \label{tab:eval_booster_layeredflow}
\end{table*}

\subsection{Results on Single-Layer Geometries}\label{sec:res_single}

We first compare with state-of-the-art methods on two datasets: Middlebury 2014~\cite{scharstein2014middlebury} (indoor) and DrivingStereo~\cite{yang2019driving} (outdoor).
Table \ref{tab:eval_middlebury_drivingstereo} collects both iSQoE and MEt3R metrics achieved by the four baselines and our StereoSpace framework on the two datasets.

On Middlebury 2014, StereoSpace achieves the lowest iSQoE (0.6829) and MEt3R (0.0893), improving upon the scores of  GenStereo~\cite{qiao2025genstereo} and Lyra~\cite{bahmani2025lyra} in the latter metric by more than 30\% and 20\%, respectively. ZeroStereo~\cite{wang2025zerostereo} and StereoDiffusion~\cite{wang2024stereodiffusion} trail on this benchmark, with the highest errors in both metrics. For DrivingStereo, our method again yields the best iSQoE and attains a MEt3R score of 0.0717, while ZeroStereo is competitive in geometry (better than StereoDiffusion), but remains behind GenStereo (0.0728) and ours. We can notice how the margins on this dataset are lower due to the simpler geometry characteristic of driving environments.

These trends align with the qualitative behavior of the baselines. StereoDiffusion often produces overly smooth, monochrome completions: such artifacts can inflate pixel similarity measures yet harm stereo viewing comfort. This is reflected by poorer iSQoE and notably higher MEt3R scores, especially on Middlebury. ZeroStereo preserves warped structure where visible, but its inpainted regions are frequently inconsistent, which elevates iSQoE and, on Middlebury, also degrades MEt3R. It fares better on DrivingStereo where geometry is simpler, yet still lags behind GenStereo and StereoSpace. GenStereo’s warped conditioning is less susceptible to these issues: benefiting from a strong monocular depth prior, it provides accurate disparity guidance in latent space and achieves the best DrivingStereo MEt3R among the baselines. Nevertheless, our end-to-end formulation surpasses GenStereo in overall quality, delivering the strongest iSQoE on both datasets, as well as the best geometric consistency. 
Notably, Lyra mostly competes with GenStereo for the second place on Middlebury, probably a consequence of its open-world capabilities avoiding specialization on the considered stereo setting.
These results suggest that incorporating geometry directly into the generative model yields stereo pairs that are both perceptually appealing and geometrically accurate in typical settings.

\begin{figure}[t]
  \centering
  \begin{overpic}
      [clip,trim=0cm 5.2cm 0cm 0cm, width=\linewidth]{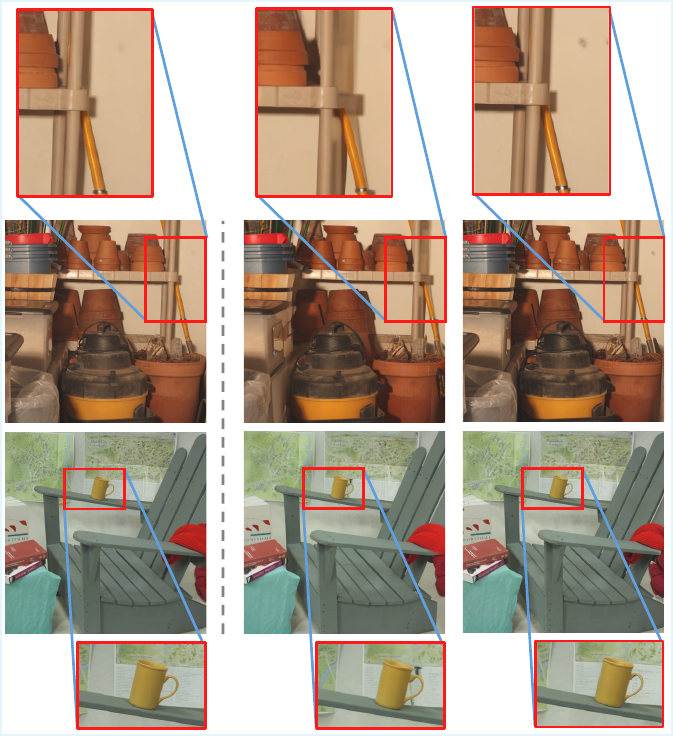}
      \put (2,65) {Ground truth}      \put (35,65) {GenStereo \cite{qiao2025genstereo}}
      \put (67,65) {\textbf{StereoSpace (ours)}}
      \put (37,5) {\tiny \textcolor{yellow}{\textbf{PSNR = 18.3323}}}
      \put (37,3) {\tiny \textcolor{yellow}{\textbf{SSIM = 0.6061}}}
      \put (70,5) {\tiny \textcolor{yellow}{\textbf{PSNR = 14.3611}}}
      \put (70,3) {\tiny \textcolor{yellow}{\textbf{SSIM = 0.4211}}}
  \end{overpic}\vspace{-0.3cm}
  \caption{
  \textbf{Qualitative results on Middlebury 2014 \cite{scharstein2014middlebury}.} 
  Compared to GenStereo (left), StereoSpace (right) preserves realistic image details, as shown on top.  
  We also report PSNR and SSIM to highlight the inability of such metrics to account for it.
  }\vspace{-0.3cm}
  \label{fig:qualitative_middlebury}
\end{figure} 

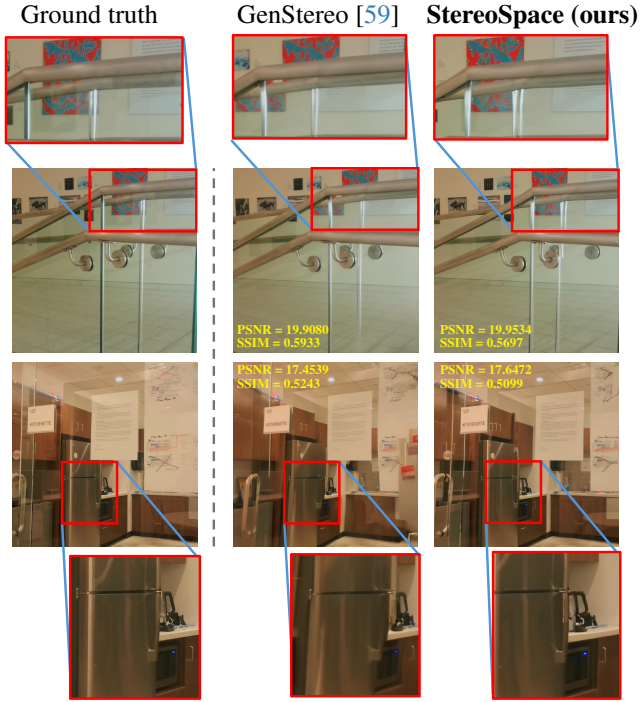
\begin{figure}[t]
  \centering
  \begin{overpic}
      [clip,trim=0cm 6.5cm 0cm 0cm,width=\linewidth]{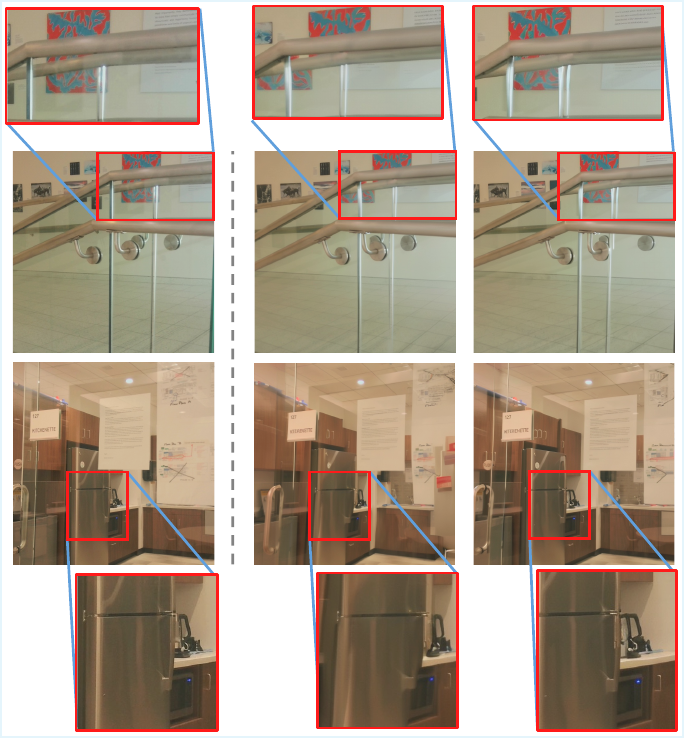}
      \put (2,53) {Ground truth}      \put (35,53) {GenStereo \cite{qiao2025genstereo}}
      \put (67,53) {\textbf{StereoSpace (ours)}}
      \put (38,4) {\tiny \textcolor{yellow}{\textbf{PSNR = 19.9080}}}
      \put (38,2) {\tiny \textcolor{yellow}{\textbf{SSIM = 0.5933}}}
      \put (70,4) {\tiny \textcolor{yellow}{\textbf{PSNR = 19.9534}}}
      \put (70,2) {\tiny \textcolor{yellow}{\textbf{SSIM = 0.5697}}}
  \end{overpic}\vspace{-0.3cm}
  \caption{
  \textbf{Qualitative results on LayeredFlow \cite{wen2024layeredflow}.}
  While GenStereo (left) fails at modeling layered structures in the presence of transparent surfaces, StereoSpace (right) excels at this.
  }\vspace{-0.3cm}
  \label{fig:qualitative_layered}
\end{figure}

\begin{figure*}[t]
    \centering
    \setlength{\tabcolsep}{1pt}
    \begin{tabular}{cc}
        \multicolumn{2}{c}{\includegraphics[width=0.8\linewidth, clip, trim=0.5cm 0cm 1cm 0.75cm]{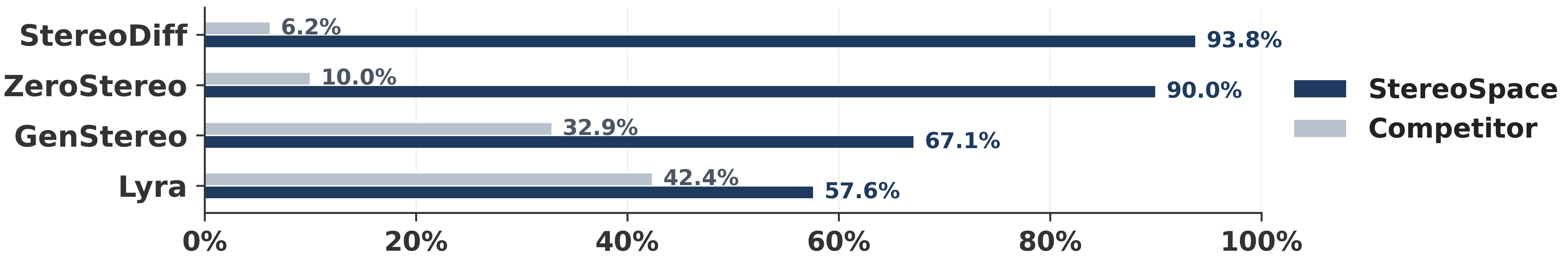}} \vspace{-0.2cm}\\
        \multicolumn{2}{c}{\textbf{(a)}} \\
        \includegraphics[height=0.24\linewidth, clip, trim=0cm 0cm 0.7cm 0.75cm]{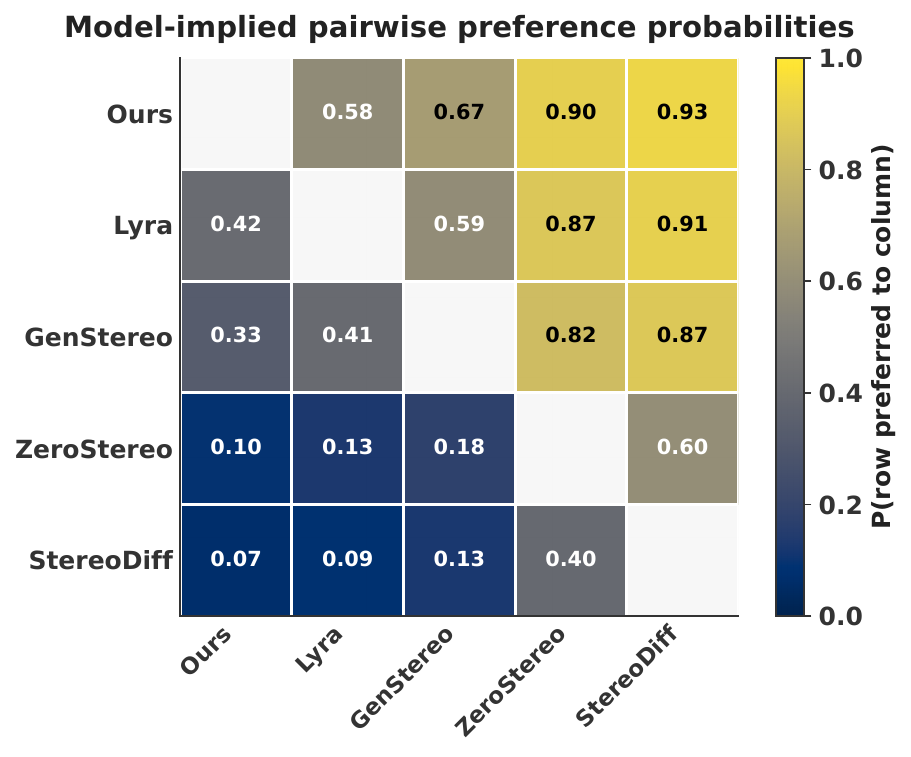} & \includegraphics[height=0.24\linewidth, clip, trim=0cm 0cm 0cm 0.75cm]{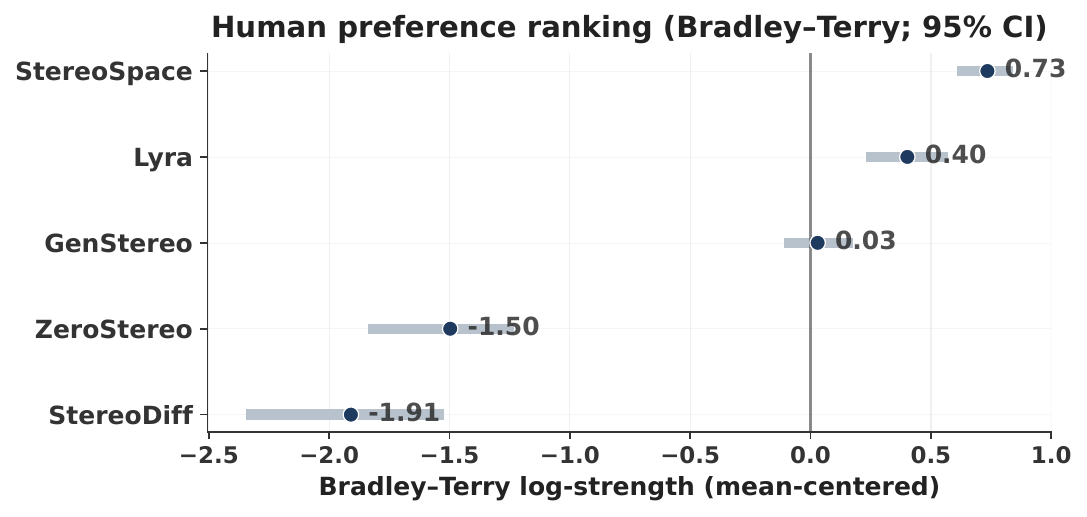} \vspace{-0.2cm}\\
        {\textbf{(b)}} & {\textbf{(c)}} \vspace{-0.1cm}\\
    \end{tabular}
    \vspace{-0.3cm}
    \caption{\textbf{User study through pairwise comparisons.} 
    70 participants were involved, for a total of 1.4K pairwise comparisons. (a) StereoSpace winrate against other methods; (b) Bradley-Terry win probabilities, (c) ranking according to Bradley-Terry scores. }
    \vspace{-0.3cm}
    \label{fig:plot}
\end{figure*}

\subsection{Results on Multi-Layer Geometries}\label{sec:res_multi}

As a final comparison, we evaluate against the same baselines on Booster~\cite{ramirez2022booster} and LayeredFlow~\cite{wen2024layeredflow}, showcasing complex geometry and transparency-induced depth layers.
Table \ref{tab:eval_booster_layeredflow} collects both iSQoE and MEt3R metrics achieved by the four baselines and StereoSpace on the two datasets. 

On these more challenging benchmarks, all methods exhibit degradation consistent with multi-level depth, transparency, and irregular reflectance. These phenomena violate the single-surface assumption inherent in warping pipelines. A single disparity map cannot model semi-transparent layers, specular highlights, or view-dependent reflections simultaneously, leading to geometric drift and compensatory blur. As a consequence, StereoSpace attains the best scores on both metrics across both datasets by a large margin. ZeroStereo is the most affected, with its reliance on explicit warping and impainting yielding the highest errors. While StereoDiffusion performs slightly better but still suffers from oversmoothing, GenStereo remains competitive, but trails our method on both iSQoE and MEt3R, mostly competing with Lyra for the second place.

This confirms that embedding stereo reasoning into the generator confers robustness to non-ideal optics and layered geometry, which are ubiquitous outside controlled settings.

\subsection{Qualitative Results}

Figure \ref{fig:qualitative_middlebury} shows qualitative samples from \texttt{Storage} scene, reporting the results rendered by GenStereo and StereoSpace. Our method accurately deals with the complex overlaying in the background, faithfully recreating the shadows effect and realistic occlusions between objects. In contrast, GenStereo produces unrealistic overlaps between foreground and background objects, mainly trying to inpaint the occlusions caused by the warping process. 

Figure \ref{fig:qualitative_layered} shows qualitative samples from the LayeredFlow dataset, which depict in particular a glass railing together with the results rendered by GenStereo and StereoSpace. 
It is clearly evident how the reliance on depth priors and warping makes GenStereo unable to handle the two different layers in the scene, i.e., the glass railing itself and the wall in the background visible through it. Indeed, we can observe how the painting on the wall is split in the rendered image, with its bottom part being moved further to the left as if it belongs to the first depth layer. In contrast, StereoSpace is not affected by this issue and can render the painting faithfully. 
We refer the reader to the supplementary material for higher-resolution qualitative results.

\noindent\textbf{The Ineffectiveness of Conventional Metrics.} Both Figures \ref{fig:qualitative_middlebury} and \ref{fig:qualitative_layered} also reports PSNR and SSIM computed on images generated by GenStereo and StereoSpace with respect to the ground truth right image -- an exhaustive evaluation is reported in the supplementary material. We can observe that, in the former case, GenStereo achieves better scores than our method, despite the evident artifacts and its inability to properly handle complex geometries and shadows.
We attribute this behavior to the tighter pixel-wise alignment between rendered and ground-truth views achieved by depth-warping methods. As a result, these metrics tend to penalize our approach, which does not rely on depth priors, yet produces more realistic and geometry-consistent images, as reflected by iSQoE and MEt3R, which we consider more appropriate for benchmarking this task.

\subsection{User Study Evaluation}

{To confirm the effectiveness of the proposed evaluation protocol, we conduct a study among 70 participants instructed to vote for the best generated image in pairwise comparisons. 
Each subject received 20 random pairs of different methods' output images alongside the corresponding left image as a reference. We report the outcome in Figure~\ref{fig:plot}, showing StereoSpace winrate against any other method (a), as well as aggregated scores using the Bradley–Terry preference model (b), used then to rank methods (c). These results align well with the trend highlighted by MEt3R and iSQoE: we outperform Lyra about 60-40, while others lag behind.
}

\section{Conclusion}\label{cpt6}

In this paper, we presented StereoSpace, an end-to-end approach to stereo view synthesis. While existing solution to this task \cite{wang2025zerostereo,wang2024stereodiffusion,qiao2025genstereo,bahmani2025lyra} heavily rely on depth priors to exploit warping and turn the generative task into an inpainting problem, our framework removes this requirement and only relies on view conditioning to effectively learn stereo geometry. StereoSpace generates more realistic novel views, in particular when running on images depicting multi-layered geometry, as confirmed by both perceptual and geometry based metrics. Future developments will aim at extending StereoSpace to stereo video generation. 

\noindent\textbf{Acknowledgments.} We acknowledge the EuroHPC Joint Undertaking for awarding us access to Leonardo at CINECA, Italy, under Proposal ID EHPC-DEV-2025D05-054.
The authors also acknowledge the CINECA award under the ISCRA initiative.

{
    \small
    \bibliographystyle{ieeenat_fullname}
    \bibliography{main}

@String(CVPR= {IEEE/CVF Conference on Computer Vision and Pattern Recogition (CVPR)})

@String(ICCV= {IEEE/CVF International Conference on Computer Vision (ICCV)})

@String(ECCV= {European Conference on Computer Vision (ECCV)})

@String(NeurIPS= {Advances in Neural Information Processing Systems (NeurIPS)})

@String(CVPRW= {IEEE/CVF Conference on Computer Vision and Pattern Recogition Workshops (CVPRW)})

@String(ICML= {International Conference on Machine Learning (ICML)})

@String(TOG= {ACM Trans. Graph.})

@String(IROS = {IEE/RSJ International Conference on Intelligent Robotics and Systems})

@String(ICLR = {International Conference on Learning Representations (ICLR)})

@String(AAAI = {AAAI})

@inproceedings{yu2025mono2stereo,
  title     = {{Mono2Stereo}: A Benchmark and Empirical Study for Stereo Conversion},
  author    = {Songsong Yu and Yuxin Chen and Zhongang Qi and Zeke Xie and Yifan Wangand Lijun Wang and Ying Shan and Huchuan Lu},
  booktitle = CVPR,
  year      = {2025}
}

@inproceedings{qiao2025genstereo,
  title={{GenStereo}: Towards Open-World Generation of Stereo Images and Unsupervised Matching},
  author={Qiao, Feng and Xiong, Zhexiao and Xing, Eric and Jacobs, Nathan},
  booktitle=CVPR,
  year={2025}
}

@inproceedings{wang2024stereodiffusion,
  title={{StereoDiffusion}: Training-Free Stereo Image Generation Using Latent Diffusion Models},
  author={Wang, Lezhong and Frisvad, Jeppe Revall and Jensen, Mark Bo and Bigdeli, Siavash Arjomand},
  booktitle=CVPR,
  year={2024}
}

@article{loshchilov2017adamw,
  title={Decoupled weight decay regularization},
  author={Loshchilov, Ilya and Hutter, Frank},
  journal={preprint arXiv:1711.05101},
  year={2017}
}

@inproceedings{salimans2022progressive,
  title={Progressive Distillation for Fast Sampling of Diffusion Models},
  author={Salimans, Tim and Ho, Jonathan},
  booktitle=ICLR,
  year={2022}
}

@inproceedings{karras2022elucidating,
  title={Elucidating the Design Space of Diffusion-Based Generative Models},
  author={Karras, Tero and Aittala, Miika and Aila, Timo and Laine, Samuli},
  booktitle=NeurIPS,
  year={2022}
}

@inproceedings{ke2024repurposing,
  title={Repurposing diffusion-based image generators for monocular depth estimation},
  author={Ke, Bingxin and Obukhov, Anton and Huang, Shengyu and Metzger, Nando and Daudt, Rodrigo Caye and Schindler, Konrad},
  booktitle=CVPR,
  year={2024}
}

@article{ke2025marigold,
  title={Marigold: Affordable Adaptation of Diffusion-Based Image Generators for Image Analysis},
  author={Ke, Bingxin and Qu, Kevin and Wang, Tianfu and Metzger, Nando and Huang, Shengyu and Li, Bo and Obukhov, Anton and Schindler, Konrad},
  journal={preprint arXiv:2505.09358},
  year={2025}
}

@InProceedings{scharstein2014middlebury,
author="Scharstein, Daniel
and Hirschm{\"u}ller, Heiko
and Kitajima, York
and Krathwohl, Greg
and Ne{\v{s}}i{\'{c}}, Nera
and Wang, Xi
and Westling, Porter",
editor="Jiang, Xiaoyi
and Hornegger, Joachim
and Koch, Reinhard",
title="High-Resolution Stereo Datasets with Subpixel-Accurate Ground Truth",
booktitle="Pattern Recognition",
year="2014",
pages="31--42"
}

@INPROCEEDINGS{yang2019driving,
  author={Yang, Guorun and Song, Xiao and Huang, Chaoqin and Deng, Zhidong and Shi, Jianping and Zhou, Bolei},
  booktitle=CVPR, 
  title={{DrivingStereo}: A Large-Scale Dataset for Stereo Matching in Autonomous Driving Scenarios}, 
  year={2019}
}

@inproceedings{zhao2024stereocrafter,
  title     = {{StereoCrafter}: Diffusion-based Generation of Long and High-fidelity Stereoscopic 3D from Monocular Videos},
  author    = {Sijie Zhao and Wenbo Hu and Xiaodong Cun and Yong Zhang and Xiaoyu Li and Zhe Kong and Xiangjun Gao and Muyao Niu and Ying Shan},
  booktitle = {preprint arXiv:2407.00367},
  year      = {2024}
}

@article{shi2024stereocrafterzero,
  title   = {{StereoCrafter-Zero}: Zero-Shot Stereo Video Generation with Noisy Restart},
  author  = {Jian Shi and Qian Wang and Zhenyu Li and Ramzi Idoughi and Peter Wonka},
  journal = {preprint arXiv:2411.14295},
  year    = {2024}
}

@inproceedings{lv2024spatialdreamer,
  title={{SpatialDreamer}: Self-supervised Stereo Video Synthesis from Monocular Input},
  author={Lv, Zhen and Long, Yangqi and Huang, Congzhentao and Li, Cao and Lv, Chengfei and Ren, Hao and Zheng, Dian},
  booktitle=CVPR,
  year={2025}
}

@inproceedings{dai2024svg,
  title={{SVG}: 3D Stereoscopic Video Generation via Denoising Frame Matrix},
  author={Peng Dai and Feitong Tan and Qiangeng Xu and David Futschik and Ruofei Du and Sean Fanello and Xiaojuan Qi and Yinda Zhang},
  booktitle=ICLR,
  year={2025}
}

@inproceedings{yang2024viewfusion,
  title={{ViewFusion}: Towards multi-view consistency via interpolated denoising},
  author={Yang, Xianghui and Zuo, Yan and Ramasinghe, Sameera and Bazzani, Loris and Avraham, Gil and van den Hengel, Anton},
  booktitle=CVPR,
  year={2024}
}

@inproceedings{asim2025met3r,
  title={{MEt3R}: Measuring multi-view consistency in generated images},
  author={Asim, Mohammad and Wewer, Christopher and Wimmer, Thomas and Schiele, Bernt and Lenssen, Jan Eric},
  booktitle=CVPR,
  year={2025}
}

@inproceedings{leroy2024grounding,
  title={Grounding image matching in 3d with mast3r},
  author={Leroy, Vincent and Cabon, Yohann and Revaud, J{\'e}r{\^o}me},
  booktitle={European Conference on Computer Vision},
  pages={71--91},
  year={2024},
  organization={Springer}
}

@inproceedings{ramirez2022booster,
  title={Open challenges in deep stereo: the booster dataset},
  author={Ramirez, Pierluigi Zama and Tosi, Fabio and Poggi, Matteo and Salti, Samuele and Mattoccia, Stefano and Di Stefano, Luigi},
  booktitle=CVPR,
  year={2022}
}

@inproceedings{schuhmann2022laionb,
  title={{LAION-5}B: An open large-scale dataset for training next generation image-text models},
  author={Christoph Schuhmann and
          Romain Beaumont and
          Richard Vencu and
          Cade W Gordon and
          Ross Wightman and
          Mehdi Cherti and
          Theo Coombes and
          Aarush Katta and
          Clayton Mullis and
          Mitchell Wortsman and
          Patrick Schramowski and
          Srivatsa R Kundurthy and
          Katherine Crowson and
          Ludwig Schmidt and
          Robert Kaczmarczyk and
          Jenia Jitsev},
  booktitle=NeurIPS,
  year={2022}
}

@inproceedings{tamir2025isqoe,
  title={What Makes for a Good Stereoscopic Image?},
  author={Tamir, Netanel and Amir, Shir and Itzhaky, Ranel and Atia, Noam and Sundaram, Shobhita and Fu, Stephanie and Sokolovsky, Ron and Isola, Phillip and Dekel, Tali and Zhang, Richard and Farber, Miriam},
  booktitle=CVPR,
  year={2025}
}

@InProceedings{Menze_2015_CVPR,
author = {Menze, Moritz and Geiger, Andreas},
title = {Object Scene Flow for Autonomous Vehicles},
booktitle = {Proceedings of the IEEE Conference on Computer Vision and Pattern Recognition (CVPR)},
month = {June},
year = {2015}
}

@article{wen2025stereo,
  title={FoundationStereo: Zero-Shot Stereo Matching},
  author={Bowen Wen and Matthew Trepte and Joseph Aribido and Jan Kautz and Orazio Gallo and Stan Birchfield},
  journal={CVPR},
  year={2025}
}

@article{replica19arxiv,
  title =   {The {R}eplica Dataset: A Digital Replica of Indoor Spaces},
  author =  {Julian Straub and Thomas Whelan and Lingni Ma and Yufan Chen and Erik Wijmans and Simon Green and Jakob J. Engel and Raul Mur-Artal and Carl Ren and Shobhit Verma and Anton Clarkson and Mingfei Yan and Brian Budge and Yajie Yan and Xiaqing Pan and June Yon and Yuyang Zou and Kimberly Leon and Nigel Carter and Jesus Briales and  Tyler Gillingham and  Elias Mueggler and Luis Pesqueira and Manolis Savva and Dhruv Batra and Hauke M. Strasdat and Renzo De Nardi and Michael Goesele and Steven Lovegrove and Richard Newcombe },
  journal = {arXiv preprint arXiv:1906.05797},
  year =    {2019}
}

@misc{diffusers,
  author = {Patrick von Platen and Suraj Patil and Anton Lozhkov and Pedro Cuenca and Nathan Lambert and Kashif Rasul and Mishig Davaadorj and Dhruv Nair and Sayak Paul and William Berman and Yiyi Xu and Steven Liu and Thomas Wolf},
  title = {Diffusers: State-of-the-art diffusion models},
  year = {2022},
  publisher = {GitHub},
  journal = {GitHub repository},
  howpublished = {\url{https://github.com/huggingface/diffusers}}
}

@inproceedings{hypersim,
title = {Hypersim: A Photorealistic Synthetic Dataset for Holistic Indoor Scene Understanding},
booktitle = {ICCV},
author = {Mike Roberts and Jason Ramapuram and Anurag Ranjan and Atulit Kumar and Miguel Angel Bautista and Nathan Paczan and Russ Webb and Joshua M. Susskind},
year = {2021},
URL = {https://arxiv.org/pdf/2011.02523.pdf}
}

@inproceedings{yeshwanthliu2023scannetpp,
  title={ScanNet++: A High-Fidelity Dataset of 3D Indoor Scenes},
  author={Yeshwanth, Chandan and Liu, Yueh-Cheng and Nie{\ss}ner, Matthias and Dai, Angela},
  booktitle = {Proceedings of the International Conference on Computer Vision ({ICCV})},
  year={2023}
}

@ARTICLE{hirschmuller2008sgbm,
  author={Hirschmuller, Heiko},
  journal={IEEE Transactions on Pattern Analysis and Machine Intelligence}, 
  title={Stereo Processing by Semiglobal Matching and Mutual Information}, 
  year={2008},
  volume={30},
  number={2},
  pages={328-341}}

@inproceedings{caron2021dino,
  title={Emerging properties in self-supervised vision transformers},
  author={Caron, Mathilde and Touvron, Hugo and Misra, Ishan and J{\'e}gou, Herv{\'e} and Mairal, Julien and Bojanowski, Piotr and Joulin, Armand},
  booktitle=ICCV,
  year={2021}
}

@article{zhou2025stable,
  title={Stable virtual camera: Generative view synthesis with diffusion models},
  author={Zhou, Jensen and Gao, Hang and Voleti, Vikram and Vasishta, Aaryaman and Yao, Chun-Han and Boss, Mark and Torr, Philip and Rupprecht, Christian and Jampani, Varun},
  journal={preprint arXiv:2503.14489},
  year={2025}
}

@inproceedings{
    fu2024featup,
    title={{FeatUp}: A Model-Agnostic Framework for Features at Any Resolution},
    author={Stephanie Fu and Mark Hamilton and Laura E. Brandt and Axel Feldmann and Zhoutong Zhang and William T. Freeman},
    booktitle=ICLR,
    year={2024}
}

@inproceedings{hu2024animateanyone,
  title={Animate anyone: Consistent and controllable image-to-video synthesis for character animation},
  author={Hu, Li},
  booktitle=CVPR,
  pages={8153--8163},
  year={2024}
}

@article{seo2024genwarp,
  title={{GenWarp}: Single image to novel views with semantic-preserving generative warping},
  author={Seo, Junyoung and Fukuda, Kazumi and Shibuya, Takashi and Narihira, Takuya and Murata, Naoki and Hu, Shoukang and Lai, Chieh-Hsin and Kim, Seungryong and Mitsufuji, Yuki},
  journal=NeurIPS,
  year={2024}
}

@inproceedings{rombach2021highresolution,
  title={High-resolution image synthesis with latent diffusion models},
  author={Rombach, Robin and Blattmann, Andreas and Lorenz, Dominik and Esser, Patrick and Ommer, Bj{\"o}rn},
  booktitle=CVPR,
  pages={10684--10695},
  year={2022}
}

@article{plucker1865,
  author    = {Julius Pl{\"u}cker},
  title     = {On a New Geometry of Space},
  journal   = {Philosophical Transactions of the Royal Society of London},
  volume    = {155},
  pages     = {725--791},
  year      = {1865}
}

@inproceedings{zheng2024free3d,
  title={{Free3D}: Consistent novel view synthesis without 3d representation},
  author={Zheng, Chuanxia and Vedaldi, Andrea},
  booktitle=CVPR,
  year={2024}
}

@misc{jia2024plucker,
  author       = {Jia, Yan-Bin},
  title        = {Plücker Coordinates for Lines in the Space},
  howpublished = {COMS 4770/5770 Notes, Iowa State University},
  year         = {2024},
  note         = {Lecture notes}
}

@inproceedings{kant2024spad,
  title={{SPAD}: Spatially aware multi-view diffusers},
  author={Kant, Yash and Siarohin, Aliaksandr and Wu, Ziyi and Vasilkovsky, Michael and Qian, Guocheng and Ren, Jian and Guler, Riza Alp and Ghanem, Bernard and Tulyakov, Sergey and Gilitschenski, Igor},
  booktitle=CVPR,
  year={2024}
}

@inproceedings{
    miyato2024gta,
    title={{GTA}: A Geometry-Aware Attention Mechanism for Multi-View Transformers},
    author={Takeru Miyato and Bernhard Jaeger and Max Welling and Andreas Geiger},
    booktitle=ICLR,
    year={2024}
}

@article{su2024roformer,
  title={{RoFormer:} Enhanced transformer with rotary position embedding},
  author={Su, Jianlin and Ahmed, Murtadha and Lu, Yu and Pan, Shengfeng and Bo, Wen and Liu, Yunfeng},
  journal={Neurocomputing},
  volume={568},
  pages={127063},
  year={2024}
}

@inproceedings{kong2024eschernet,
  title={{EscherNet}: A generative model for scalable view synthesis},
  author={Kong, Xin and Liu, Shikun and Lyu, Xiaoyang and Taher, Marwan and Qi, Xiaojuan and Davison, Andrew J},
  booktitle=CVPR,
  year={2024}
}

@article{li2025prope,
  title={Cameras as Relative Positional Encoding},
  author={Li, Ruilong and Yi, Brent and Liu, Junchen and Gao, Hang and Ma, Yi and Kanazawa, Angjoo},
  journal={preprint arXiv:2507.10496},
  year={2025}
}

@inproceedings{liu2025dms,
  title={{DMS}: Diffusion-Based Multi-Baseline Stereo Generation for Improving Self-Supervised Depth Estimation},
  author={Liu, Zihua and Li, Yizhou and Zhang, Songyan and Okutomi, Masatoshi},
  booktitle =CVPR,
  year={2025}
}

@inproceedings{xie2016deep3d,
  title={{Deep3D}: Fully automatic 2d-to-3d video conversion with deep convolutional neural networks},
  author={Xie, Junyuan and Girshick, Ross and Farhadi, Ali},
  booktitle=ECCV,
  year={2016}
}

@inproceedings{Tosi2023Nerf,
  author    = {Tosi, Fabio and Tonioni, Alessio and De Gregorio, Daniele and Poggi, Matteo},
  title     = {{NeRF}-Supervised Deep Stereo},
  booktitle = CVPR,
  year      = {2023},
}

@inproceedings{li2025scenesplat,
  title={{SceneSplat}: Gaussian Splatting-based Scene Understanding With Vision-Language Pretraining},
  author={Li, Yue and Ma, Qi and Yang, Runyi and Li, Huapeng and Ma, Mengjiao and Ren, Bin and Popovic, Nikola and Sebe, Nicu and Konukoglu, Ender and Gevers, Theo and others},
  booktitle = ICCV,
  year={2025}
}

@article{muller2022instant,
  title={Instant neural graphics primitives with a multiresolution hash encoding},
  author={M{\"u}ller, Thomas and Evans, Alex and Schied, Christoph and Keller, Alexander},
  journal={ACM Transactions on Graphics (TOG)},
  volume={41},
  number={4},
  pages={1--15},
  year={2022}
}

@inproceedings{wen2024layeredflow,
  title={{LayeredFlow}: A real-world benchmark for non-lambertian multi-layer optical flow},
  author={Wen, Hongyu and Liang, Erich and Deng, Jia},
  booktitle=ECCV,
  year={2024}
}

@inproceedings{wang2025zerostereo,
  title     = {{ZeroStereo}: Zero-shot Stereo Matching from Single Images},
  author    = {Wang, Xianqi and Yang, Hao and Xu, Gangwei and Cheng, Junda and Lin, Min and Deng, Yong and Zang, Jinliang and Chen, Yurui and Yang, Xin},
  booktitle = ICCV,
  year      = {2025},
}

@article{zhang2024spatialme,
  title={{SpatialMe}: Stereo video conversion using depth-warping and blend-inpainting},
  author={Zhang, Jiale and Jia, Qianxi and Liu, Yang and Zhang, Wei and Wei, Wei and Tian, Xin},
  journal={preprint arXiv:2412.11512},
  year={2024}
}

@article{shvetsova2025m2svid,
  title={{M2SVid}: End-to-End Inpainting and Refinement for Monocular-to-Stereo Video Conversion},
  author={Shvetsova, Nina and Bhat, Goutam and Truong, Prune and Kuehne, Hilde and Tombari, Federico},
  journal={preprint arXiv:2505.16565},
  year={2025}
}

@article{huang2025restereo,
  title={Restereo: Diffusion stereo video generation and restoration},
  author={Huang, Xingchang and Singh, Ashish Kumar and Dubost, Florian and Vasconcelos, Cristina Nader and Khattar, Sakar and Shi, Liang and Theobalt, Christian and Oztireli, Cengiz and Singh, Gurprit},
  journal={preprint arXiv:2506.06023},
  year={2025}
}

@inproceedings{wang2020tartanair,
  title={{TartanAir}: A dataset to push the limits of visual slam},
  author={Wang, Wenshan and Zhu, Delong and Wang, Xiangwei and Hu, Yaoyu and Qiu, Yuheng and Wang, Chen and Hu, Yafei and Kapoor, Ashish and Scherer, Sebastian},
  booktitle=IROS,
  year={2020}
}

@inproceedings{wang2019irs,
  title     = {{IRS}: A Large Naturalistic Indoor Robotics Stereo Dataset to Train Deep Models for Disparity and Surface Normal Estimation},
  author    = {Wang, Qiang and Zheng, Shizhen and Yan, Qingsong and Deng, Fei and Zhao, Kaiyong and Chu, Xiaowen},
  booktitle = {IEEE International Conference on Multimedia and Expo (ICME)},
  year      = {2021},
  pages     = {1--6}
}

@inproceedings{tremblay2018falling,
  title={Falling things: A synthetic dataset for 3d object detection and pose estimation},
  author={Tremblay, Jonathan and To, Thang and Birchfield, Stan},
  booktitle=CVPRW,
  year={2018}
}

@article{cabon2020virtualkitti,
  title={Virtual {Kitti} 2},
  author={Cabon, Yohann and Murray, Naila and Humenberger, Martin},
  journal={preprint arXiv:2001.10773},
  year={2020}
}

@article{jing2024infinigen,
  title={Match stereo videos via bidirectional alignment},
  author={Jing, Junpeng and Mao, Ye and Qiu, Anlan and Mikolajczyk, Krystian},
  journal={preprint arXiv:2409.20283},
  year={2024}
}

@INPROCEEDINGS{Tosi2021unreal,
  author = {Fabio Tosi and Yiyi Liao and Carolin Schmitt and Andreas Geiger},
  title = {{SMD-Nets}: Stereo Mixture Density Networks},
  booktitle = CVPR,
  year = {2021}
}

@article{tokarsky2024plt,
  title={{PLT-D3}: A high-fidelity dynamic driving simulation dataset for stereo depth and scene flow},
  author={Tokarsky, Joshua and Abdulhafiz, Ibrahim and Ayyalasomayajula, Satya and Mohsen, Mostafa and Rao, Navya G and Forbes, Adam},
  journal={preprint arXiv:2406.07667},
  year={2024}
}

@inproceedings{buttler2012sintel,
    title = {A naturalistic open source movie for optical flow evaluation},
    author = {Butler, D. J. and Wulff, J. and Stanley, G. B. and Black, M. J.},
    booktitle = ECCV,
    year = {2012}
}

@inproceedings{radford2021clip,
  title={Learning transferable visual models from natural language supervision},
  author={Radford, Alec and Kim, Jong Wook and Hallacy, Chris and Ramesh, Aditya and Goh, Gabriel and Agarwal, Sandhini and Sastry, Girish and Askell, Amanda and Mishkin, Pamela and Clark, Jack and others},
  booktitle=ICML,
  year={2021}
}

@inproceedings{mehl2023spring,
  title={Spring: A high-resolution high-detail dataset and benchmark for scene flow, optical flow and stereo},
  author={Mehl, Lukas and Schmalfuss, Jenny and Jahedi, Azin and Nalivayko, Yaroslava and Bruhn, Andr{\'e}s},
  booktitle=CVPR,
  year={2023}
}

@article{jospin2022simstereo,
  title={Active-passive simstereo-benchmarking the cross-generalization capabilities of deep learning-based stereo methods},
  author={Jospin, Laurent and Antony, Allen and Xu, Lian and Laga, Hamid and Boussaid, Farid and Bennamoun, Mohammed},
  journal=NeurIPS,
  year={2022}
}

@inproceedings{karaev2023dynamic,
  title={{DynamicStereo}: Consistent dynamic depth from stereo videos},
  author={Karaev, Nikita and Rocco, Ignacio and Graham, Benjamin and Neverova, Natalia and Vedaldi, Andrea and Rupprecht, Christian},
  booktitle=CVPR,
  year={2023}
}

@inproceedings{barron2021mipnerf,
    title={{Mip-NeRF}: A Multiscale Representation 
           for Anti-Aliasing Neural Radiance Fields},
    author={Jonathan T. Barron and Ben Mildenhall and 
            Matthew Tancik and Peter Hedman and 
            Ricardo Martin-Brualla and Pratul P. Srinivasan},
    booktitle=ICCV,
    year={2021}
}

@article{barron2022mipnerf360,
    title={{Mip-NeRF 360}: Unbounded Anti-Aliased Neural Radiance Fields},
    author={Jonathan T. Barron and Ben Mildenhall and 
            Dor Verbin and Pratul P. Srinivasan and Peter Hedman},
    journal=CVPR,
    year={2022}
}

@inproceedings{chen2021mvsnerf,
  title={{MVSNeRF}: Fast generalizable radiance field reconstruction from multi-view stereo},
  author={Chen, Anpei and Xu, Zexiang and Zhao, Fuqiang and Zhang, Xiaoshuai and Xiang, Fanbo and Yu, Jingyi and Su, Hao},
  booktitle=ICCV,
  year={2021}
}

@inproceedings{chen2022tensorf,
  title={{TensoRF}: Tensorial radiance fields},
  author={Chen, Anpei and Xu, Zexiang and Geiger, Andreas and Yu, Jingyi and Su, Hao},
  booktitle=ECCV,
  year={2022}
}

@inproceedings{yu2021pixelnerf,
  title={{pixelNeRF}: Neural radiance fields from one or few images},
  author={Yu, Alex and Ye, Vickie and Tancik, Matthew and Kanazawa, Angjoo},
  booktitle=CVPR,
  year={2021}
}

@inproceedings{park2021nerfies,
  title={Nerfies: Deformable neural radiance fields},
  author={Park, Keunhong and Sinha, Utkarsh and Barron, Jonathan T and Bouaziz, Sofien and Goldman, Dan B and Seitz, Steven M and Martin-Brualla, Ricardo},
  booktitle=ICCV,
  year={2021}
}

@inproceedings{pumarola2020d,
    title={{D-NeRF}: Neural Radiance Fields for Dynamic Scenes},
    author={Pumarola, Albert and Corona, Enric and Pons-Moll, Gerard and Moreno-Noguer, Francesc},
    booktitle=CVPR,
    year={2020}
}

@inproceedings{gao2021dynamic,
  title={Dynamic view synthesis from dynamic monocular video},
  author={Gao, Chen and Saraf, Ayush and Kopf, Johannes and Huang, Jia-Bin},
  booktitle=ICCV,
  year={2021}
}

@article{kerbl20233d,
  title={3D Gaussian splatting for real-time radiance field rendering},
  author={Kerbl, Bernhard and Kopanas, Georgios and Leimk{\"u}hler, Thomas and Drettakis, George},
  journal={ACM Transactions on Graphics (TOG)},
  volume={42},
  number={4},
  pages={139--1},
  year={2023}
}

@inproceedings{huang20242d,
  title={2d gaussian splatting for geometrically accurate radiance fields},
  author={Huang, Binbin and Yu, Zehao and Chen, Anpei and Geiger, Andreas and Gao, Shenghua},
  booktitle={ACM SigGraph},
  pages={1--11},
  year={2024}
}

@inproceedings{jiang2024gaussianshader,
  title={{GaussianShader}: 3d gaussian splatting with shading functions for reflective surfaces},
  author={Jiang, Yingwenqi and Tu, Jiadong and Liu, Yuan and Gao, Xifeng and Long, Xiaoxiao and Wang, Wenping and Ma, Yuexin},
  booktitle=CVPR,
  year={2024}
}

@inproceedings{luiten2024dynamic,
  title={Dynamic 3d gaussians: Tracking by persistent dynamic view synthesis},
  author={Luiten, Jonathon and Kopanas, Georgios and Leibe, Bastian and Ramanan, Deva},
  booktitle={International Conference on 3D Vision (3DV)},
  year={2024}
}

@inproceedings{charatan2024pixelsplat,
  title={{pixelSplat}: 3d gaussian splats from image pairs for scalable generalizable 3d reconstruction},
  author={Charatan, David and Li, Sizhe Lester and Tagliasacchi, Andrea and Sitzmann, Vincent},
  booktitle=CVPR,
  year={2024}
}

@inproceedings{chen2024mvsplat,
  title={{MVSplat}: Efficient 3d gaussian splatting from sparse multi-view images},
  author={Chen, Yuedong and Xu, Haofei and Zheng, Chuanxia and Zhuang, Bohan and Pollefeys, Marc and Geiger, Andreas and Cham, Tat-Jen and Cai, Jianfei},
  booktitle=ECCV,
  year={2024}
}

@article{ho2020denoising,
  title={Denoising diffusion probabilistic models},
  author={Ho, Jonathan and Jain, Ajay and Abbeel, Pieter},
  journal=NeurIPS,
  year={2020}
}

@inproceedings{dhariwal2021diffusion,
  title={Diffusion models beat gans on image synthesis},
  author={Dhariwal, Prafulla and Nichol, Alexander},
  booktitle=NeurIPS,
  year={2021}
}

@inproceedings{rombach2022latent,
  title={High-resolution image synthesis with latent diffusion models},
  author={Rombach, Robin and Blattmann, Andreas and Lorenz, Dominik and Esser, Patrick and Ommer, Bj{\"o}rn},
  booktitle=CVPR,
  year={2022}
}

@article{ho2022video,
  title={Video diffusion models},
  author={Ho, Jonathan and Salimans, Tim and Gritsenko, Alexey and Chan, William and Norouzi, Mohammad and Fleet, David J},
  journal=NeurIPS,
  year={2022}
}

@inproceedings{blattmann2023align,
  title={Align your latents: High-resolution video synthesis with latent diffusion models},
  author={Blattmann, Andreas and Rombach, Robin and Ling, Huan and Dockhorn, Tim and Kim, Seung Wook and Fidler, Sanja and Kreis, Karsten},
  booktitle=CVPR,
  year={2023}
}

@inproceedings{zhang2023adding,
  title={Adding conditional control to text-to-image diffusion models},
  author={Zhang, Lvmin and Rao, Anyi and Agrawala, Maneesh},
  booktitle=ICCV,
  year={2023}
}

@inproceedings{qiu2024richdreamer,
  title={{RichDreamer}: A generalizable normal-depth diffusion model for detail richness in text-to-3d},
  author={Qiu, Lingteng and Chen, Guanying and Gu, Xiaodong and Zuo, Qi and Xu, Mutian and Wu, Yushuang and Yuan, Weihao and Dong, Zilong and Bo, Liefeng and Han, Xiaoguang},
  booktitle=CVPR,
  year={2024}
}

@inproceedings{fu2024geowizard,
  title={Geowizard: Unleashing the diffusion priors for 3d geometry estimation from a single image},
  author={Fu, Xiao and Yin, Wei and Hu, Mu and Wang, Kaixuan and Ma, Yuexin and Tan, Ping and Shen, Shaojie and Lin, Dahua and Long, Xiaoxiao},
  booktitle=ECCV,
  year={2024}
}

@inproceedings{
      baranchuk2021label,
      title={Label-Efficient Semantic Segmentation with Diffusion Models},
      author={Dmitry Baranchuk and Andrey Voynov and Ivan Rubachev and Valentin Khrulkov and Artem Babenko},
      booktitle=ICLR,
      year={2022}
}

@inproceedings{tian2024diffuse,
  title={Diffuse attend and segment: Unsupervised zero-shot segmentation using stable diffusion},
  author={Tian, Junjiao and Aggarwal, Lavisha and Colaco, Andrea and Kira, Zsolt and Gonzalez-Franco, Mar},
  booktitle=CVPR,
  year={2024}
}

@inproceedings{rahman2023ambiguous,
  title={Ambiguous medical image segmentation using diffusion models},
  author={Rahman, Aimon and Valanarasu, Jeya Maria Jose and Hacihaliloglu, Ilker and Patel, Vishal M},
  booktitle=CVPR,
  year={2023}
}

@inproceedings{songdenoising,
  title={Denoising Diffusion Implicit Models},
  author={Song, Jiaming and Meng, Chenlin and Ermon, Stefano},
  booktitle={International Conference on Learning Representations},
  year=2021
}

@inproceedings{chen2023diffusiondet,
  title={{DiffusionDet}: Diffusion model for object detection},
  author={Chen, Shoufa and Sun, Peize and Song, Yibing and Luo, Ping},
  booktitle=ICCV,
  year={2023}
}

@inproceedings{xu2023open,
  title={Open-vocabulary panoptic segmentation with text-to-image diffusion models},
  author={Xu, Jiarui and Liu, Sifei and Vahdat, Arash and Byeon, Wonmin and Wang, Xiaolong and De Mello, Shalini},
  booktitle=CVPR,
  year={2023}
}

@inproceedings{lugmayr2022repaint,
  title={Repaint: Inpainting using denoising diffusion probabilistic models},
  author={Lugmayr, Andreas and Danelljan, Martin and Romero, Andres and Yu, Fisher and Timofte, Radu and Van Gool, Luc},
  booktitle=CVPR,
  year={2022}
}

@inproceedings{shi2023mvdream,
  title     = {{MVDream}: Multi-view Diffusion for 3D Generation},
  author    = {Yichun Shi and Peng Wang and Jianglong Ye and Long Mai and Kejie Li and Xiao Yang},
  booktitle = ICLR,
  year      = {2024}
}

@article{yang2024depth,
  title={Depth anything v2},
  author={Yang, Lihe and Kang, Bingyi and Huang, Zilong and Zhao, Zhen and Xu, Xiaogang and Feng, Jiashi and Zhao, Hengshuang},
  journal=NeurIPS,
  year={2024}
}

@article{gao2024cat3d,
    title={{CAT3D}: Create Anything in 3D with Multi-View Diffusion Models},
    author={Ruiqi Gao* and Aleksander Holynski* and Philipp Henzler and Arthur Brussee and Ricardo Martin-Brualla and Pratul P. Srinivasan and Jonathan T. Barron and Ben Poole*
    },
    journal=NeurIPS,
    year={2024}
}

@article{bahmani2025lyra,
  title={Lyra: Generative 3d scene reconstruction via video diffusion model self-distillation},
  author={Bahmani, Sherwin and Shen, Tianchang and Ren, Jiawei and Huang, Jiahui and Jiang, Yifeng and Turki, Haithem and Tagliasacchi, Andrea and Lindell, David B and Gojcic, Zan and Fidler, Sanja and others},
  journal={preprint arXiv:2509.19296},
  year={2025}
}

@inproceedings{sargent2024zeronvs,
  title={{ZeroNVS}: Zero-shot 360-degree view synthesis from a single image},
  author={Sargent, Kyle and Li, Zizhang and Shah, Tanmay and Herrmann, Charles and Yu, Hong-Xing and Zhang, Yunzhi and Chan, Eric Ryan and Lagun, Dmitry and Fei-Fei, Li and Sun, Deqing and Wu, Jiajun},
  booktitle=CVPR,
  year={2024}
}

@article{tang2023mvdiffusion,
  author = {Tang, Shitao and Zhang, Fuyang and Chen, Jiacheng and Wang, Peng and Furukawa, Yasutaka},
  title = {{MVDiffusion}: Enabling Holistic Multi-view Image Generation with Correspondence-Aware Diffusion},
  journal = NeurIPS,
  year = {2023},
}

@inproceedings{wang2024motionctrl,
  title={{MotionCtrl}: A unified and flexible motion controller for video generation},
  author={Wang, Zhouxia and Yuan, Ziyang and Wang, Xintao and Li, Yaowei and Chen, Tianshui and Xia, Menghan and Luo, Ping and Shan, Ying},
  booktitle={ACM SigGraph},
  year={2024}
}

@article{liang2024wonderland,
  title={Wonderland: Navigating 3D Scenes from a Single Image},
  author={Liang, Hanwen and Cao, Junli and Goel, Vidit and Qian, Guocheng and Korolev, Sergei and Terzopoulos, Demetri and Plataniotis, Konstantinos N and Tulyakov, Sergey and Ren, Jian},
  journal=CVPR,
  year={2025}
}

@article{he2024cameractrl,
  title={{CameraCtrl}: Enabling camera control for text-to-video generation},
  author={He, Hao and Xu, Yinghao and Guo, Yuwei and Wetzstein, Gordon and Dai, Bo and Li, Hongsheng and Yang, Ceyuan},
  journal={preprint arXiv:2404.02101},
  year={2024}
}

@article{yang2023diffusion,
  title={Diffusion models: A comprehensive survey of methods and applications},
  author={Yang, Ling and Zhang, Zhilong and Song, Yang and Hong, Shenda and Xu, Runsheng and Zhao, Yue and Zhang, Wentao and Cui, Bin and Yang, Ming-Hsuan},
  journal={ACM Computing Surveys},
  volume={56},
  number={4},
  pages={1--39},
  year={2023}
}

@inproceedings{bahmani2025ac3d,
  title={{AC3D}: Analyzing and improving 3d camera control in video diffusion transformers},
  author={Bahmani, Sherwin and Skorokhodov, Ivan and Qian, Guocheng and Siarohin, Aliaksandr and Menapace, Willi and Tagliasacchi, Andrea and Lindell, David B and Tulyakov, Sergey},
  booktitle=CVPR,
  year={2025}
}

@article{xu2024camco,
  title={{CamCo}: Camera-Controllable 3D-Consistent Image-to-Video Generation},
  author={Xu, Dejia and Nie, Weili and Liu, Chao and Liu, Sifei and Kautz, Jan and Wang, Zhangyang and Vahdat, Arash},
  journal={preprint arXiv:2406.02509},
  year={2024}
}

@inproceedings{liu2023zero,
  title={Zero-1-to-3: Zero-shot one image to 3d object},
  author={Liu, Ruoshi and Wu, Rundi and Van Hoorick, Basile and Tokmakov, Pavel and Zakharov, Sergey and Vondrick, Carl},
  booktitle=ICCV,
  year={2023}
}

@inproceedings{mou2024t2i,
  title={{T2I-Adapter}: Learning adapters to dig out more controllable ability for text-to-image diffusion models},
  author={Mou, Chong and Wang, Xintao and Xie, Liangbin and Wu, Yanze and Zhang, Jian and Qi, Zhongang and Shan, Ying},
  booktitle={AAAI Conference on Artificial Intelligence},
  year={2024}
}

@inproceedings{wang2025moge,
    title={{MoGe-2}: Accurate Monocular Geometry with Metric Scale and Sharp Details},
    author={Ruicheng Wang and Sicheng Xu and Yue Dong and Yu Deng and Jianfeng Xiang and Zelong Lv and Guangzhong Sun and Xin Tong and Jiaolong Yang},
    booktitle=NeurIPS,
    year={2025},
}

@inproceedings{he2024diffusion,
  title={A diffusion-based framework for multi-class anomaly detection},
  author={He, Haoyang and Zhang, Jiangning and Chen, Hongxu and Chen, Xuhai and Li, Zhishan and Chen, Xu and Wang, Yabiao and Wang, Chengjie and Xie, Lei},
  booktitle={AAAI Conference on Artificial Intelligence},
  year={2024}
}
}

\clearpage
\maketitlesupplementary

\begin{table}[t]
\centering
\small
\setlength{\intextsep}{6pt}
\begin{tabular}{@{}lcccc@{}}
\toprule
\textbf{Dataset} & \textbf{Baseline} & \textbf{Setting} & \textbf{\#Samples} & \textbf{Year} \\
\midrule
NeRF-Stereo~\cite{Tosi2023Nerf}          & \baseshuffle & \envtree \envhouse & 27K  & 2023 \\
SceneSplat~\cite{li2025scenesplat}       & \baseshuffle & \envhouse & 5K   & 2025 \\
\noalign{\vskip 0.4em}
\arrayrulecolor{rulegray}\cdashline{1-5}[.5pt/1.6pt]
\noalign{\vskip 0.4em}
\arrayrulecolor{black}
TartanAir~\cite{wang2020tartanair}       & $25$~cm & \envtree \envhouse & 306K & 2020 \\
Dynamic Replica~\cite{karaev2023dynamic} & \baseshuffle & \envhouse & 145K & 2023 \\
IRS~\cite{wang2019irs}                   & $10$~cm & \envhouse & 103K & 2021 \\
Falling Things~\cite{tremblay2018falling}& $6$~cm & \envtree \envhouse & 61K  & 2018 \\
LayeredFlow~\cite{wen2024layeredflow}    & \baseshuffle & \envhouse & 31K  & 2024 \\
VKITTI2~\cite{cabon2020virtualkitti}     & $53.3$~cm & \envtree & 21K  & 2020 \\
InfinigenSV~\cite{jing2024infinigen}     & \baseshuffle & \envtree & 17K  & 2024 \\
SimStereo~\cite{jospin2022simstereo}     & $16$~cm & \envhouse & 14K  & 2022 \\
UnrealStereo4K~\cite{Tosi2021unreal}     & \baseshuffle & \envtree \envhouse & 8K   & 2021 \\
Spring~\cite{mehl2023spring}             & \baseshuffle & \envtree & 5K   & 2023 \\
PLT-D3~\cite{tokarsky2024plt}            & $12$~cm & \envtree & 3K   & 2024 \\
Sintel~\cite{buttler2012sintel}          & $10$~cm & \envtree & 1K   & 2012 \\
\bottomrule
\end{tabular}\vspace{-0.3cm}
\caption{\textbf{Training data.} Stereo datasets used for mixed training of \emph{StereoSpace}. For each dataset, we report the baseline (fixed value when available, \baseshuffle\ indicates a variable baseline), scene type (indoor~\envhouse~or outdoor~\envtree), the number of samples, and the release year. Datasets above the dotted line \protect\tabdividerlegend are multi-baseline.}
\label{tab:datasets}
\end{table}


\noindent 
Section~\ref{sec:supp_training} reports additional details concerning the training of StereoSpace, Section~\ref{sec:supp_conditioning} describes the main conditioning mechanisms used to allow for camera control, Section~\ref{sec:supp_evaluation} describes the evaluation split composition in detail and finally Section~\ref{sec:supp_qualitative} shows further qualitative results.

\section{Training Details}
\label{sec:supp_training}

\noindent\textbf{Dataset Composition.} Our training data is drawn from 14 publicly available data sources. Here, we provide additional statistics in Tab.~\ref{tab:datasets} for the stereo datasets. We report the nominal baseline, whether a dataset primarily contains indoor or outdoor scenes, the number of training stereo pairs and the release year, respectively. Together, these sources cover a diverse range of camera baselines, scene scales, and photorealism levels.

\noindent\textbf{Disparity Supervision.} Most datasets provide ground-truth left-to-right and right-to-left disparity maps. When some direction is missing, we infer the respective disparity using a strong off-the-shelf stereo matcher (FoundationStereo~\cite{wen2025stereo}), and treat the resulting pair as pseudo-ground-truth. Having both directions is required to compute the left–right consistency mask used in the warping loss 
(Sec. 3.3 in the main paper),
which restricts supervision to pixels that are co-visible in the source and target views. The synthetic LayeredFlow dataset is a notable exception: its multi-layer depth representation violates the single-surface assumption underlying our view warping formulation. Therefore, for LayeredFlow samples, we disable the warping loss and re-weight the training batches accordingly.

\begin{figure}[t]
 \centering
 \includegraphics[width=0.4\textwidth]{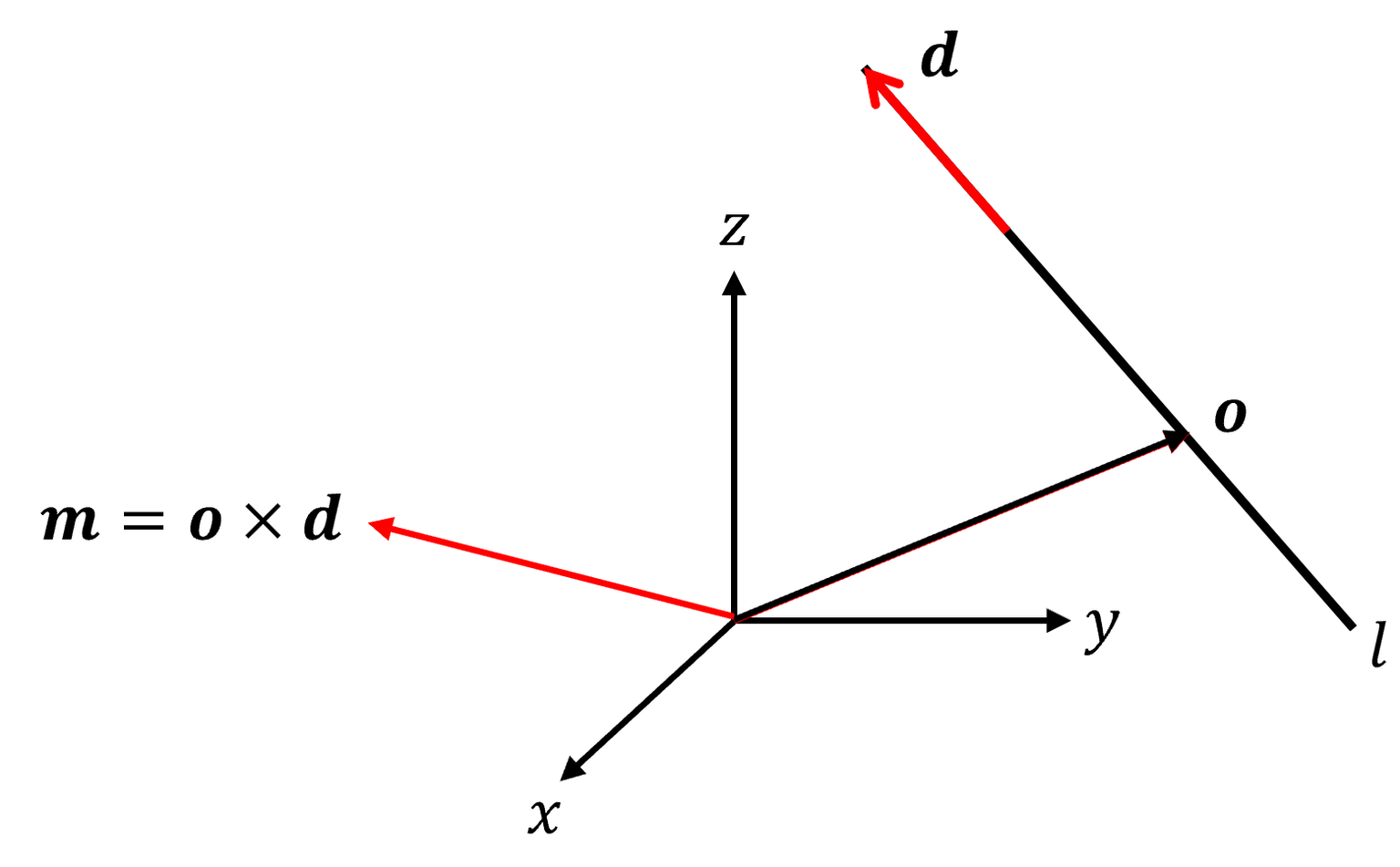}\vspace{-0.3cm}
 \caption{\textbf{Plücker coordinates} of line $\ell$ are given by the 6D homogeneous vector $(\mathbf{d},\,\mathbf{m})$.}
 \label{fig:pluecker}
\end{figure}

\noindent\textbf{Rendered Multi-Baseline Tuples.} For SceneSplat-7K~\cite{li2025scenesplat}, we derive multi-baseline, rectified stacks from the pre-optimized Gaussian splats. We restrict to the Hypersim~\cite{hypersim}, Replica~\cite{replica19arxiv}, and ScanNet++~\cite{yeshwanthliu2023scannetpp} subsets, filtering scenes using the dataset-provided PSNR, SSIM, LPIPS, and depth $\ell_1$ metrics to discard photometrically or geometrically unstable reconstructions. Within each retained scene, we form candidate stacks with moderate baselines and subsamples up to $20$ diverse bundles via $k$-means over stack centroids per splat. Each selected stack is rendered to RGB and depth with a small global focal-length scaling jitter, and stacks with insufficient geometric support are removed based on simple depth- and opacity-based heuristics.

\noindent\textbf{Training Schedule.} StereoSpace, including all its tested variations, is trained for $3$ epochs, corresponding to approximately $48.6$~K optimizer steps. This schedule matches the GenStereo~\cite{qiao2025genstereo} setup and ensures a comparable training budget between variants in the ablation study (Sec. 4.4). 
We also experimented with the extension of training to $5$ epochs, but observed saturated performance and no consistent improvement relative to added computation, so all reported results adhere to the $3$-epoch schedule. In contrast to prior work, StereoSpace is trained in both stereo directions (left$\rightarrow$right and right$\rightarrow$left). Empirically, this bidirectional supervision did not degrade performance, while enabling inference-time stereo generation in either direction, whereas competing approaches typically support only the left$\rightarrow$right direction.

\begin{table*}[t]
\centering
\small
\setlength{\tabcolsep}{10pt}
\renewcommand{\arraystretch}{1.15}
\resizebox{\linewidth}{!}{
\begin{tabular}{llrrrrr}
\toprule
\textbf{Dataset} & \textbf{Metric} & \textbf{GenStereo} & \textbf{Lyra} & \textbf{StereoSpace (ours)} & \textbf{StereoDiffusion} & \textbf{ZeroStereo} \\
\noalign{\vspace{-0.75ex}}
\midrule
\multirow{3}{*}{Middlebury} & PSNR $\uparrow$ & \fst{18.54} & \snd{18.06} & 17.04 & 17.06 & \trd{17.48} \\
\noalign{\vspace{-0.75ex}}
 & SSIM $\uparrow$ & \fst{0.5989} & \snd{0.5793} & \trd{0.5410} & 0.5246 & 0.5372 \\
\noalign{\vspace{-0.75ex}}
 & LPIPS $\downarrow$ & \fst{0.184} & \trd{0.247} & \snd{0.234} & 0.304 & 0.264 \\
\noalign{\vspace{-0.75ex}}
\midrule
\multirow{3}{*}{DrivingStereo} & PSNR $\uparrow$ & \fst{24.95} & \snd{24.25} & 23.44 & 22.89 & \trd{23.64} \\
\noalign{\vspace{-0.75ex}}
 & SSIM $\uparrow$ & \fst{0.7939} & \snd{0.7792} & 0.7424 & 0.7365 & \trd{0.7511} \\
\noalign{\vspace{-0.75ex}}
 & LPIPS $\downarrow$ & \fst{0.136} & 0.172 & \snd{0.164} & 0.193 & \trd{0.170} \\
\noalign{\vspace{-0.75ex}}
\midrule
\multirow{3}{*}{Booster} & PSNR $\uparrow$ & \trd{21.74} & \fst{23.21} & \snd{21.91} & 20.42 & 18.69 \\
\noalign{\vspace{-0.75ex}}
 & SSIM $\uparrow$ & \trd{0.7348} & \fst{0.7813} & \snd{0.7351} & 0.6986 & 0.5733 \\
\noalign{\vspace{-0.75ex}}
 & LPIPS $\downarrow$ & \trd{0.237} & \fst{0.196} & \snd{0.202} & 0.322 & 0.443 \\
\noalign{\vspace{-0.75ex}}
\midrule
\multirow{3}{*}{LayeredFlow} & PSNR $\uparrow$ & \trd{17.44} & \fst{18.67} & \snd{17.67} & 17.08 & 15.83 \\
\noalign{\vspace{-0.75ex}}
 & SSIM $\uparrow$ & 0.5912 & \fst{0.6327} & \trd{0.5940} & \snd{0.5957} & 0.4421 \\
\noalign{\vspace{-0.75ex}}
 & LPIPS $\downarrow$ & \trd{0.370} & \fst{0.321} & \snd{0.335} & 0.433 & 0.521 \\
\noalign{\vspace{-0.75ex}}
\bottomrule
\end{tabular}}\vspace{-0.3cm}
\caption{\textbf{Evaluation with conventional metrics -- PSNR, SSIM, LPIPS.} Experiments on Middlebury \cite{scharstein2014middlebury}, DrivingStereo \cite{yang2019driving}, Booster \cite{ramirez2022booster} and LayeredFlow \cite{wen2024layeredflow}.}
\label{tab:stereo-metrics}
\end{table*}

\section{Viewpoint Conditioning}
\label{sec:supp_conditioning}

\noindent In this section, we recall the principles behind the conditioning mechanisms used by StereoSpace.

\noindent\textbf{Plücker rays.}
A 3D line $\ell$ can be represented by a point $\mathbf{o}\in\mathbb{R}^3$ and a (unit) direction $\mathbf{d}\in\mathbb{R}^3$ (Fig.~\ref{fig:pluecker}). Its Plücker (Grassmann) coordinates are the homogeneous 6D vector
\begin{equation}
   \ell \equiv (\mathbf{d},\,\mathbf{m}), \qquad \mathbf{m} = \mathbf{o} \times \mathbf{d}.
\end{equation}
By construction $\mathbf{d}\!\cdot\!\mathbf{m} = 0$ (the Plücker constraint), and for any other point $\mathbf{o}'=\mathbf{o}+\lambda \mathbf{d}$ on the same line we have $(\mathbf{o}'\times \mathbf{d})=\mathbf{m}$. Hence, Plücker coordinates are invariant to sliding $\mathbf{o}$ along the line, which makes them a natural parametrization of camera rays~\cite{jia2024plucker}.

For a pinhole camera with center $\mathbf{c}$ (in world coordinates), the ray through the pixel $(i,j)$ has Plücker coordinates $\ell_{ij}=(\mathbf{d}_{ij},\mathbf{m}_{ij})$ with
\begin{equation}
   \mathbf{m}_{ij}=\mathbf{c}\times \mathbf{d}_{ij}.
\end{equation}
We form dense Plücker embeddings $\mathbf{F}_{\text{plucker}}\in\mathbb{R}^{6\times H\times W}$ by concatenating $(\mathbf{d}_{ij},\mathbf{m}_{ij})$ for each pixel of an image of size $(H,W)$. Because $(\mathbf{d},\mathbf{m})$ are homogeneous, $s(\mathbf{d},\mathbf{m})$ with $s\!\neq\!0$ represents the same (unoriented) line. For rays, we fix this gauge by normalizing $\|\mathbf{d}\|=1$ and choosing the sign so that $\mathbf{d}$ points from the camera into the scene.

This representation encodes the camera geometry in a distributed way: instead of a single global pose vector, each pixel’s ray is tagged with a 6D vector that implicitly contains the camera intrinsics and extrinsics for that viewline. Thus, the diffusion model can, in principle, attend to the 3D configuration of rays when conditioning on the view.

Plücker coordinates also admit simple expressions for line-line relations. Given two rays
$\ell_k=(\mathbf{d}_k,\mathbf{m}_k)$, the reciprocal product
\begin{equation}
   \langle \ell_1,\ell_2\rangle \;=\; \mathbf{d}_1\!\cdot\!\mathbf{m}_2 \;+\; \mathbf{d}_2\!\cdot\!\mathbf{m}_1
\end{equation}
vanishes iff the two rays are coplanar (i.e., they intersect or are parallel)~\cite{jia2024plucker}. For non-parallel rays, their shortest distance is
\begin{equation}
   d(\ell_1,\ell_2)\;=\;\frac{\big|\mathbf{d}_1\!\cdot\!\mathbf{m}_2+\mathbf{d}_2\!\cdot\!\mathbf{m}_1\big|}{\|\mathbf{d}_1\times \mathbf{d}_2\|}.
\end{equation}
Rays that image the same 3D point are coplanar and intersect, and rays to nearby points have small reciprocal product and line-line distance. Importantly, these quantities are bilinear in $(\mathbf{d},\mathbf{m})$, so they can be implemented by linear projections and dot products on $\mathbf{F}_{\text{plucker}}$.

Although the computation of such expressions is not enforced explicitly in the network, Plücker ray embeddings provide an inductive bias that makes cross-view geometric consistency easy to test in the input space. Empirically, such per-pixel ray conditioning has been shown to improve camera pose accuracy and 3D consistency in generative models~\cite{zheng2024free3d,kant2024spad}, and we observe similar benefits. By contrast, global pose encodings (e.g., Euler angles) offer no direct notion of pixel-to-pixel correspondences across views and often suffer from symmetries such as front-back ambiguity.

\begin{figure*}[t]
  \centering
  \begin{tabular}{c}
       \vspace{-0.2cm}{\includegraphics[clip,trim=0cm 1cm 0cm 0cm, width=\linewidth]{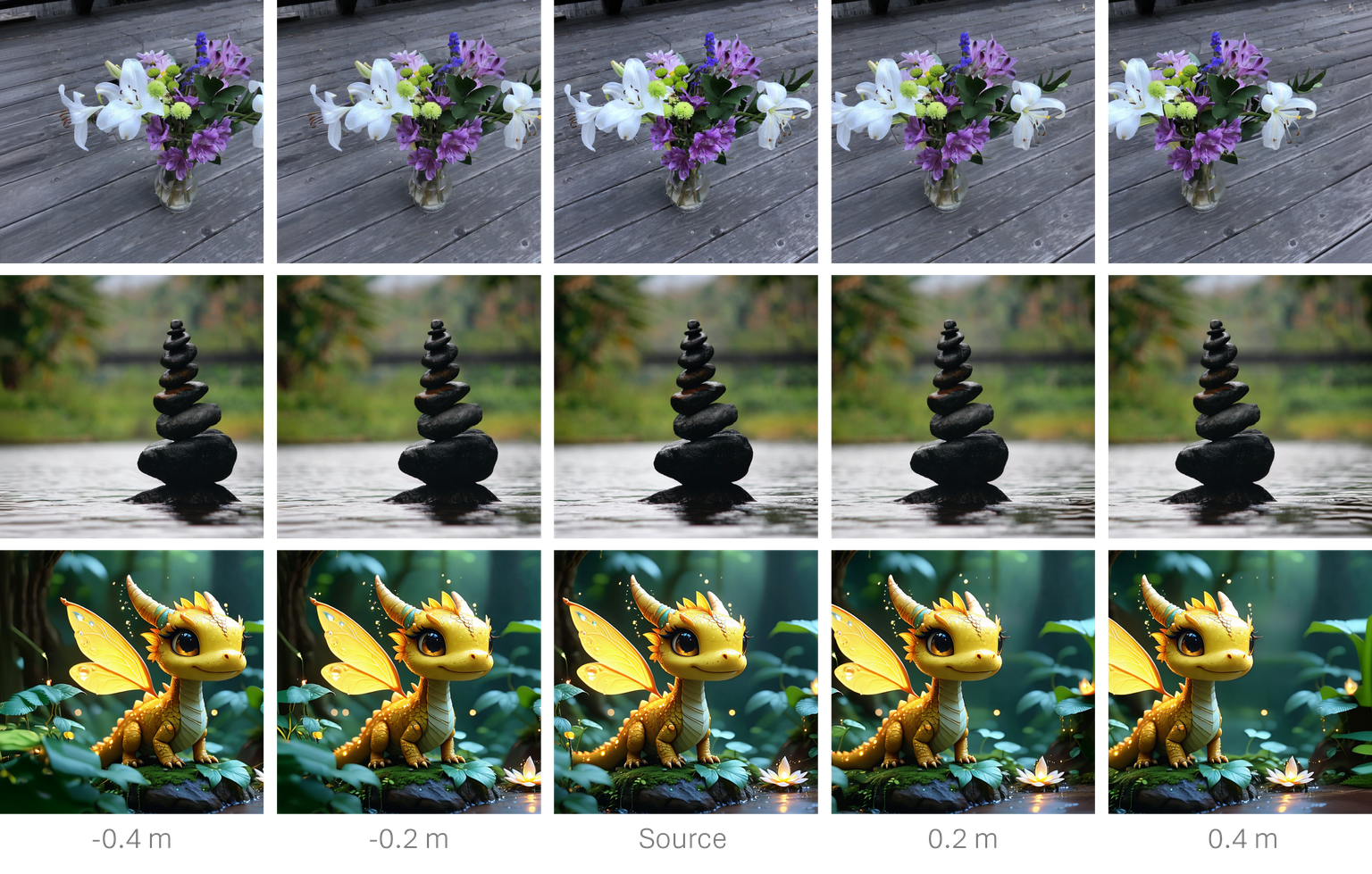}} \\ 
       -0.4m \hspace{0.1\linewidth} \quad\quad -0.2m \hspace{0.1\linewidth} \quad\quad Source \hspace{0.1\linewidth} \quad\quad 0.2m \hspace{0.1\linewidth} \quad\quad 0.4m \\
  \end{tabular}
   \vspace{-0.4cm}
  \caption{\textbf{Qualitative results of multiple inferences with varying baseline.} StereoSpace naturally supports rendering images captured with arbitrary baselines, including viewpoints located to the left (negative baseline) and to the right (positive baseline) of the source image.} 
  \label{fig:multi-baseline-inference}
\end{figure*}

\noindent\textbf{PRoPE attention.}
Besides Plücker-ray conditioning, we also evaluate an attention-level camera encoding based on Geometric Transform Attention (GTA)~\cite{miyato2024gta}, CaPE~\cite{kong2024eschernet}, RoPE~\cite{su2024roformer}, and the recent PRoPE~\cite{li2025prope}. In this variant, each token $t$ from camera $i(t)$ is associated with (i) a projective transform derived from the camera’s projection matrix and (ii) a rotary position embedding of its 2D image coordinates $(x_t,y_t)$. Following the GTA framework, these per-token transforms $D_t$ are multiplied into $Q,K,V$ so that attention logits depend on the \emph{relative} projective transform between the cameras of the query and key tokens, while RoPE handles within-image spatial relations.

We apply PRoPE-style attention in all cross-attention layers where the denoising stream attends to reference views. The PRoPE mechanism complements Plücker embeddings: both encode the full camera frustum (intrinsics \(+\) extrinsics) but at different stages: Plücker rays provide per-pixel geometry in the input channels and ResNet blocks, while PRoPE aligns token features via the relative projective transform inside attention. 
However, as reported in our ablation study, the combination of the two does not produce any significant improvement, but Plücker embeddings alone achieve both the lowest iSQoE and MEt3R scores.

\begin{figure*}[t]
    \centering
    \renewcommand{\tabcolsep}{1pt}
    \begin{tabular}{ccccccc}
        & Left Image & ZeroStereo \cite{wang2025zerostereo} & StereoDiffusion \cite{wang2024stereodiffusion} & GenStereo \cite{qiao2025genstereo} & Lyra \cite{bahmani2025lyra} & \bf StereoSpace (ours) \smallskip\\
    
         \rotatebox[origin=l]{90}{\quad\texttt{Adirondack}} & \includegraphics[width=0.16\textwidth]{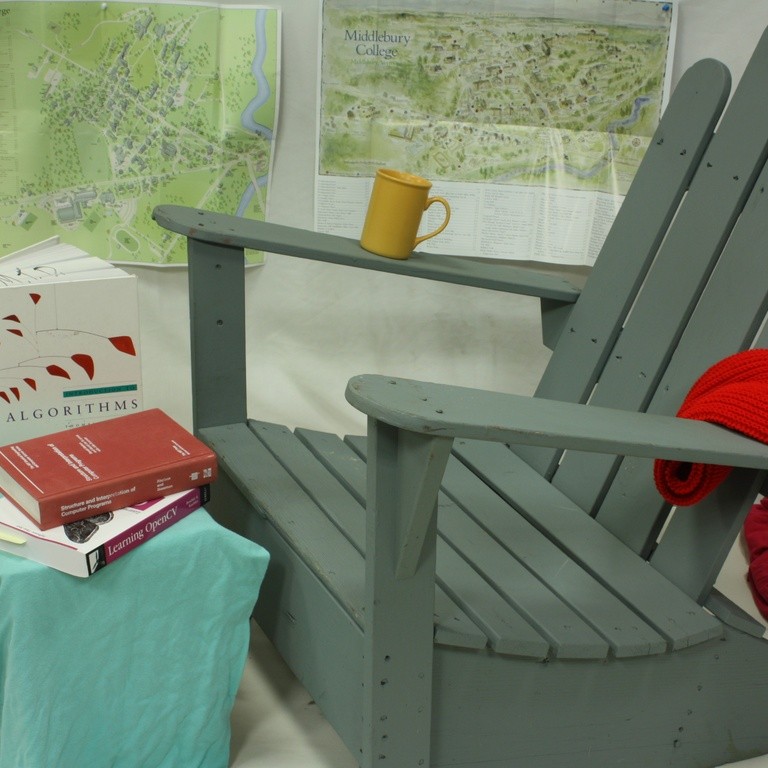} &
         \includegraphics[width=0.16\textwidth]{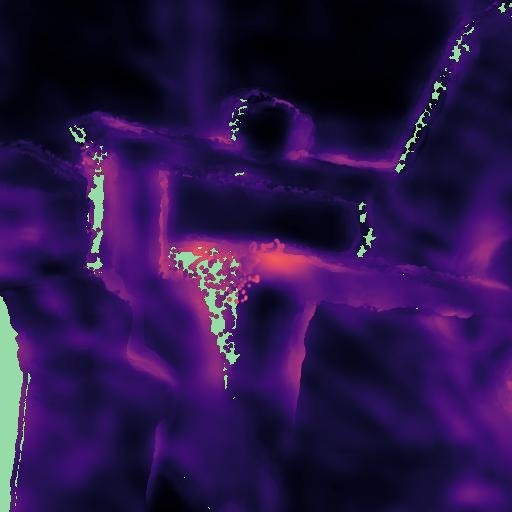} &
         \includegraphics[width=0.16\textwidth]{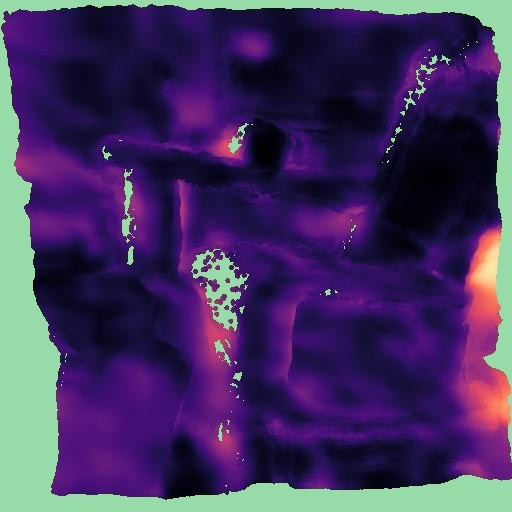} &
         \includegraphics[width=0.16\textwidth]{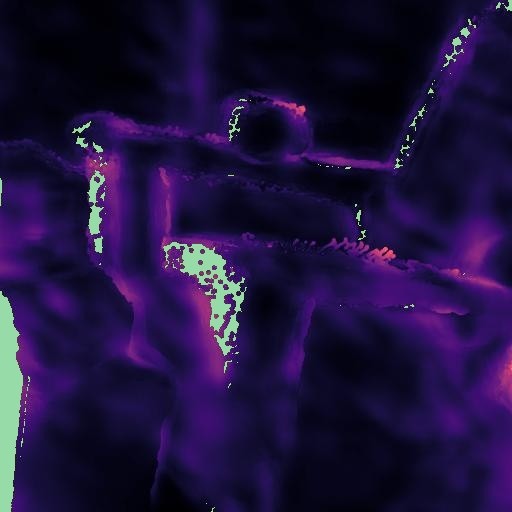} &
         \includegraphics[width=0.16\textwidth]{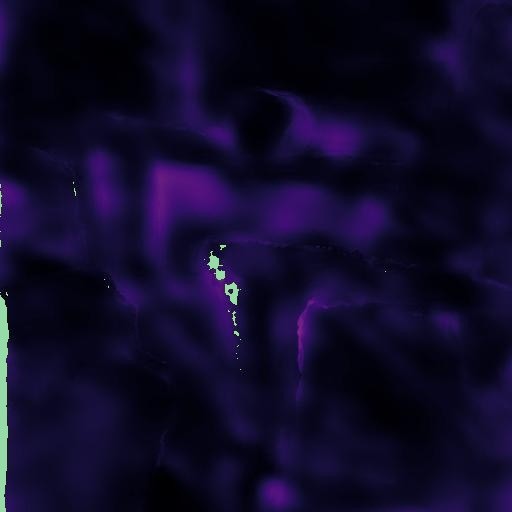} &
         \includegraphics[width=0.16\textwidth]{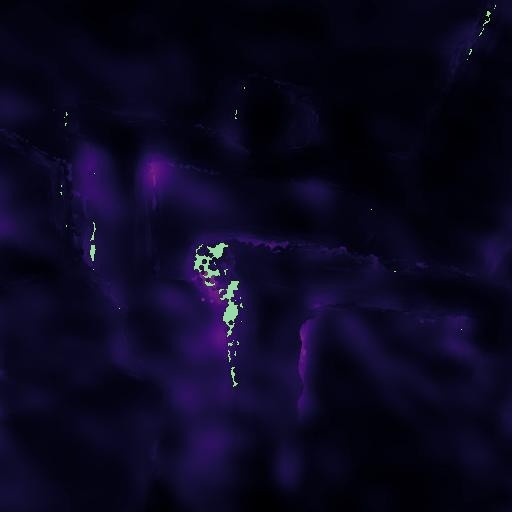} \vspace{-0.1cm}\\
         & MEt3R scores:  & 0.1670 & 0.2045 & 0.1311 & 0.1016 & 0.0747 \smallskip\\

         \rotatebox[origin=l]{90}{\quad\quad\texttt{Piano}} & \includegraphics[width=0.16\textwidth]{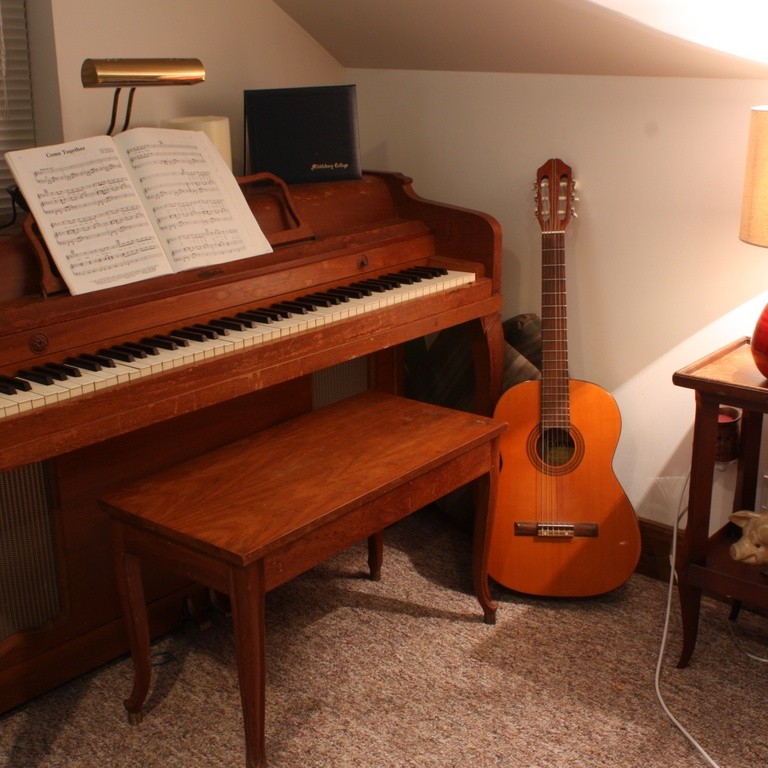} &
         \includegraphics[width=0.16\textwidth]{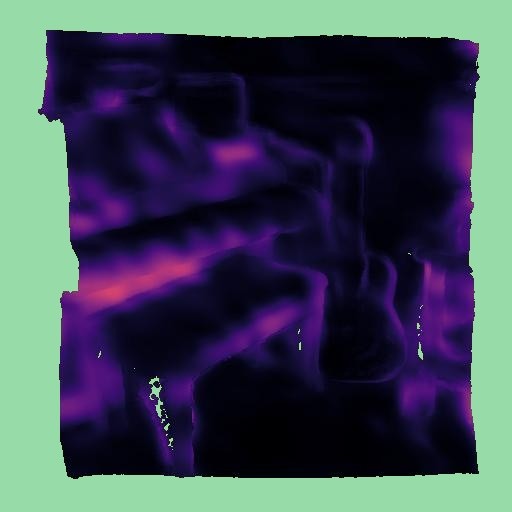} &
         \includegraphics[width=0.16\textwidth]{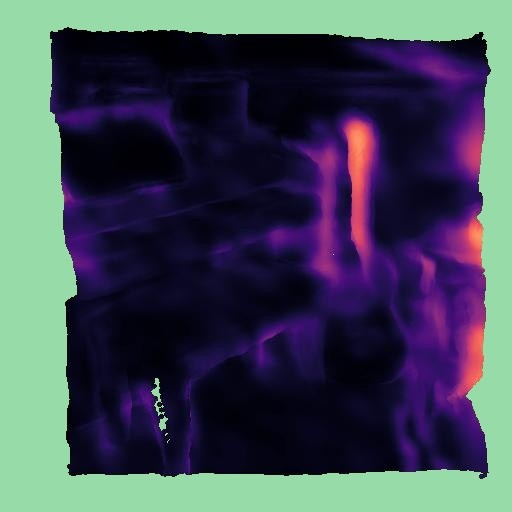} &
         \includegraphics[width=0.16\textwidth]{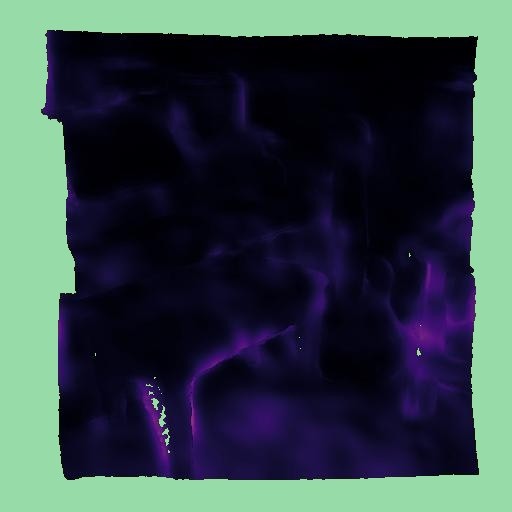} &
         \includegraphics[width=0.16\textwidth]{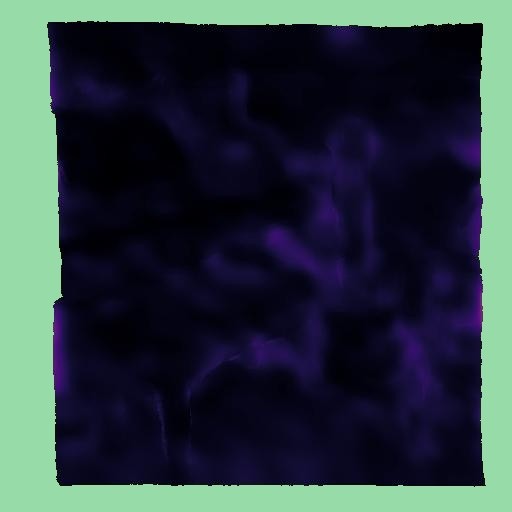} &
         \includegraphics[width=0.16\textwidth]{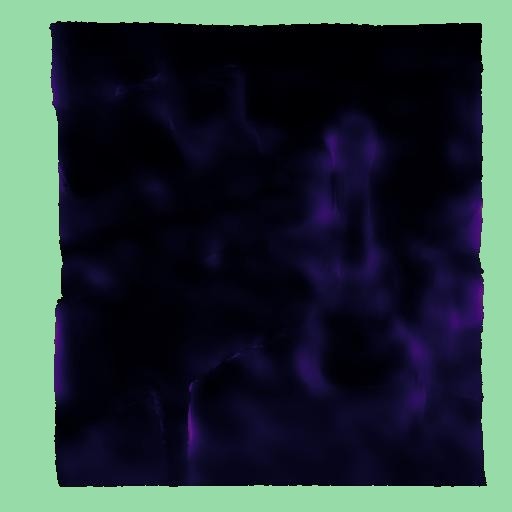} \vspace{-0.1cm}\\
         & MEt3R scores:  & 0.1241 & 0.1285 & 0.0709 & 0.0755 & 0.0582 \smallskip\\

         \rotatebox[origin=l]{90}{\quad\quad\texttt{Pipes}} & \includegraphics[width=0.16\textwidth]{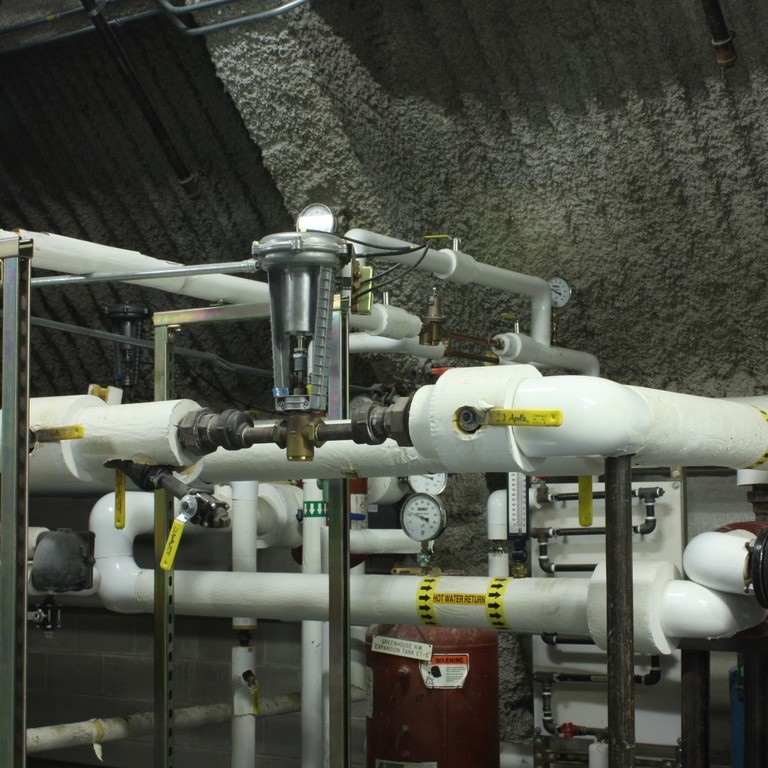} &
         \includegraphics[width=0.16\textwidth]{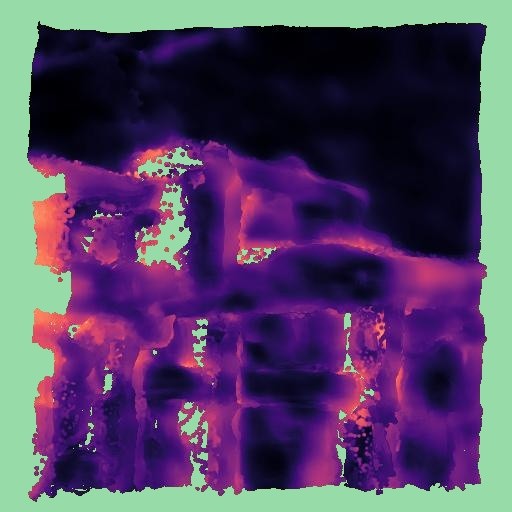} &
         \includegraphics[width=0.16\textwidth]{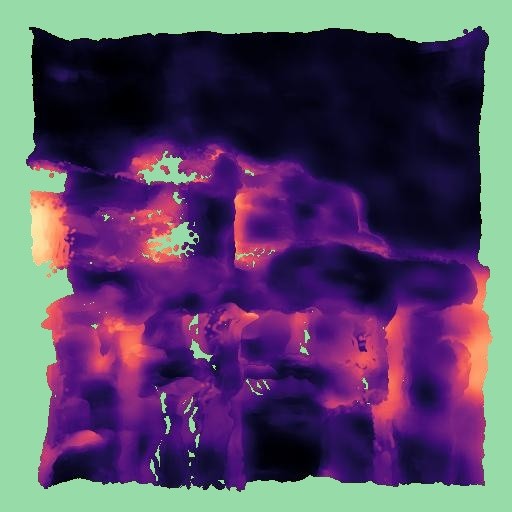} &
         \includegraphics[width=0.16\textwidth]{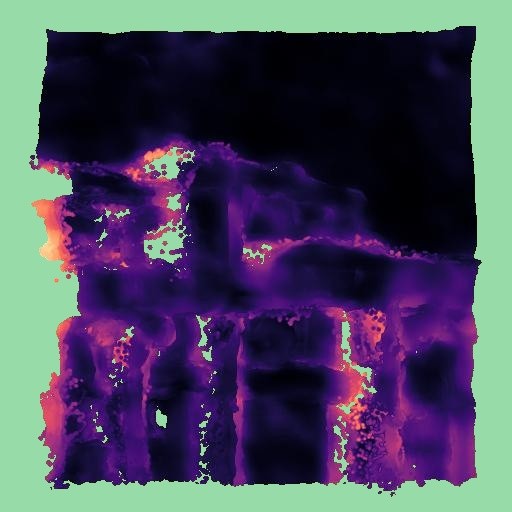} &
         \includegraphics[width=0.16\textwidth]{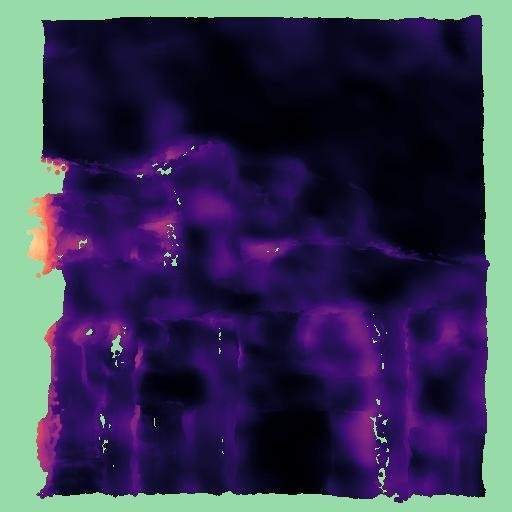} &
         \includegraphics[width=0.16\textwidth]{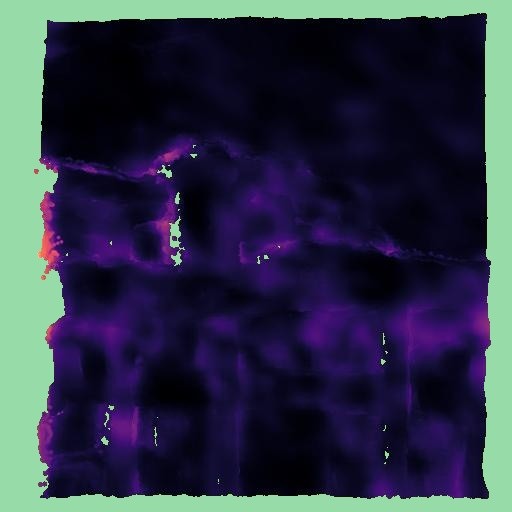} \vspace{-0.1cm}\\
         & MEt3R scores:  & 0.2342 & 0.2486 & 0.1683 & 0.1573 & 0.1137 \smallskip\\

         \rotatebox[origin=l]{90}{\quad\quad\texttt{Shopvac}} & \includegraphics[width=0.16\textwidth]{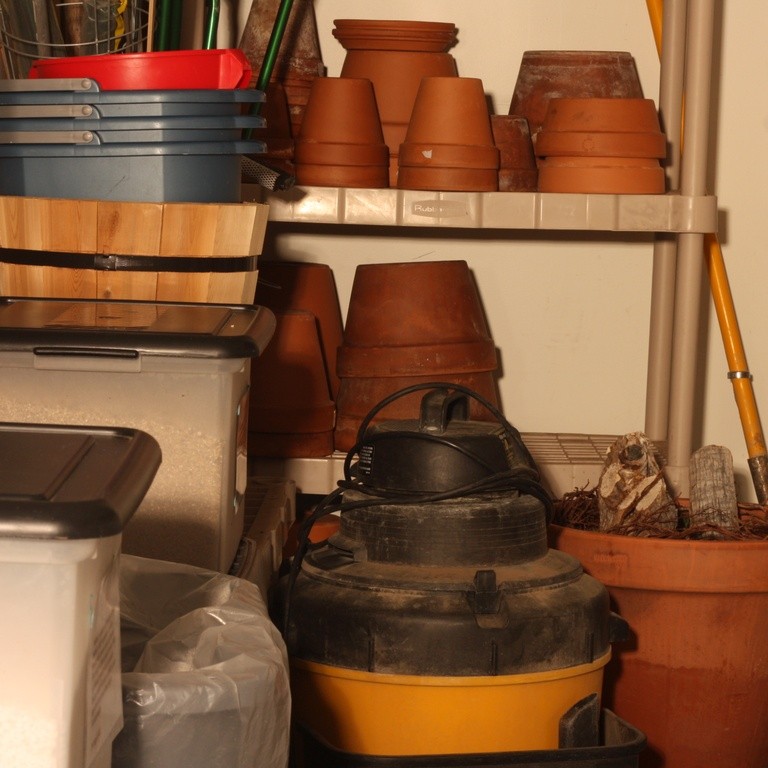} &
         \includegraphics[width=0.16\textwidth]{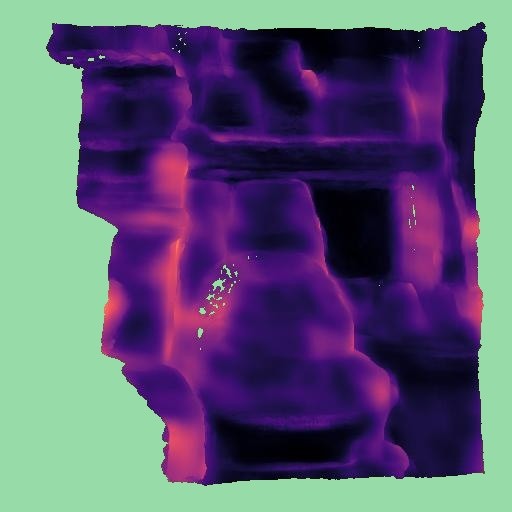} &
         \includegraphics[width=0.16\textwidth]{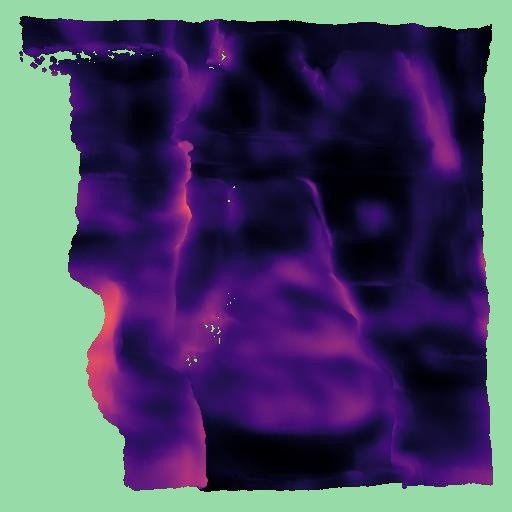} &
         \includegraphics[width=0.16\textwidth]{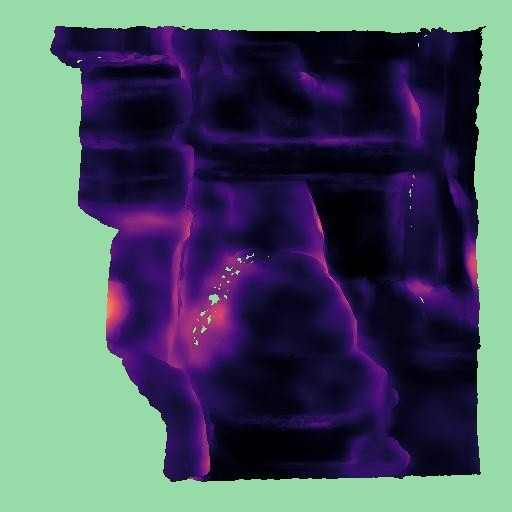} &
         \includegraphics[width=0.16\textwidth]{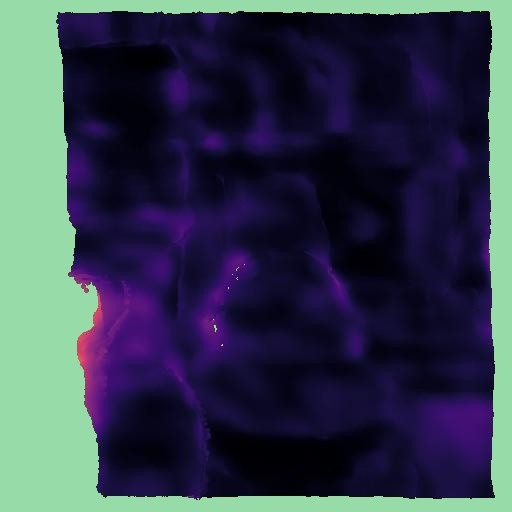} &
         \includegraphics[width=0.16\textwidth]{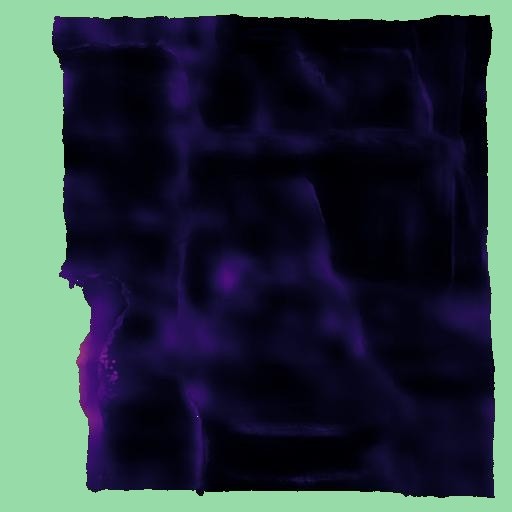} \vspace{-0.1cm}\\
         & MEt3R scores:  & 0.2318 & 0.1919 & 0.1471 & 0.1135 & 0.0732 \\

         \rotatebox[origin=l]{90}{\quad\quad\texttt{Sticks}} & \includegraphics[width=0.16\textwidth]{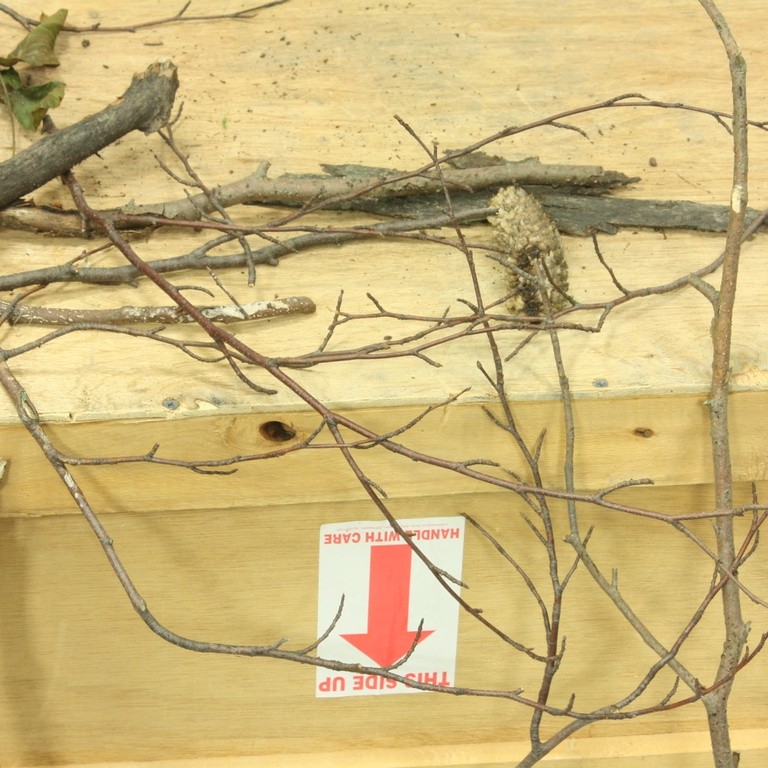} &
         \includegraphics[width=0.16\textwidth]{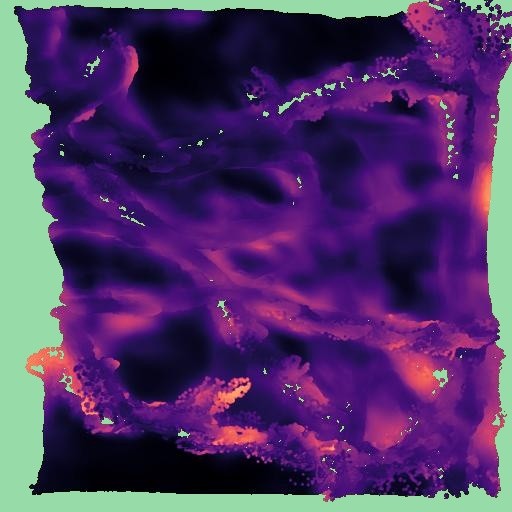} &
         \includegraphics[width=0.16\textwidth]{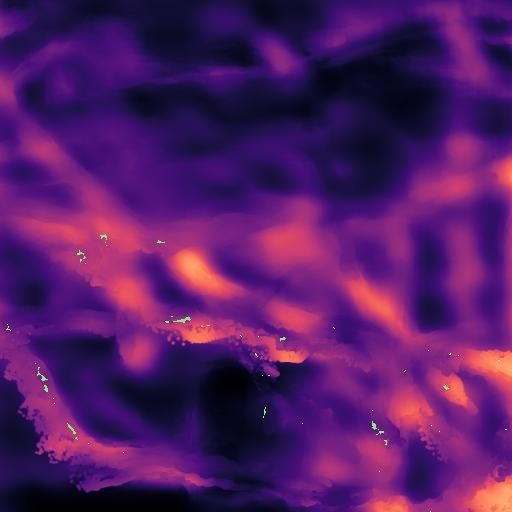} &
         \includegraphics[width=0.16\textwidth]{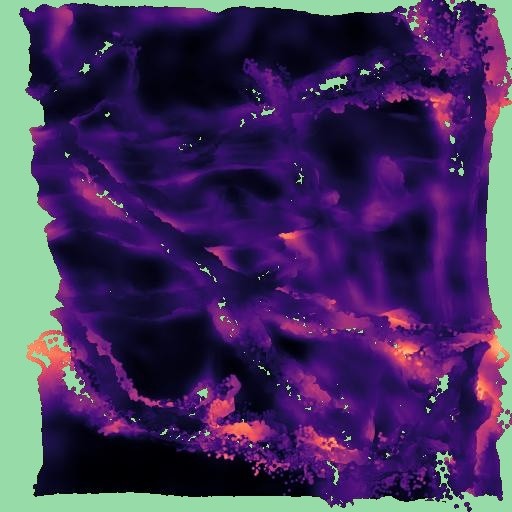} &
         \includegraphics[width=0.16\textwidth]{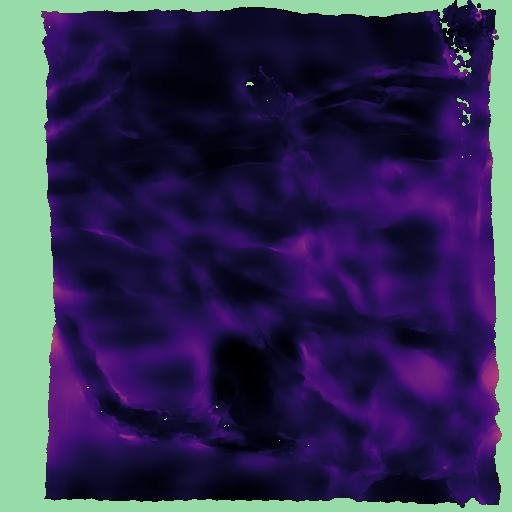} &
         \includegraphics[width=0.16\textwidth]{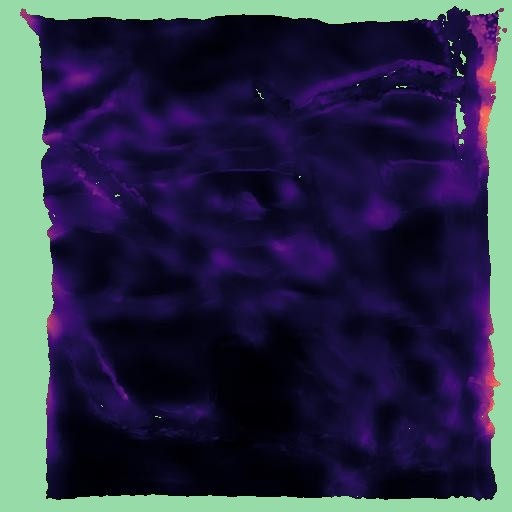} \vspace{-0.1cm}\\
         & MEt3R scores:  & 0.2394 & 0.3091 & 0.1938 & 0.1568 & 0.1118 \\
         
    \end{tabular}\vspace{-0.3cm}
    \caption{\textbf{Visualization of MEt3R score \cite{asim2025met3r} maps on Middlebury dataset \cite{scharstein2014middlebury}.} We report, from left to right, the original left image for four samples in the dataset, followed by the MEt3R score maps computed between it and the right images generated by different methods. The coloring is according to the \texttt{magma} colormap, with green regions representing occlusions (discarded by MEt3R when computing the average score). Under each score map, we report the global score computed by MEt3R (the lower, the better).}
    \label{fig:met3r_middlebury}
\end{figure*}

\begin{figure*}[t]
    \centering
    \renewcommand{\tabcolsep}{1pt}
    \begin{tabular}{ccccccc}
        & Left Image & ZeroStereo \cite{wang2025zerostereo} & StereoDiffusion \cite{wang2024stereodiffusion} & GenStereo \cite{qiao2025genstereo} & Lyra \cite{bahmani2025lyra} & \bf StereoSpace (ours) \smallskip\\
    
         \rotatebox[origin=l]{90}{\quad\quad\quad\texttt{215}} & \includegraphics[width=0.16\textwidth]{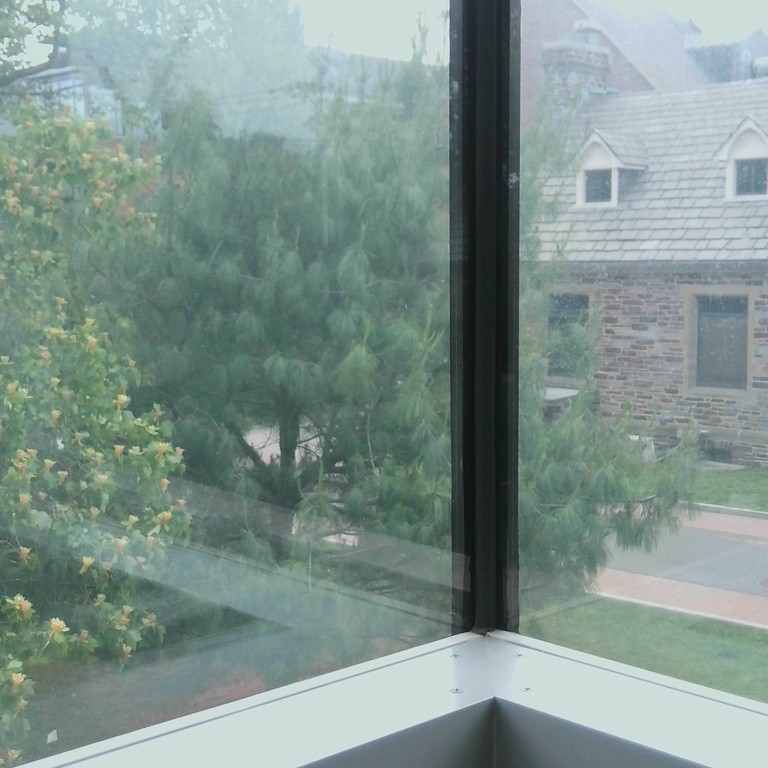} &
         \includegraphics[width=0.16\textwidth]{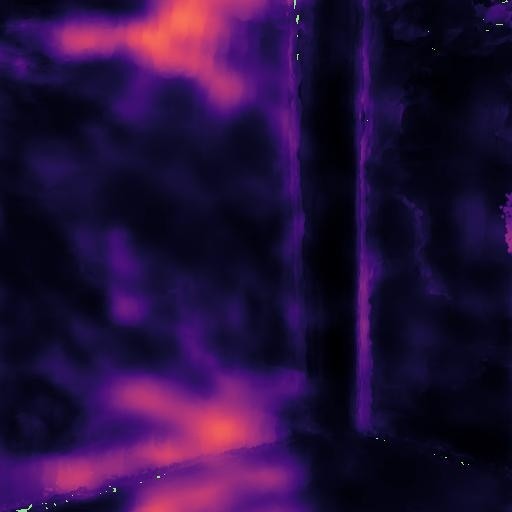} &
         \includegraphics[width=0.16\textwidth]{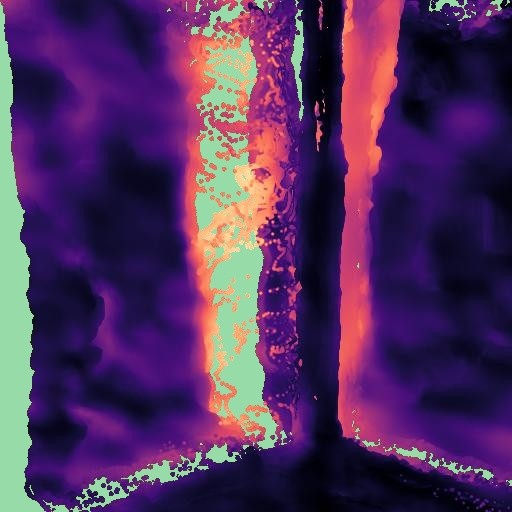} &
         \includegraphics[width=0.16\textwidth]{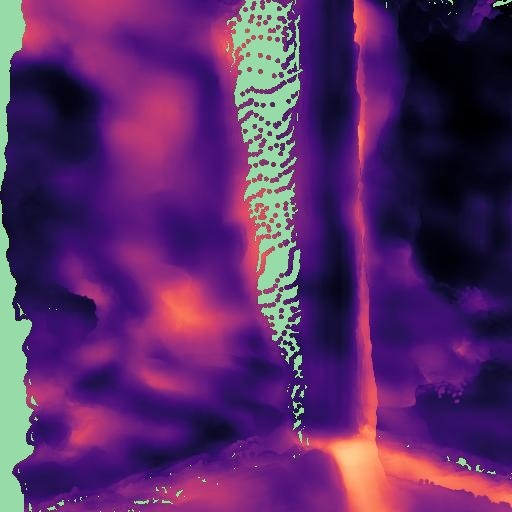} &
         \includegraphics[width=0.16\textwidth]{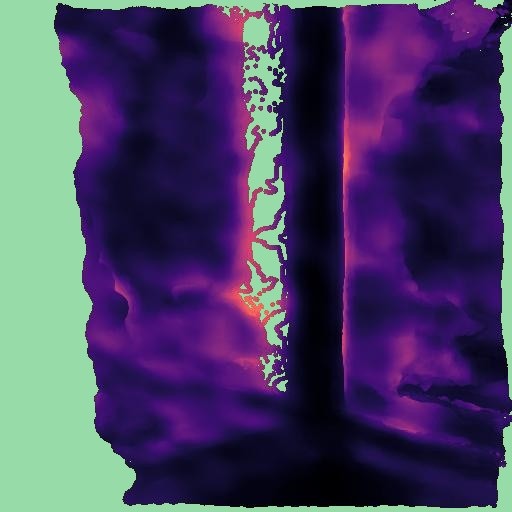} &
         \includegraphics[width=0.16\textwidth]{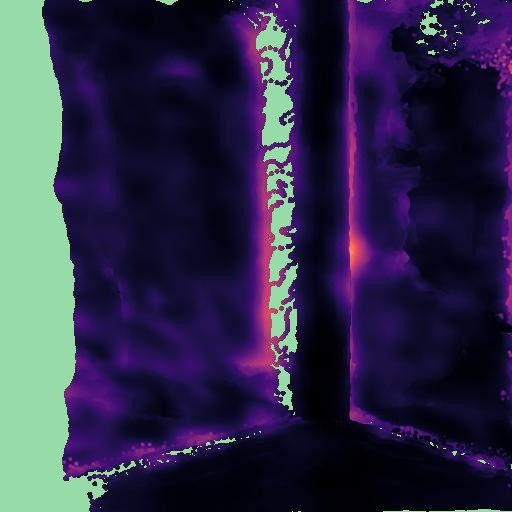} \vspace{-0.1cm}\\
         & MEt3R scores:  & 0.1553 & 0.3044 & 0.3088 & 0.1901 & 0.1367 \smallskip\\

         \rotatebox[origin=l]{90}{\quad\quad\quad\texttt{249}} &\includegraphics[width=0.16\textwidth]{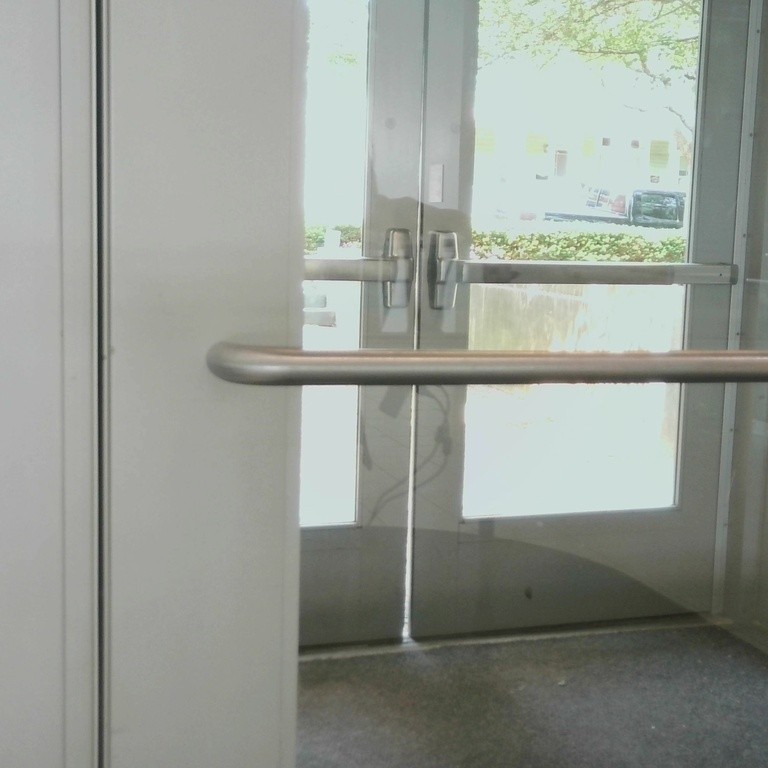} &
         \includegraphics[width=0.16\textwidth]{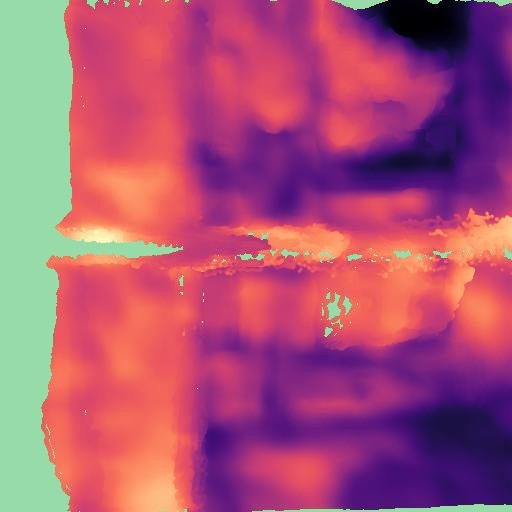} &
         \includegraphics[width=0.16\textwidth]{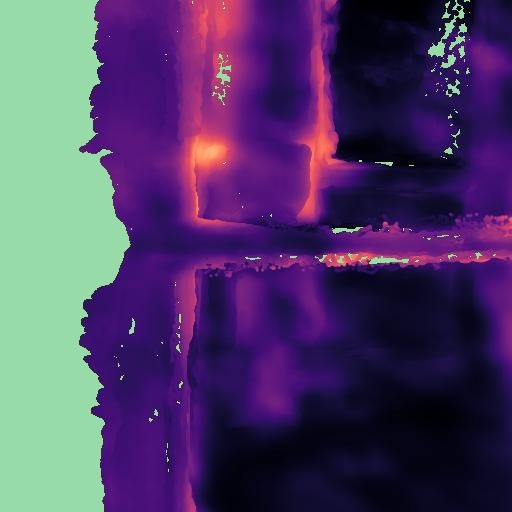} &
         \includegraphics[width=0.16\textwidth]{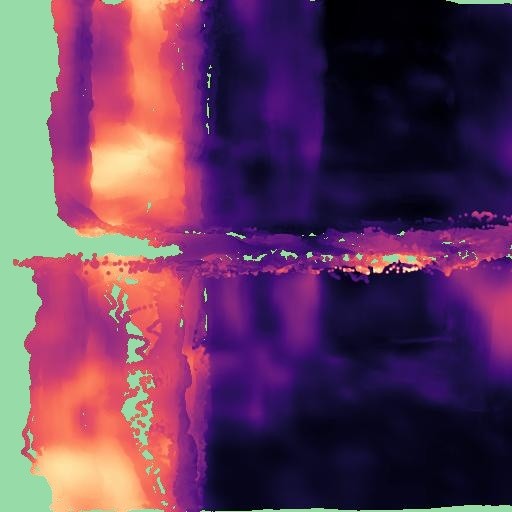} &
         \includegraphics[width=0.16\textwidth]{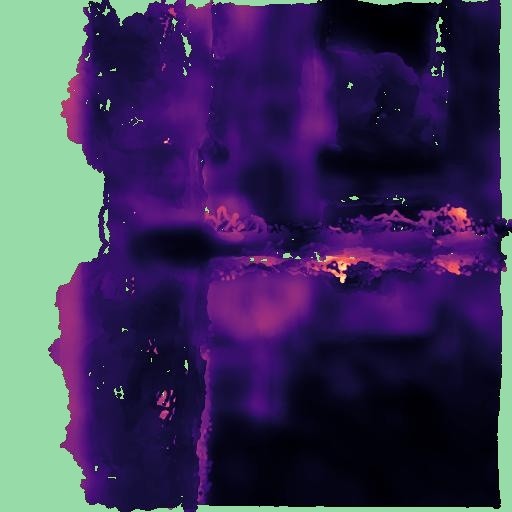} &
         \includegraphics[width=0.16\textwidth]{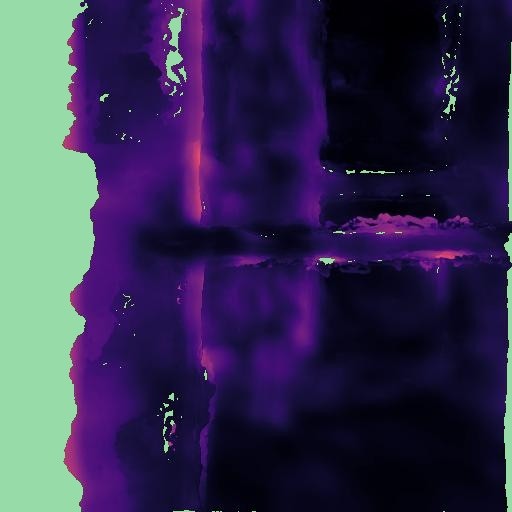} \vspace{-0.1cm}\\
         & MEt3R scores:  & 0.5380 & 0.2258 & 0.3114 & 0.1697 & 0.1351 \smallskip\\

         \rotatebox[origin=l]{90}{\quad\quad\quad\texttt{250}} & \includegraphics[width=0.16\textwidth]{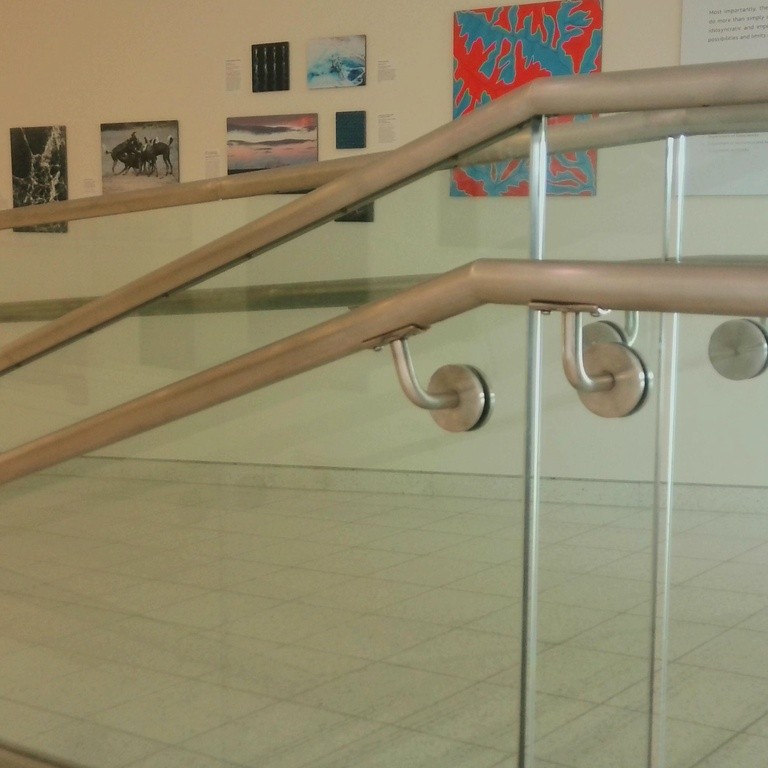} &
         \includegraphics[width=0.16\textwidth]{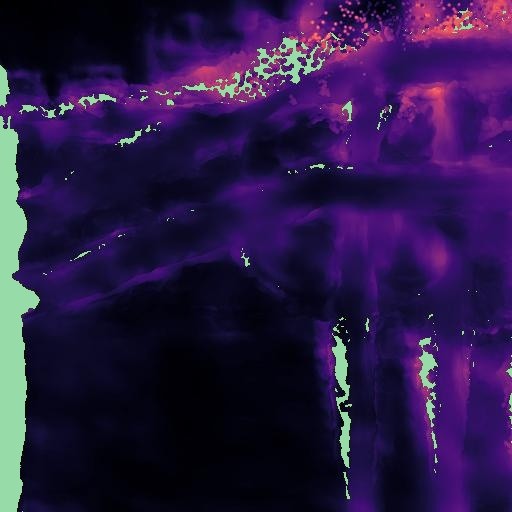} &
         \includegraphics[width=0.16\textwidth]{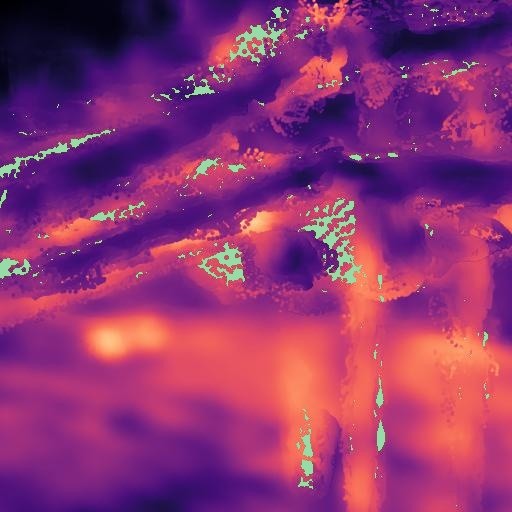} &
         \includegraphics[width=0.16\textwidth]{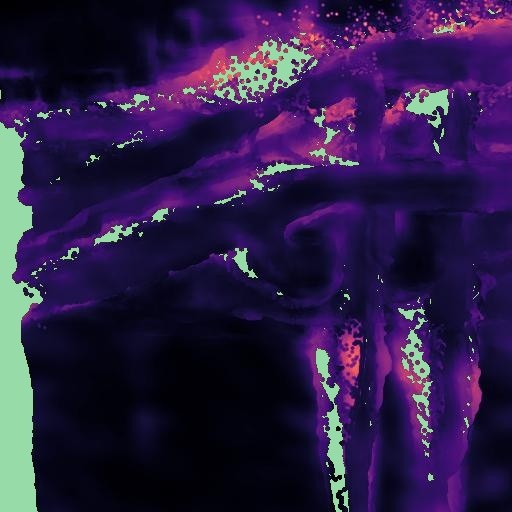} &
         \includegraphics[width=0.16\textwidth]{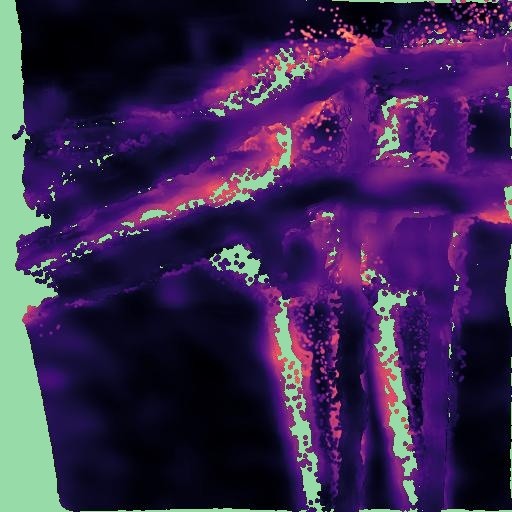} &
         \includegraphics[width=0.16\textwidth]{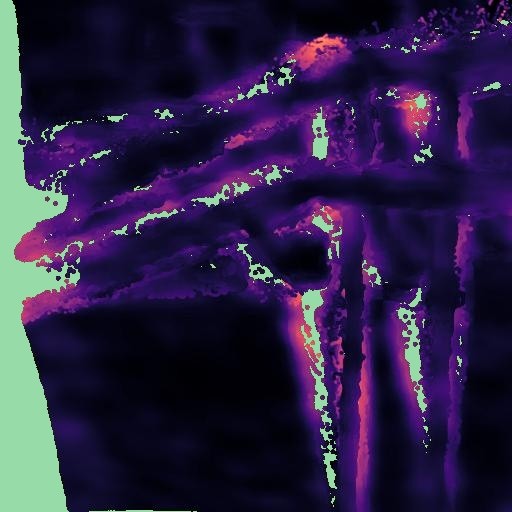} \vspace{-0.1cm}\\
         & MEt3R scores:  & 0.1542 & 0.4504 & 0.1523 & 0.1923 & 0.1449 \smallskip\\

         \rotatebox[origin=l]{90}{\quad\quad\quad\texttt{252}} &\includegraphics[width=0.16\textwidth]{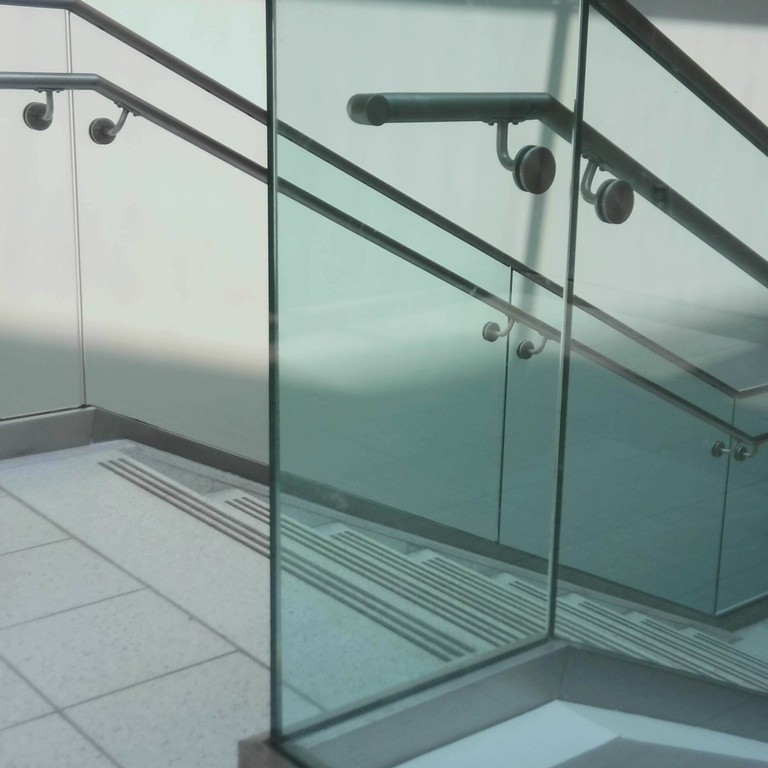} &
         \includegraphics[width=0.16\textwidth]{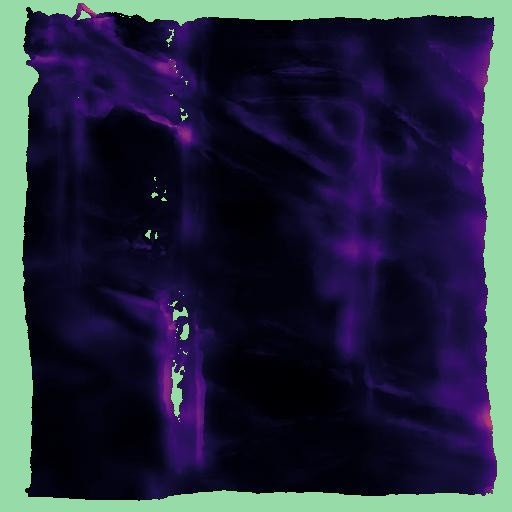} &
         \includegraphics[width=0.16\textwidth]{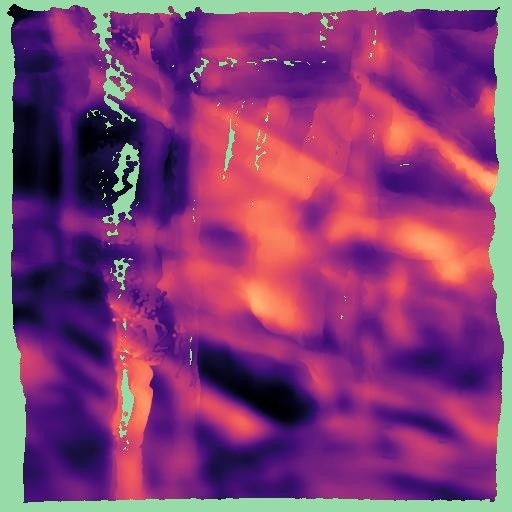} &
         \includegraphics[width=0.16\textwidth]{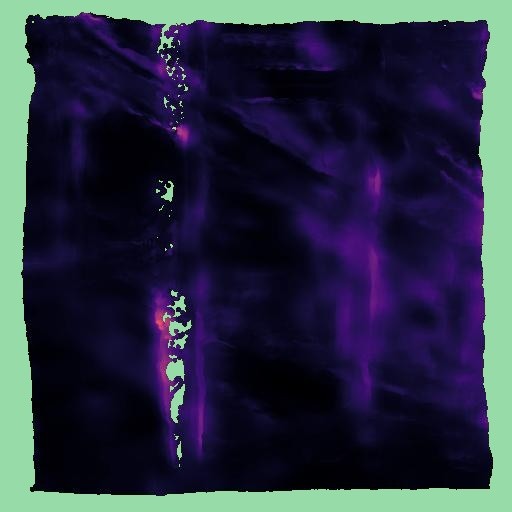} &
         \includegraphics[width=0.16\textwidth]{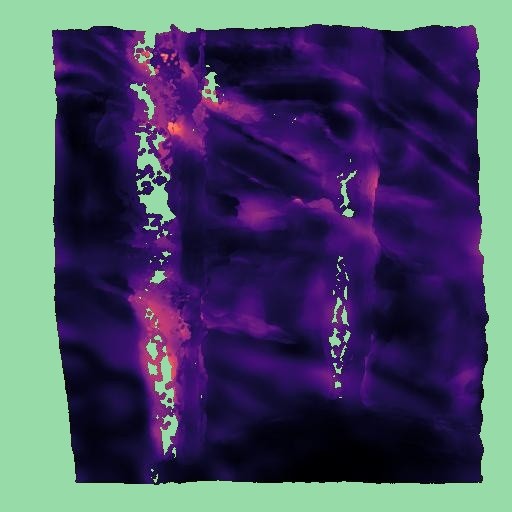} &
         \includegraphics[width=0.16\textwidth]{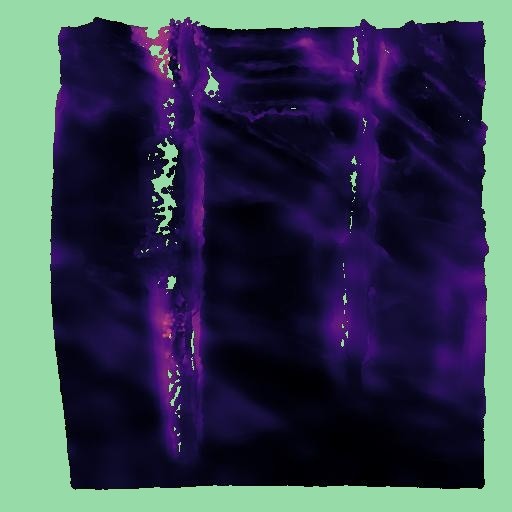} \vspace{-0.1cm}\\
         & MEt3R scores:  & 0.0893 & 0.4622 & 0.0942 & 0.1717 & 0.1114 \smallskip\\

         \rotatebox[origin=l]{90}{\quad\quad\quad\texttt{285}} &\includegraphics[width=0.16\textwidth]{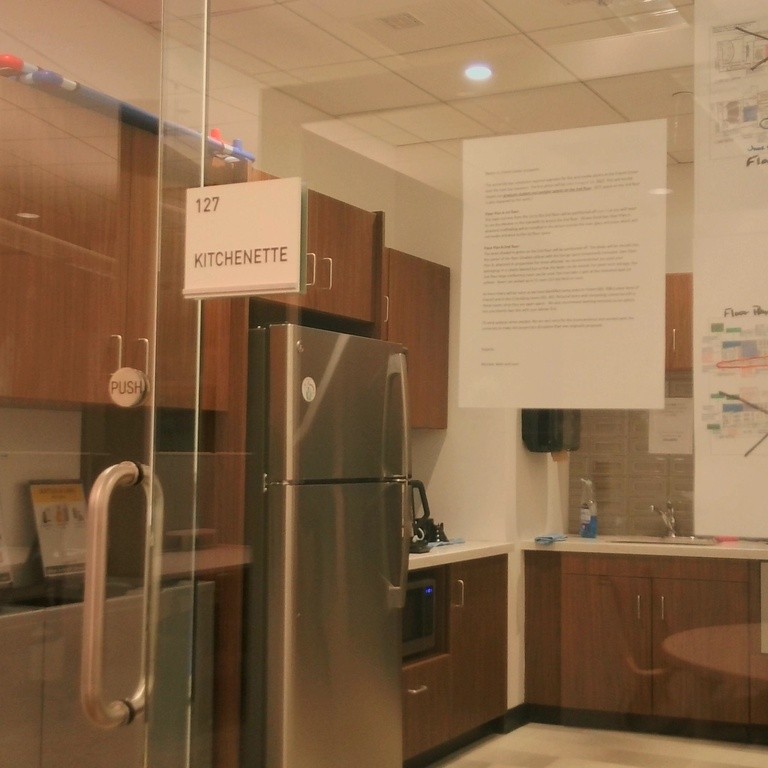} &
         \includegraphics[width=0.16\textwidth]{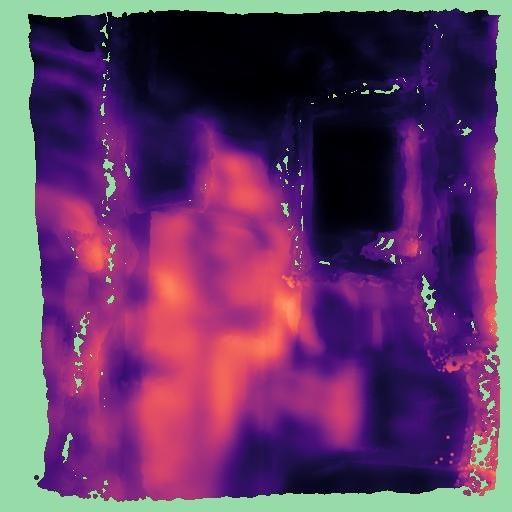} &
         \includegraphics[width=0.16\textwidth]{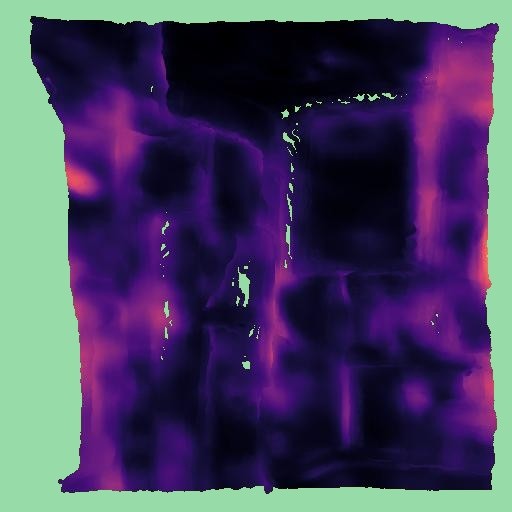} &
         \includegraphics[width=0.16\textwidth]{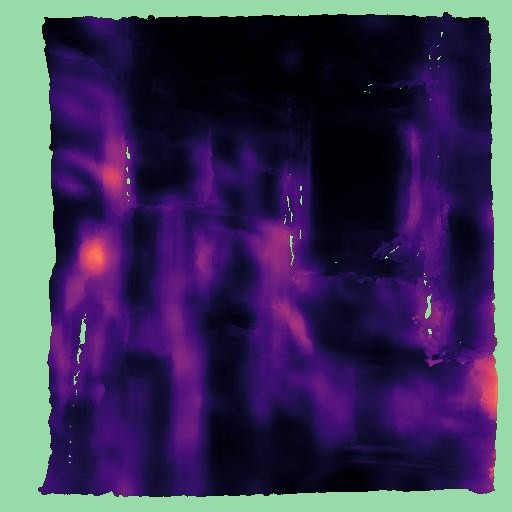} &
         \includegraphics[width=0.16\textwidth]{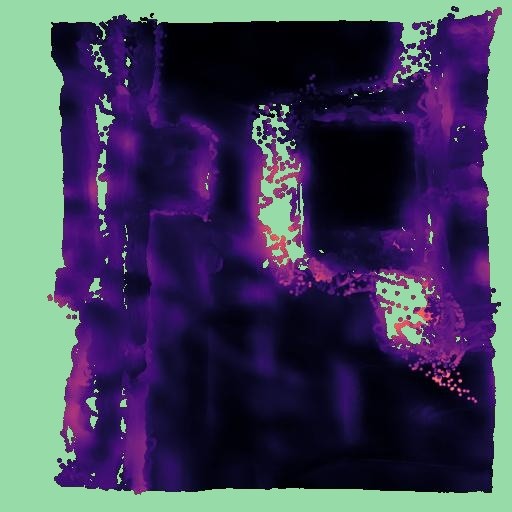} &
         \includegraphics[width=0.16\textwidth]{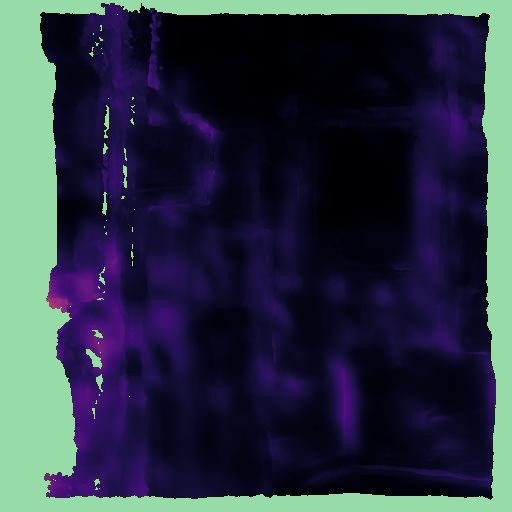} \vspace{-0.1cm}\\
         & MEt3R scores:  & 0.2926 & 0.1685 & 0.1493 & 0.1454 & 0.0852 \smallskip\\

         \rotatebox[origin=l]{90}{\quad\quad\quad\texttt{297}} &\includegraphics[width=0.16\textwidth]{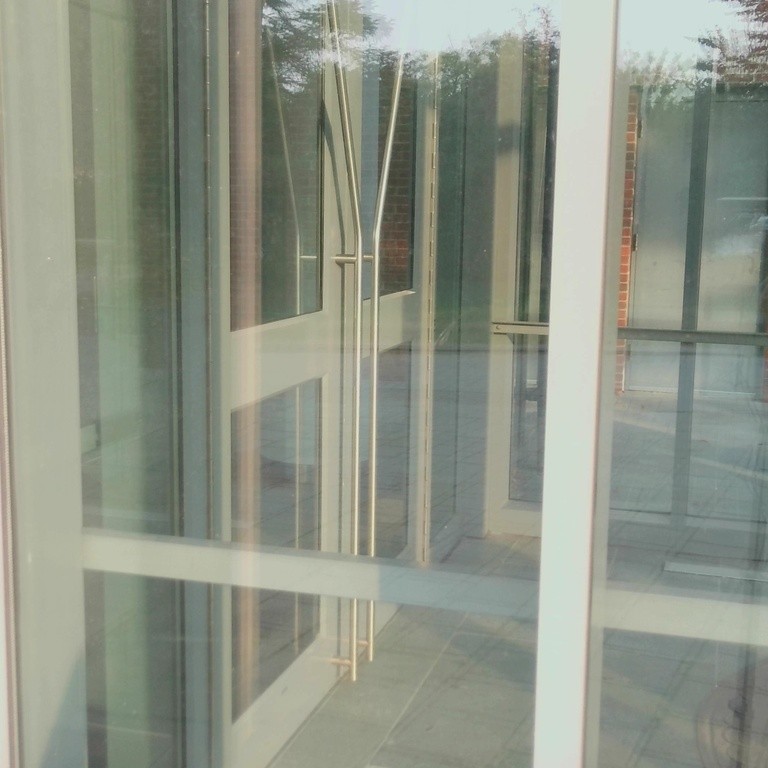} &
         \includegraphics[width=0.16\textwidth]{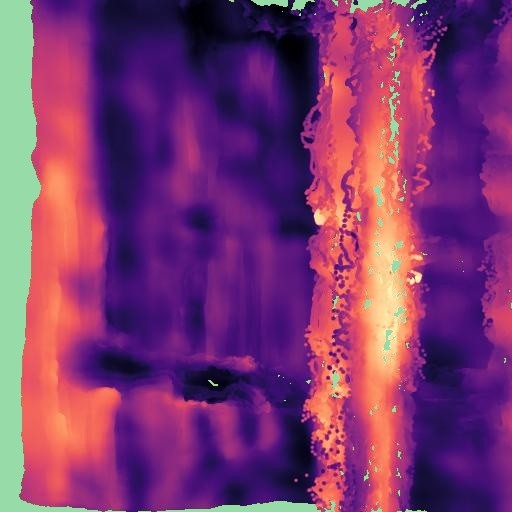} &
         \includegraphics[width=0.16\textwidth]{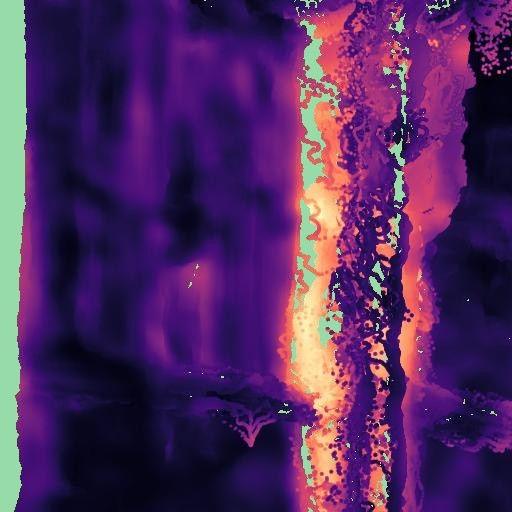} &
         \includegraphics[width=0.16\textwidth]{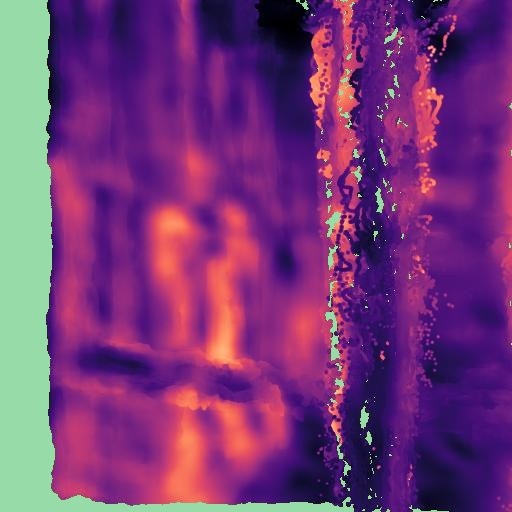} &
         \includegraphics[width=0.16\textwidth]{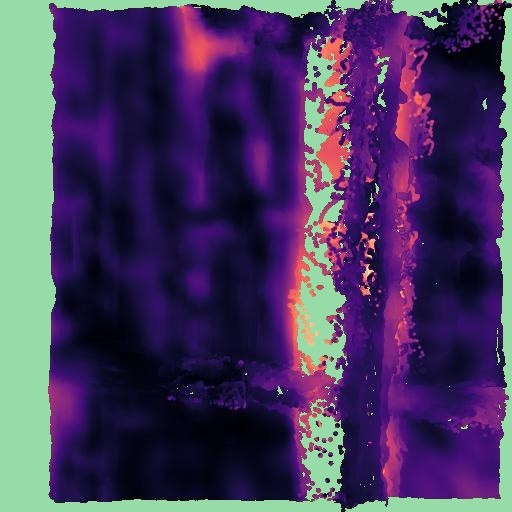} &
         \includegraphics[width=0.16\textwidth]{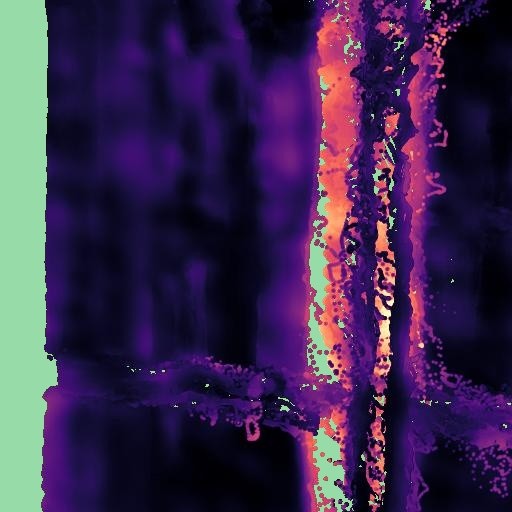} \vspace{-0.1cm}\\
         & MEt3R scores:  & 0.4125 & 0.2727 & 0.3697 & 0.2200 & 0.1936 \\
    \end{tabular}\vspace{-0.3cm}
    \caption{\textbf{Visualization of MEt3R score \cite{asim2025met3r} maps on LayeredFlow dataset \cite{wen2024layeredflow}.} We report, from left to right, the original left image for four samples in the dataset, followed by the MEt3R score maps computed between it and the right images generated by different methods. The coloring is according to the \texttt{magma} colormap, with green regions representing occlusions (discarded by MEt3R when computing the average score). Under each score map, we report the global score computed by MEt3R (the lower, the better).}
    \label{fig:met3r_layered}
\end{figure*}

\begin{figure*}[t]
  \centering
  \begin{overpic}[width=\linewidth]{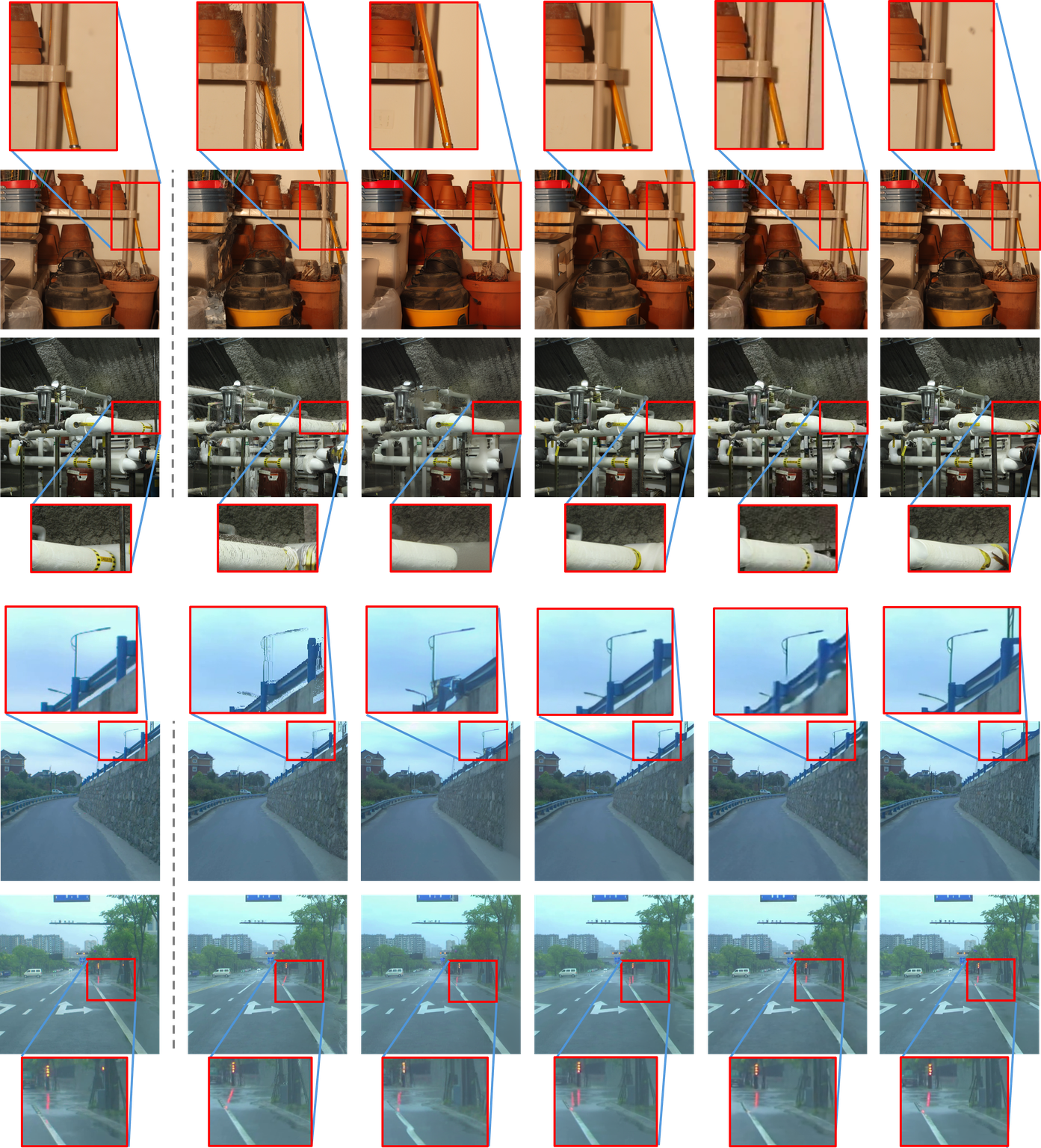}
  \put (1,101) {Ground truth}
  \put (17,101) {ZeroStereo \cite{wang2025zerostereo}}
  \put (31,101) {StereoDiffusion \cite{wang2024stereodiffusion}}
  \put (47,101) {GenStereo \cite{qiao2025genstereo}}
  \put (64,101) {Lyra \cite{bahmani2025lyra}}
  \put (75,101) {\textbf{StereoSpace (ours)}}
  \end{overpic}
  \caption{\textbf{Qualitative results on Middlebury \cite{scharstein2014middlebury} and DrivingStereo \cite{yang2019driving} datasets.} 
  }
  \label{fig:qualitative_suppl_1}
\end{figure*}

\begin{figure*}[t]
  \centering
  \begin{overpic}[width=\linewidth]{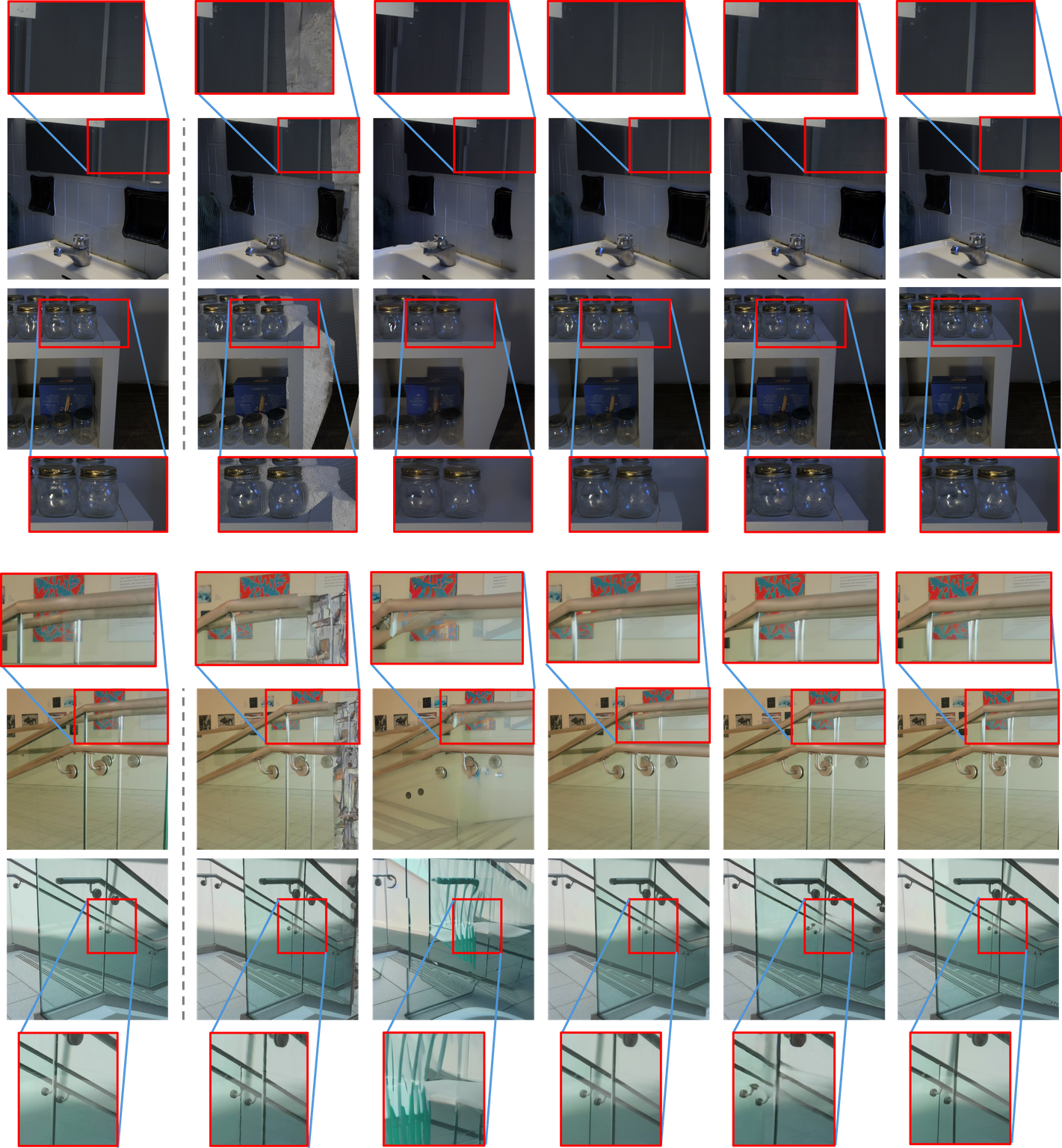}
  \put (2,101) {Ground truth}
  \put (17,101) {ZeroStereo \cite{wang2025zerostereo}}
  \put (31,101) {StereoDiffusion \cite{wang2024stereodiffusion}}
  \put (49,101) {GenStereo \cite{qiao2025genstereo}}
  \put (66,101) {Lyra \cite{bahmani2025lyra}}
  \put (77,101) {\textbf{StereoSpace (ours)}}
  \end{overpic}
  \caption{\textbf{Qualitative results on Booster \cite{ramirez2022booster} and LayeredFlow \cite{wen2024layeredflow} datasets.}  
  }
  \label{fig:qualitative_suppl_2}
\end{figure*}

\section{Evaluation Sets Composition}
\label{sec:supp_evaluation}

We evaluate on four real-world stereo benchmarks: Middlebury 2014~\cite{scharstein2014middlebury}, DrivingStereo~\cite{yang2019driving}, Booster~\cite{ramirez2022booster}, and LayeredFlow~\cite{wen2024layeredflow}, none of these datasets are used during training. For Middlebury 2014, we follow the official evaluation protocol and use the same split as GenStereo and StereoDiffusion. For DrivingStereo, we uniformly sample 50 stereo pairs from the 500 released pairs for each weather condition, yielding 200 evaluation images in total. For Booster, we report results on the union of the official train and test splits. For LayeredFlow, we evaluate on the 300 real-world stereo pairs from the official validation and evaluation splits of the benchmark. In our mixed training set, we only include the separate synthetic stereo split of LayeredFlow rendered in Blender.

\section{Evaluation with Conventional Metrics}

For the sake of completeness, we report classical metrics such as PSNR, SSIM, and LPIPS in Tab. \ref{tab:stereo-metrics}. In most cases, StereoSpaces ranks 2nd, just behind either Lyra or GenStereo. However, as discussed in the paper, these metrics fail at faithfully assessing both geometric consistency and perceptual comfort. 

\section{Qualitative Results}
\label{sec:supp_qualitative}

\noindent We conclude by reporting additional qualitative results.

\noindent\textbf{Native Multi-baseline Inference.} Thanks to our viewpoint conditioning mechanism and training schedule, StereoSpace natively supports  the generation images at arbitrary horizontal baselines on either side of the input view (Fig.~\ref{fig:multi-baseline-inference}). Warping-based frameworks can also be extended to this setting, but require either manually rescaling the monocular disparity used for warping or  flipping the image to synthesize views on the opposite side of the reference image.

\noindent\textbf{MEt3R Score Map Visualization.}
To better illustrate the margin in MEt3R~\cite{asim2025met3r} scores across methods, we visualize the per-pixel score maps produced by MEt3R before averaging, which makes local differences between methods directly observable.
Fig.~\ref{fig:met3r_middlebury} shows results on five samples from the Middlebury dataset.
Overall, warping-based methods produce higher MEt3R scores, particularly near depth discontinuities, whereas both Lyra and StereoSpace exhibit substantially lower scores.
We also observe that the overall quality of the generated images affects the ability of MEt3R (through its MASt3R~\cite{leroy2024grounding} backbone) to accurately estimate the relative transformation between the original and synthesized images.
This is visible in the green regions, which represent areas of the scene that are not overlapping between the two images.
Warping-based solutions show larger non-overlapping regions at both the top and bottom of the frame, although these areas should overlap in a stereo setup.
By contrast, such regions are often much narrower (or absent) in the Lyra and StereoSpace score maps.

Fig.~\ref{fig:met3r_layered} collects six samples from the LayeredFlow dataset \cite{wen2024layeredflow}. The multi-layered geometry peculiar to these scenes leads to a significant increase in the MEt3R scores, in particular for warping-based solutions. 
As discussed, this effect is mostly related to the use of estimated depth, which fails to account for the multiple layers in the scene. 
On the contrary, StereoSpace shines in these cases, as it does not inherit the limitations inherent to warping-based methods.

\noindent\textbf{Further Qualitative Comparisons.} Finally, to highlight the superior realism of StereoSpace, we present extensive qualitative comparisons with images generated by all competing methods considered in our evaluation. Fig.~\ref{fig:qualitative_suppl_1} reports two samples from Middlebury \cite{scharstein2014middlebury} and two from the DrivingStereo \cite{yang2019driving} datasets. In the Middlebury examples, we highlight the bleeding artifacts introduced by ZeroStereo, as well as the interpolation effects between foreground and background objects produced by StereoDiffusion, clearly visible in the \texttt{Shopvac} scene, and both caused by warping. We can also appreciate how Lyra itself, despite its high realism, tends to introduce oversmoothing. 
On DrivingStereo, we observe fewer artifacts due to simpler scene geometry and smaller disparities from larger depths. Nevertheless, StereoSpace still demonstrates superior reconstruction of thin structures.

Fig.~\ref{fig:qualitative_suppl_2} shows two scenes from Booster \cite{ramirez2022booster} and two from LayeredFlow \cite{wen2024layeredflow}. At the top, we can notice how reflective and transparent objects in general are a significant challenge for methods relying on estimated depth, as highlighted by the artifacts produced in correspondence with the mirror or the deformations visible in the jars. At the bottom, depth-based approaches again struggle to deal with transparent surfaces that induce multiple depth layers, while StereoSpace maintains more faithful geometry and appearance.

\end{document}